\documentclass[runningheads]{llncs}

\usepackage{chngcntr}

\usepackage{eccv}

\usepackage{eccvabbrv}

\usepackage{graphicx}
\usepackage{booktabs}
\usepackage{multirow}
\usepackage{array}
\usepackage{amsmath}
\usepackage{amsopn}
\usepackage{pifont}%
\usepackage{arydshln}

\usepackage[accsupp]{axessibility}  %

\usepackage{hyperref}

\usepackage{orcidlink}

\begin{document}

\def \islam {$I^2$-SLAM}

\title{\islam: Inverting Imaging Process for\\Robust Photorealistic Dense SLAM}

\titlerunning{\islam}

\author{Gwangtak Bae$^\star$\inst{1}\orcidlink{0000-0001-8943-3205} \and
Changwoon Choi$^\star$\inst{1}\orcidlink{0000-0001-5748-6003} \and
Hyeongjun Heo\inst{1}\orcidlink{0009-0009-2934-1064} \and \\Sang Min Kim\inst{1}\orcidlink{0009-0002-9718-1196} \and Young Min Kim$^\dagger$\inst{1,2}\orcidlink{0000-0002-6735-8539}}

\def\thefootnote{$\star$}\footnotetext{Authors contributed equally to this work.}
\def\thefootnote{$\dagger$}\footnotetext{Young Min Kim is the corresponding author.}
\authorrunning{G.~Bae, C.~Choi, H.~Heo, S.M.~Kim, and Y.M.~Kim}

\institute{Dept. of Electrical and Computer Engineering, Seoul National University
\and
INMC \& IPAI, Seoul National University}

\maketitle
\begin{figure}
    \centering
    \includegraphics[width=\textwidth]{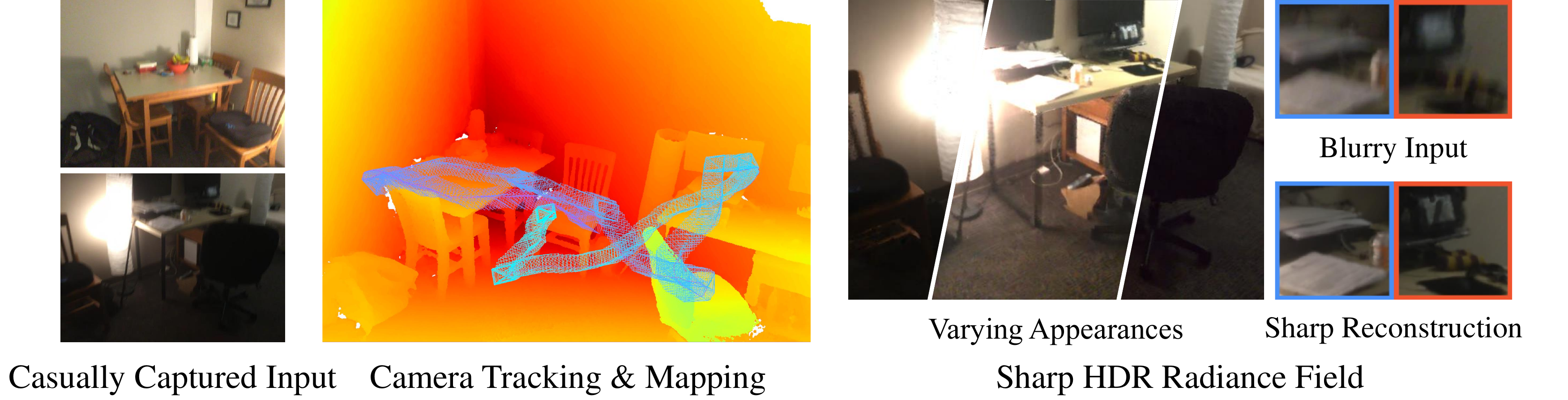}
    \caption{We propose \islam, a SLAM pipeline with a physical image formation process. We can reconstruct photorealistic and sharp HDR maps from casually captured videos which contain severe motion blur and varying appearances.}
    \label{fig:teaser}
\end{figure}

\begin{abstract}
 We present an inverse image-formation module that can enhance the robustness of existing visual SLAM pipelines for casually captured scenarios.
Casual video captures often suffer from motion blur and varying appearances, which degrade the final quality of coherent 3D visual representation.
We propose integrating the physical imaging into the SLAM system, which employs linear HDR radiance maps to collect measurements.
Specifically, individual frames aggregate images of multiple poses along the camera trajectory to explain prevalent motion blur in hand-held videos.
Additionally, we accommodate per-frame appearance variation by dedicating explicit variables for image formation steps, namely white balance, exposure time, and camera response function.
Through joint optimization of additional variables, the SLAM pipeline produces high-quality images with more accurate trajectories.
Extensive experiments demonstrate that our approach can be incorporated into recent visual SLAM pipelines using various scene representations, such as neural radiance fields or Gaussian splatting.
\href{https://3d.snu.ac.kr/publications/I2SLAM}{{Project website}}

\end{abstract}

\section{Introduction}
\label{sec:introduction}

Simultaneous localization and mapping (SLAM) builds a map of the environment during deployment, which can be utilized in various applications, including VR/AR~\cite{covolan2020mapping, jinyu2019survey}, robotic navigation~\cite{gupta2017cognitive, oleynikova2017voxblox, chaplot2020learning}, and collision handling~\cite{chen2023catnips}.
Traditional 3D SLAM approaches typically use geometric representations like points/surfels~\cite{keller2013real, whelan2015elasticfusion, whelan2016elasticfusion, schops2019bad, teed2021droid}, mesh~\cite{bloesch2019learning}, voxel grids~\cite{newcombe2011kinectfusion, bylow2013real}, or voxel hashing~\cite{niessner2013real, dai2017bundlefusion}.
Recent visual SLAM approaches additionally capture visual appearances incorporating advances in Neural Radiance Fields (NeRF)~\cite{mildenhall2021nerf} and its variants~\cite{muller2022instant, kerbl20233d}.
They can synthesize photorealistic images of the environment and open up new possibilities in complex downstream tasks such as detailed semantic scene understanding~\cite{jatavallabhula2023conceptfusion}, language-guided manipulation~\cite{shen2023F3RM}, or visual navigation~\cite{shafiullah2022clip}.
Additionally, neural representations can fill unseen regions with smooth geometric estimation and require low-memory footprint~\cite{sucar2021imap, Ortiz:etal:iSDF2022, wang2023co}.
3D visual representations can achieve real-time performance using voxelized hash grid~\cite{muller2022instant} or 3D Gaussian Splatting (3DGS)~\cite{kerbl20233d}.

Despite many works that build visual representations using the SLAM framework, most do not maintain their performance in real-world scenarios.
Casually captured videos, the standard input for visual SLAM systems, suffer from two prevalent challenges: 1) \textit{motion blur} due to camera movement and 2) \textit{varying appearances} resulting from auto exposure and white balancing adjustments as demonstrated in~\cref{fig:teaser}.
The degradation in images serves as a critical bottleneck for the accuracy of the map and the pose estimation, and the error accumulates due to the incremental nature of SLAM, significantly reducing the overall quality.

This work tackles the prevailing challenges by attaching the physical image formation process that directly models the aforementioned variations.
Then, we can directly optimize for the correct camera poses and raw measurements via an analysis-by-synthesis approach.
Specifically, the motion-blurred image is compared against observations integrated along the estimated camera trajectories during a window of exposure time instead of an image from a single camera pose.
Our pipeline approximates the camera poses as a linear movement and optimizes the start and end poses.
The global trajectory estimated in the SLAM pipeline guides the initial blur movement with in-camera parameters, such as exposure time.
At the same time, we match the per-frame appearance variation against simulated images and jointly optimize a differentiable tone mapper composed of exposure time, white balance function, and camera response function (CRF).
We employ high dynamic range (HDR) radiance fields as a map representation to linearize the color space, which reflects the actual light intensity maps.
The HDR maps simplify modeling appearance variations and produce significantly more realistic motion blur effects~\cite{10.1145/258734.258884}.

As our formulation, coined \islam, inverts the actual measurement steps, the module can augment any dense visual SLAM pipelines that use image inputs, including implicit neural networks and 3D Gaussians.
Our extensive experiments demonstrate that it can robustly reconstruct sharp HDR maps from RGB/RGBD streams afflicted with severe motion blur and varying appearances.

Our technical contributions can be summarized as follows:
\begin{itemize}
    \item We present \islam, integrating the image formation process into the visual SLAM approaches to overcome dominant challenges in real-world captures.
    \item Incorporating 3D maps composed of linear radiance values, our formulation jointly optimizes the approximate movement for the motion blur and tone-mapping functions for appearance variations during the SLAM framework.
    \item We propose an initialization and regularization method to stabilize the blur movement optimization to align with the estimated global camera trajectory, utilizing the SLAM setup.
    \item We enhance the robustness and performance of recent visual SLAM approaches in the real-world and our synthetic dataset, which contains severe motion blur and varying appearance.
\end{itemize}

\section{Related Works}
\label{sec:related_works}
\subsubsection{Dense Visual SLAM}

Visual SLAM methods use images as input and first started using multi-view geometry between sparse image features to estimate camera trajectories.
DTAM~\cite{newcombe2011dtam} further demonstrated building a projective photometric cost volume, a dense map representation that unlocks the opportunity for combining various image-based applications. 
As deep neural networks prosper in computer vision, parts of the visual SLAM pipeline have also been successfully deployed to use latent representations~\cite{bloesch2018codeslam} or assist the depth estimation~\cite{teed2021droid}.

Subsequent advancements in neural visual SLAM benefit from the technical innovations in neural implicit representation or improved novel representations. 
iMAP~\cite{sucar2021imap} employed implicit neural map representation for SLAM.
NICE-SLAM~\cite{zhu2022nice} demonstrates that the multi-resolution feature grid representation can improve the speed and resolve the forgetting problem in large-scale scenes. 
Co-SLAM~\cite{wang2023co} explores the coordinate and parametric encodings to achieve efficient and accurate mapping. 
NeRF-SLAM~\cite{rosinol2023nerf} eliminates the dependency on depth by utilizing DROID-SLAM~\cite{teed2021droid}.
Recent results favor explicit representations when high speed and photometric quality are desired.
Point-SLAM~\cite{sandstrom2023point} creates photorealistic 3D maps using a dynamic neural point cloud. Since the introduction of 3D Gaussian Splatting~\cite{kerbl20233d}, several concurrent works rapidly deploy them for 3D map representation for SLAM~\cite{keetha2023splatam, matsuki2023gaussian, huang2023photo, yugay2023gaussian}.

\subsubsection{Motion Deblurring}

Motion blur significantly affects the quality of many computer vision tasks and has long been investigated as an active research topic.
Traditional approaches assume a convolution kernel for the motion blur and convert the operation via deconvolution~\cite{fergus2006removing, shan2008high, cho2009fast, whyte2012non}. 
The convolution operation naturally transfers to convolutional neural networks (CNN), and more recent works estimate the non-uniform motion blur field~\cite{sun2015learning}, complex Fourier coefficients~\cite{chakrabarti2016neural}, or dense motion flow~\cite{chen2018learning}.
Recent neural representations~\cite{mildenhall2021nerf, muller2022instant} also suffer from performance degradation when handling blurred input images. 
Various works account for the blur operation and reconstruct a sharp 3D map even when input images are blurred~\cite{ma2022deblur, wang2023bad}. 

The input measurements for the visual SLAM are susceptible to motion blur, as a moving camera captures an image for a duration of time. 
We linearize the camera trajectory during the exposure time and optimize the poses to match the blurred input, similar to~\cite{wang2023bad}.
However, we further exploit the SLAM setup and regularize the blur movement to be aligned with the estimated camera trajectory.

\subsubsection{High Dynamic Range Recovery}
Another critical issue for high-quality visual SLAM is that the pixel values are inconsistent with different exposures in input frames.
Classical works on HDR imaging demonstrate that a color space of an HDR radiance map can successfully combine multiple images of different exposures~\cite{10.1145/258734.258884}.
HDR radiances are linear to the scene radiance values and lead to better results in image processing and image-based modeling~\cite{chen2018learning, hu2018exposure, zhang2019synthetic}.
Our map representation also models the actual HDR radiance values, and we provide explicit variables to model the physical process to map the pixel values, namely the exposure time, white balance, and camera response function per frame.
Recent works on novel-view synthesis and 3D reconstruction also achieve improved results as they optimize the HDR radiances with exposure times~\cite{jun2022hdr,huang2022hdr,ruckert2022adop, mildenhall2022nerf}.
Our work integrates factors such as exposure time into a coherent formulation for tone mapping and motion blur and stabilizes the overall optimization.

\begin{figure}[t]
    \centering
    \includegraphics[width=\linewidth]
    {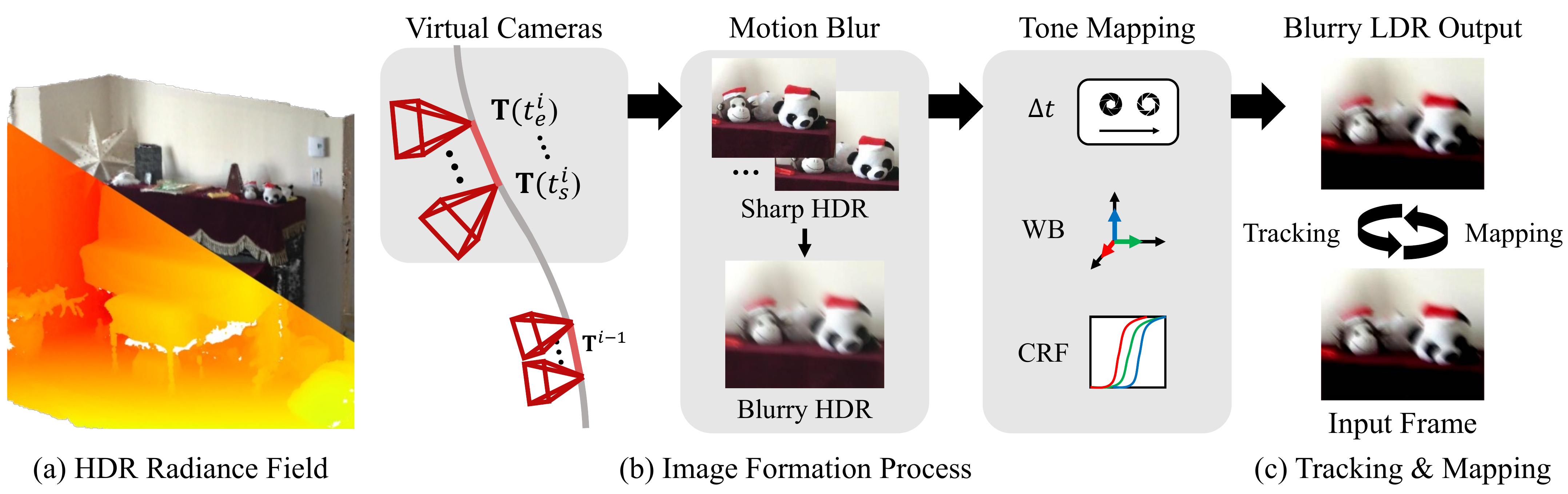}
    \caption{Method overview. (a) We reconstruct a sharp HDR radiance field map. (b) Motion blur is simulated by integration of sharp images, which are obtained from virtual camera poses during the exposure time. Then we obtain the blurry LDR image by applying differentiable tone mapping module. (c) SLAM methods simultaneously perform tracking and mapping from degraded images to reconstruct a sharp HDR map.}
    \label{fig:method_overview}
\end{figure}

\section{Method}
\label{sec:method}
\islam~is a generic method that can be combined with any existing photo-realistic dense SLAM methods.
In \cref{subsec:map_represenatation}, we first review the rendering techniques for two representative map representations used in our experiments: Neural Radiance Fields (NeRF)~\cite{mildenhall2022nerf} and 3D Gaussian Splatting (3DGS)~\cite{kerbl20233d}.
Then we describe our physical image formation process to render motion-blurred and appearance-varying images in \cref{subsec:image_formation_process}.
Finally, we explain how to integrate our image formation process into existing SLAM pipelines in \cref{subsec:mapping_and_tracking}.

\subsection{Preliminaries: Rendering for Photorealistic Dense SLAM}
\label{subsec:map_represenatation}

Our approach seamlessly integrates with various methods for generating color $\mathbf{c}$ and, optionally, depth $d$ for each pixel for a given camera pose.
Owing to their photorealistic quality and ease of training, most recent visual SLAM approaches are based on variations stemming from either NeRF~\cite{mildenhall2021nerf} or 3DGS~\cite{kerbl20233d}.
This section reviews the general formulations of the two approaches, elucidating the process of generating color and depth information.

Let $\mathbf{c}(\mathbf{T}, \mathbf{p})$  and $d(\mathbf{T}, \mathbf{p})$ denote the color and depth at pixel location $\mathbf{p}$ with camera pose $\mathbf{T}$.
For NeRF, we can obtain the color and depth of target pixels by marching rays and utilizing the volume rendering technique: 

\begin{equation}
    \mathbf{c}(\mathbf{T},\mathbf{p}) = \sum_{i=1}^{N}\tau_i(1-e^{-\sigma_i\delta_i})\mathbf{c}_i \quad \text{where} \;\: \tau_i = e^{-\sum_{j=1}^{i-1}\sigma_j\delta_j},
    \label{eq:rendering_nerf}
\end{equation}
\begin{equation}
    d(\mathbf{T},\mathbf{p}) = \sum_{i=1}^{N}\tau_i(1-e^{-\sigma_i\delta_i})d_i,
\end{equation}
where $N$ is the number of samples for each ray, and the camera determines the ray pose $\mathbf{T}$ and pixel location $\mathbf{p}$. $\mathbf{c}_i$ and $d_i$ are the corresponding color and depth of each sample along a ray observed from the camera center.
$\delta$ is the distance between consecutive samples. 
Color $\mathbf{c}_i$ and volume density $\sigma_i$ can be queried from MLP as in iMAP~\cite{sucar2021imap}, hierarchical feature grid with small MLP~\cite{zhu2022nice}, or multi-resolution hash grid~\cite{rosinol2023nerf}.

3DGS~\cite{kerbl20233d} also adopts the volume rendering formulation, so we can train the map differentiable to input image measurements.
However, 3DGS processes sparse explicit samples from 3D Gaussian primitives, which are faster.
We can render images by compositing $\mathcal{N}$ ordered 3D Gaussians as follows:
\begin{equation}
    \mathbf{c}(\mathbf{T},\mathbf{p}) = \sum_{i=1}^{\mathcal{N}}\mathbf{c}_i\alpha_i\prod_{j=1}^{i-1}(1-\alpha_j), \,\text{where}\,\alpha = oe^{-(\mathbf{x}-\mu)^T\Sigma^{-1}(\mathbf{x}-\mu)/2},
    \label{eq:rendering_gs}
\end{equation}

\begin{equation}
    d(\mathbf{T},\mathbf{p}) = \sum_{i=1}^{\mathcal{N}}d_i\alpha_i\prod_{j=1}^{i-1}(1-\alpha_j),
\end{equation}
where $\mathbf{c}$, $o\in[0,1]$, $\mathbf{\mu}\in\mathbb{R}^3$, and $\Sigma$ are the color, opacity, center position, and the covariance of a 3D Gaussian, respectively.

\subsection{Image Formation Process}
\label{subsec:image_formation_process}

When running a visual SLAM framework, we can augment additional mapping processes from HDR radiance to image measurements.
We reconstruct an HDR radiance map by making $c(\textbf{T},\textbf{p})$ in \cref{subsec:map_represenatation} to output linear HDR color.
The HDR pixel intensity $C^i_{\text{HDR}}(\textbf{p})$ captured during the exposure time $[t_s^i, t_e^i]$ for the $i$th frame is 
\begin{equation}
    C^i_{\text{HDR}}(\mathbf{p})=\int_{t_s^i}^{t_e^i} \mathbf{c}(\mathbf{T}(t), \mathbf{p}) dt,
\end{equation}
where $\mathbf{T}(t)$ is the camera pose at time $t$, $\mathbf{c}(\mathbf{T}, \mathbf{p})$ is HDR color for camera pose $\mathbf{T}$ at pixel location $\mathbf{p}$ defined in \cref{eq:rendering_nerf,eq:rendering_gs}.
Motion blur occurs if $\mathbf{T}(t)$ changes during exposure.
We numerically approximate the integral with quadrature:
\begin{equation}
    C^i_{\text{HDR}}(\mathbf{p}) = \Delta t^i\cdot\frac{1}{N_{\text{cam}}}\sum_{j=1}^{N_{\text{cam}}}\mathbf{c}\left(\mathbf{T}\left(t_s^i + \frac{j - 1}{N_{\text{cam}} - 1} (t_e^i - t_s^i)\right), \mathbf{p}\right),
\end{equation}
where $N_{\text{cam}}$ is the number of virtual cameras to approximate the continuous integral with discrete summation and $\Delta t^i = t_e^i - t_s^i$ is exposure time. Throughout, we empirically set $N_{\text{cam}}=5$ for experiments.
If we assume that the camera velocity is constant during the short exposure time, the camera poses within time range $t\in[t_s^i, t_e^i]$ can be obtained by linearly interpolating between the start pose $\mathbf{T}(t_s^i)$ and the end pose $\mathbf{T}(t_e^i)$. We interpolate the camera pose at time $t$, $\mathbf{T}(t)$, by decomposing it into rotation $\mathbf{R}(t)$ and translation $\mathbf{t}(t)$:
\begin{equation}
    \mathbf{R}(t)=\text{Slerp}{\left(\mathbf{R}(t_s^i),\mathbf{R}(t_e^i),\frac{t - t_s^i}{t_e^i-t_s^i}\right)}, \quad \mathbf{t}(t)=\text{Lerp}{\left(\mathbf{t}(t_s^i),\mathbf{t}(t_e^i),\frac{t - t_s^i}{t_e^i-t_s^i}\right)},
\end{equation}
where \text{Slerp} stands for spherical linear interpolation, \text{Lerp} is linear interpolation.

Then, the final observed color value of the pixel $\mathbf{p}$ is obtained by applying tone mapping operator $\Psi^i$:
\begin{equation}
    C^i_{\text{LDR}}(\mathbf{p})=\Psi^i(C^i_{\text{HDR}}(\mathbf{p})).
\end{equation}
The tone mapping operator $\Psi^i$ clips the HDR color to low dynamic range (LDR) and maps from linear color space to non-linear color space.
Specifically, the tone mapping operation is composed of white balancing $\text{WB}^i$ and camera response function $\text{CRF}^i$ with dynamic range clipping:
\begin{equation}
    \Psi^i(\Delta t^i \cdot \mathbf{c}) = \text{CRF}^i(\text{WB}^i(\Delta t^i\cdot\mathbf{c})).
\end{equation}
White balance function $\text{WB}^i$ is an element-wise product to each color channel:
\begin{equation}
    \text{WB}^i(\mathbf{c})=\begin{bmatrix}wb_r^i&wb_g^i&wb_b^i\end{bmatrix}^T \odot \begin{bmatrix} c_r&c_g&c_b \end{bmatrix}^T.
\end{equation}
We parameterize non-linear $\text{CRF}^i$ with uniformly sampled 256-dimensional grid $g^i$ between $[0, 1]$ for each color channel.
CRF should satisfy the two properties: (1) monotonically increasing function and (2) $\text{CRF}(0)=0$ and $\text{CRF}(1)=1$~\cite{10.1145/258734.258884}.
We further adjust CRF to be physically plausible following~\cite{liu2020single}.
We shift the derivatives by the smallest negative derivative and normalize CRF to satisfy the two properties.
We employ differentiable grid sampling~\cite{spatial_transformer} to query our CRF value.
Instead of hard clipping the dynamic range between $[0,1]$, we use a leaky clipping function with CRF to backpropagate the gradient:
\begin{equation}
    \text{CRF}_{\text{leaky}}(c) = 
    \begin{cases}
        \alpha c &c<0\\
        \text{CRF}(c) & 0\leq c\leq 1\\
        -\frac{\alpha}{\sqrt{c}} + \alpha + 1 & 1<c
    \end{cases}
    ,
\end{equation}
where $\alpha$ is a constant parameter. Throughout, we set $\alpha = 0.01$ for experiments.\\
To summarize, our image formation process can transform the HDR radiances in our map representation into LDR camera pixel values captured with continuous exposure.
The learnable parameters are the HDR radiance map, start and end poses $\mathbf{T}(t_s), \mathbf{T}(t_e)$, exposure time $\Delta t$, white balance parameters WB, and control points for CRF $g$.
Since all modules are fully differentiable, we can optimize the parameters with a simple gradient-based optimization method by minimizing the objective function with proper regularization, as described in~\cref{subsec:mapping_and_tracking}.

\subsection{Tracking and Mapping}
\label{subsec:mapping_and_tracking}

 We optimize camera trajectories during exposure time and reconstruct a sharp HDR 3D map using the rendering loss defined by our image formation process. To optimize the camera trajectory during exposure time, we additionally propose a trajectory loss and a camera trajectory initialization method. Overall loss is sum of an image rendering loss, a depth rendering loss, and a trajectory loss:
\begin{equation}
    \mathcal{L} = \lambda_{\text{img}} \mathcal{L}_{\text{img}}+\lambda_{\text{depth}} \mathcal{L}_{\text{depth}} + \lambda_{\text{traj}}\mathcal{L}_{\text{traj}}.
\end{equation}

\subsubsection{Image Rendering Loss}
We jointly optimize the in-camera parameters and the map representation within the SLAM's tracking and mapping pipeline by applying an image rendering loss function as follows:
\begin{equation}
    \mathcal{L}_{\text{img}}=\sum_{i=1}^N \sum_{\mathbf{p}} |C_{\text{LDR}}^i(\mathbf{p})-\hat{C}^i(\mathbf{p})|,
\end{equation}
where $C_{\text{LDR}}^i(\mathbf{p})$ is a rendered color output from our image formation process and $\hat{C}^i(\mathbf{p})$ is the observed color at pixel location $\mathbf{p}$ of frame $i$.
Even if the input images are degraded, the proposed rendering loss accounts for the physical degradation process and can reconstruct a sharp HDR radiance map.

\subsubsection{Depth Rendering Loss} 
We constrain the depth camera pose to be aligned with the color camera's trajectory during exposure time.
Specifically, we assign a pose with a minimum depth error as the depth camera pose and apply depth rendering loss for the selected pose as follows:
\begin{equation}
    \mathcal{L}_{\text{depth}}=\sum_{i=1}^N \sum_{\mathbf{p}} |d(\mathbf{T}(t_{\text{d}}^i), \mathbf{p})-\hat{d}^i(\mathbf{p})|,
    \text{where }t_{\text{d}}^i = \underset{t\in[t_s^i,t_e^i]}{\arg\min}\lvert d(\mathbf{T}(t),\mathbf{p})-\hat{d}^i(\mathbf{p})\rvert,
\end{equation}
where $d(\mathbf{T}(t_{\text{d}}^i), \mathbf{p})$ is a rendered depth from the depth camera pose $\mathbf{T}(t_{\text{d}}^i)$ and $\hat{d}^i(\mathbf{p})$ is the ground-truth depth at pixel location $\mathbf{p}$ of frame $i$.
Note that the depth rendering loss affects the color camera's trajectory since the depth camera pose is interpolated from $\textbf{T}(t_s^i)$ and $\textbf{T}(t_e^i)$. 
Depth rendering loss ensures that at least one pose along the camera poses during the exposure time outputs accurate depth rendering.

Unlike the integration process for color pixels, we optimize a single pose for a depth camera, assuming that the depth information is captured at a single moment within the time window.
Although the depth camera measures some inaccurate depth values in fast-moving scenarios, most of the noises are filtered as invalid pixels by sensor manufacturers~\cite{sarbolandi2015kinect}.
Depth sensors usually output invalid pixels on object boundaries and black objects, and most RGBD-SLAM methods ignore those regions.

\subsubsection{Trajectory Regularization}
We propose a trajectory loss that regularizes the camera trajectory during the exposure time.
We design the trajectory loss function with two insights.
First, the camera poses during the exposure time should be aligned with the global trajectory.
As SLAM progressively optimizes the camera poses, the global trajectory can be obtained from previous localization results without additional cost and it gives meaningful information to the temporal sensor movement.
We estimate the global trajectory by connecting the pose of the midpoint during each exposure time in the previous frames.
Second, the size of the motion blur kernel is determined by the temporal velocity and the exposure time.
Namely, the longer the exposure time $\Delta t^i$ and the faster the temporal velocity, the further the distance between the start and end camera poses.

Our trajectory loss regularizes the temporal camera trajectory during the exposure time to be aligned with the global trajectory and the length of it to be proportional to the exposure time and temporal velocity, assuming piecewise linear velocity:

\begin{equation}
    \mathcal{L}_{\text{traj}} = \mathcal{L}_{\text{traj}}^{\mathbf{t}} + \mathcal{L}_{\text{traj}}^{\mathbf{R}},
\end{equation}

\begin{equation}
\small
    \mathcal{L}_{\text{traj}}^{\mathbf{t}} = \left\lVert\mathbf{t}(t_e^{i-1}) - \text{Lerp}(\mathbf{t}^{i-1}, \mathbf{t}^{i}, a \Delta t^{i-1})\right\rVert_2^2 + \left\lVert\mathbf{t}(t_s^{i}) - \text{Lerp}(\mathbf{t}^{i-1}, \mathbf{t}^{i}, 1-a \Delta t^{i})\right\rVert_2^2,
\end{equation}

\begin{equation}
\small
    \mathcal{L}_{\text{traj}}^{\mathbf{R}} = \left\lVert\mathbf{R}(t_e^{i-1}) - \text{Slerp}(\mathbf{R}^{i-1}, \mathbf{R}^{i}, a\Delta t^{i-1})\right\rVert_2^2 + \left\lVert\mathbf{R}(t_s^{i}) - \text{Slerp}(\mathbf{R}^{i-1}, \mathbf{R}^{i}, 1-a \Delta t^{i})\right\rVert_2^2,
\end{equation}
where $\mathbf{t}$ and $\mathbf{R}$ are the translation and rotation vector of camera pose $\mathbf{T}$, and $a$ is an unknown global scale parameter that is related to the input frame rate.
$\mathbf{t}^i$ and $\mathbf{R}^i$ are the center pose between the start and end pose of frame $i$.
The scale parameter $a$ is also jointly optimized within the tracking process.
We further describe the relation between our scale parameter $a$ and the input frame rate in the supplementary material.

We also propose an initialization strategy for robust optimization exploiting the global trajectory.
With a constant velocity assumption, we initialize the $i+1$th frame's camera poses $\mathbf{T}(t^{i+1}_s)$ and $\mathbf{T}(t^{i+1}_e)$ by extrapolating estimated camera poses of $i$ and $i-1$th frames.
We initialize the start and end poses to be separated with a small predefined distance along the global trajectory.

\section{Experiments}
\label{sec:experiments}
We demonstrate that \islam~can reconstruct a sharp HDR radiance map from casually captured videos with various input modalities.
We describe our experimental setup in challenging datasets in \cref{subsec:experimental_setup}.
We show that \islam~enhances state-of-the-art dense visual SLAM methods in \cref{subsec:results}.
We conduct ablation studies and runtime analyses in \cref{subsec:ablation} and \cref{subsec:runtime}, respectively.

\subsection{Experimental Setup}
\label{subsec:experimental_setup}
\paragraph{Baselines}
\islam~serves as a versatile module that can be attached to dense visual SLAM methods to improve its performance.
We apply our approach to the state-of-the-art RGB-SLAM method, NeRF-SLAM~\cite{rosinol2023nerf}, which employs NeRF as a map representation and uses DROID-SLAM~\cite{teed2021droid} as a tracking backbone.
We re-implemented NeRF-SLAM with torch-ngp~\cite{torch-ngp} and notate as $\text{NeRF-SLAM}^\dagger$. 
$\text{NeRF-SLAM}^\dagger$ uses same loss functions of NeRF-SLAM.
We additionally test \islam~with an RGBD-SLAM approach to tackle challenging sequences in ScanNet~\cite{dai2017scannet} where robust learning-based SLAM method, DROID-SLAM, often fails without using additional depth channel.
We employ a 3DGS-based RGBD-SLAM method, SplaTAM~\cite{keetha2023splatam}, to test \islam~in RGBD inputs.

\paragraph{Datasets}
Most of the existing synthetic datasets for SLAM evaluation assume the ideal capturing setup, and do not exhibit any camera motion blur or dynamic appearance changes.
We therefore propose a new dataset incorporating these effects.
The new dataset contains realistic image degradation by simulating motion blur and auto exposure.
We render the images and depth information with Cycles path tracer~\cite{blender}.
Also, we test \islam~on challenging real-world datasets.
We use TUM-RGBD~\cite{sturm2012benchmark} and ScanNet~\cite{dai2017scannet} dataset for evaluating RGB and RGBD scenarios, respectively.

\paragraph{Evaluation}
We report the average values of three runs with different random seeds for all the quantitative evaluations except the ablation study and the runtime analysis.
We measure PSNR, SSIM~\cite{wang2004image}, and LPIPS~\cite{zhang2018unreasonable} between rendered images from the reconstructed map and sharp ground-truth images which can be obtained from our synthetic dataset.
Since there are no ground-truth sharp images in the real-world dataset, we evaluate the view synthesis performance only for the sharp frames that are manually annotated.
Also, we run test-time optimization~\cite{lin2021barf} to factor out pose errors in measuring the rendering quality of RGB-SLAMs.
We report ATE RMSE~\cite{sturm2012benchmark} for tracking performance evaluation.
We use the center of camera poses $\mathbf{T}(t_s^i)$ and $\mathbf{T}(t_e^i)$ to evaluate our method and use scale-aligned ground-truth trajectory for RGB-SLAM to handle the scale ambiguity.
Further details can be found in the supplementary material.

\subsection{Experimental Results}
\label{subsec:results}

\begin{table}[t]
    \caption{Rendering quality comparison against the RGB-SLAM baseline on TUM-RGBD~\cite{sturm2012benchmark} and synthetic dataset. \islam~represents our RGB-SLAM model which is incorporated into our re-implementation of NeRF-SLAM~\cite{rosinol2023nerf}, $\text{NeRF-SLAM}^\dagger$.}
    \label{tab:quantitative_rgb_rendering}
    \centering
    \resizebox{0.95\textwidth}{!}{
    \begin{tabular}{lwc{6em} wc{4.5em}wc{4.5em}wc{4.5em}wc{3.0em}wc{3.0em}wc{3.0em}wc{3.0em}wc{3.0em}}
        \toprule
        \multirow{2}{*}{Methods} & \multirow{2}{*}{Metrics} & \multicolumn{3}{c}{TUM-RGBD~\cite{sturm2012benchmark}} & \multicolumn{5}{c}{Synthetic} \\
         & & \texttt{fr1/desk} & \texttt{fr2/xyz} & \texttt{fr3/office} & \texttt{SP} & \texttt{LOU} & \texttt{IF0} & \texttt{IF1} & \texttt{IF2} \\
        \midrule
        \multirow{4}{*}{$\text{NeRF-SLAM}^\dagger$~\cite{rosinol2023nerf}} & PSNR & 25.97 & 29.97 & 24.72 & 28.64 & 25.43 & 30.20 & 26.09 & 26.70 \\
         & SSIM & 0.825 & 0.900 & 0.727 & 0.810 & 0.832 & 0.867 & 0.789 & 0.842 \\
         & LPIPS & 0.222 & 0.093 & 0.366 & 0.328 & 0.323 & 0.327 & 0.302 & 0.270 \\
         & Depth L1 & 10.97 & \textbf{17.83} & 30.92 & 40.68 & 55.80 & 20.72 & 5.02 & 19.76 \\
         \midrule
        \multirow{4}{*}{\islam} & PSNR & \textbf{27.23} & \textbf{32.06} & \textbf{28.91} & \textbf{28.99} & \textbf{27.59} & \textbf{32.33} & \textbf{30.16} & \textbf{28.89} \\
         & SSIM & \textbf{0.835} & \textbf{0.916} & \textbf{0.833} & \textbf{0.827} & \textbf{0.875} & \textbf{0.902} & \textbf{0.898} & \textbf{0.887} \\
         & LPIPS & \textbf{0.186} & \textbf{0.074} & \textbf{0.193} & \textbf{0.284} & \textbf{0.260} & \textbf{0.286} & \textbf{0.211} & \textbf{0.241} \\
         & Depth L1 & \textbf{9.04} & 17.94 & \textbf{17.60} & \textbf{20.33} & \textbf{41.92} & \textbf{20.48} & \textbf{3.55} & \textbf{17.28} \\
         \bottomrule
    \end{tabular}
    }
\end{table}

\begin{table}[t]
    \caption{Rendering quality comparison against the RGBD-SLAM baseline on ScanNet~\cite{dai2017scannet} and synthetic dataset. \islam~in this table represents our RGBD-SLAM model which is incorporated into SplaTAM~\cite{keetha2023splatam}.  
    }
    \label{tab:rgbd_rendering}
    \centering
    \resizebox{0.9\textwidth}{!}{
    \begin{tabular}{lcccccccccc}
        \toprule
        \multirow{2}{*}{Methods} & \multirow{2}{*}{Metrics} & \multicolumn{4}{c}{ScanNet~\cite{dai2017scannet}} & \multicolumn{5}{c}{Synthetic}\\
        & & \texttt{0024-01} & \texttt{0031-00} & \texttt{0736-00} & \texttt{0785-00} & \texttt{SP} & \texttt{LOU} & \texttt{IF0} & \texttt{IF1} & \texttt{IF2} \\
         \midrule
        \multirow{3}{*}{SplaTAM~\cite{keetha2023splatam}} & PSNR & 21.60 & 24.64 & \textbf{24.50} & 19.63 & 21.38 & 19.78 & \textbf{24.22} & 22.36 & 23.82 \\
         & SSIM  & \textbf{0.786} & 0.773 & \textbf{0.847} & 0.719 & 0.820 & \textbf{0.790} & \textbf{0.855} & 0.766 & 0.824 \\
         & LPIPS  & 0.236 & 0.275 & 0.182 & 0.340 & 0.232 & 0.246 & \textbf{0.227} & 0.281 & 0.230 \\
         \midrule
        \multirow{3}{*}{\islam} & PSNR  & \textbf{23.39} & \textbf{26.89} & 24.07 & \textbf{26.40} & \textbf{26.18} & \textbf{21.98} & 23.88 & \textbf{23.72} & \textbf{24.07} \\
         & SSIM & 0.780 & \textbf{0.796} & 0.828 & \textbf{0.762} & \textbf{0.842} & 0.770 & 0.798 & \textbf{0.796} & \textbf{0.826} \\
         & LPIPS & \textbf{0.180} & \textbf{0.236} & \textbf{0.175} & \textbf{0.238} & \textbf{0.193} &\textbf{0.231} & 0.263 & \textbf{0.233} & \textbf{0.205} \\
         \bottomrule
    \end{tabular}
    }
\end{table}

\begin{table}[t]
    \caption{We evaluate the tracking performance of RGB-SLAM methods on TUM-RGBD~\cite{sturm2012benchmark} and synthetic dataset in terms of ATE-RMSE (cm).}
    \label{tab:rgb_tracking}
    \centering
    \resizebox{0.7\textwidth}{!}{
    \begin{tabular}{lwc{4.5em}wc{4.5em}wc{4.5em}wc{2.0em}wc{2.0em}wc{2.0em}wc{2.0em}wc{2.0em}}
        \toprule
        \multirow{2}{*}{Methods} & \multicolumn{3}{c}{TUM-RGBD~\cite{sturm2012benchmark}} & \multicolumn{5}{c}{Synthetic}\\ 
         & \texttt{fr1/desk} & \texttt{fr2/xyz} & \texttt{fr3/office} & \texttt{SP} & \texttt{LOU} & \texttt{IF0} & \texttt{IF1} & \texttt{IF2} \\
        \midrule
        $\text{NeRF-SLAM}^\dagger$~\cite{rosinol2023nerf} &  2.08 & 0.41 & 7.13 & 3.97 & \textbf{3.20} & 3.38 & 1.14 & 0.65 \\
         \midrule
        \islam & \textbf{1.64} & \textbf{0.26} & \textbf{1.95} & \textbf{1.50} & 3.23 & \textbf{1.59} & \textbf{0.74} & \textbf{0.33} \\
         \bottomrule
    \end{tabular}
    }
\end{table}

\begin{table}[t]
    \caption{Tracking accuracy comparison against the RGBD-SLAM baseline on ScanNet~\cite{dai2017scannet} and synthetic dataset. ATE-RMSE~(cm) is measured as an evaluation metric.}
    \label{tab:rgbd_tracking}
    \centering
    \resizebox{0.7\textwidth}{!}{
    \begin{tabular}{lwc{3.5em}wc{3.5em}wc{3.5em}wc{3.5em}wc{2.0em}wc{2.0em}wc{2.0em}wc{2.0em}wc{2.0em}}
        \toprule
        \multirow{2}{*}{Methods} & \multicolumn{4}{c}{ScanNet~\cite{dai2017scannet}} & \multicolumn{5}{c}{Synthetic}\\ 
         & \texttt{0024-01} & \texttt{0031-00} & \texttt{0736-00} & \texttt{0785-00} & \texttt{SP} & \texttt{LOU} & \texttt{IF0} & \texttt{IF1} & \texttt{IF2} \\
        \midrule
        SplaTAM~\cite{keetha2023splatam} & 1.80 & \textbf{3.01} & 1.13 & 5.91 & 1.11 & \textbf{1.50} & 2.04 & \textbf{1.02} & \textbf{0.61} \\
         \midrule
        \islam & \textbf{1.41} & 3.25 & \textbf{1.00} & \textbf{4.59} & \textbf{0.86} & 1.55 & \textbf{1.96} & 1.05 & 0.87 \\
         \bottomrule
    \end{tabular}
    }
\end{table}

\subsubsection{Quantitative Results}
We report the quantitative results of map rendering performance for keyframes of RGB and RGBD datasets in \cref{tab:quantitative_rgb_rendering,tab:rgbd_rendering}, respectively.
We observe that \islam~enhances the rendering quality of $\text{NeRF-SLAM}^\dagger$ across all metrics on the both synthetic and real-world TUM-RGBD~\cite{sturm2012benchmark} datasets.
\islam~also substantially enhances the depth accuracy for RGB-SLAM.
In the RGBD dataset scenario, \islam~also shows superior rendering quality in most scenes when attached to SplaTAM~\cite{keetha2023splatam}.
Especially, \islam~improves the view-synthesis performance of SplaTAM in a large margin in \texttt{SP} of synthetic dataset and \texttt{0785-00} of ScanNet~\cite{dai2017scannet} dataset, whose appearances are changing dynamically within the videos.
It shows that our in-camera model is especially advantageous when combined with the models without view-dependent effects.
Furthermore, we report the camera trajectory error of RGB-SLAM methods in \cref{tab:rgb_tracking}.
The results show that \islam~substantially improves the tracking performance of $\text{NeRF-SLAM}^\dagger$ even if $\text{NeRF-SLAM}^\dagger$ takes accurate initial trajectory from DROID-SLAM as initial camera poses.
It supports our claim that image formation process, when appropriately modeled, can enhance the tracking accuracy of inputs that are casually captured.
In RGBD datasets, however, the trajectory accuracy improvement is subtle since the depth information mostly determines the camera poses as reported in \cref{tab:rgbd_tracking}.

\begin{figure}[!ht]
    \centering
    \begin{subfigure}{0.9\linewidth}
        \begin{tabular}{@{}c@{\,}c@{\,}c@{\,}c@{\,}}
            & \resizebox{0.12\textwidth}{!}{Input Frame} & \resizebox{0.18\textwidth}{!}{$\text{NeRF-SLAM}^\dagger$~\cite{rosinol2023nerf}} & \resizebox{0.09\textwidth}{!}{\islam} \\
            \rotatebox{90}{\quad\;\; \resizebox{0.07\textwidth}{!}{\texttt{Sponza}}} & \includegraphics[width=0.32\linewidth]{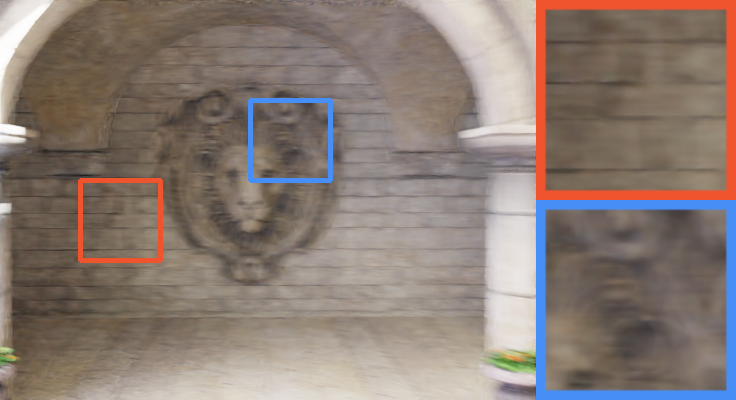}& \includegraphics[width=0.32\linewidth]{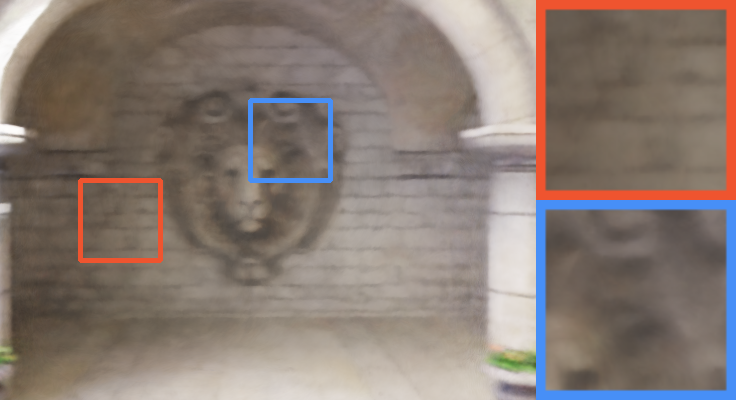} & \includegraphics[width=0.32\linewidth]{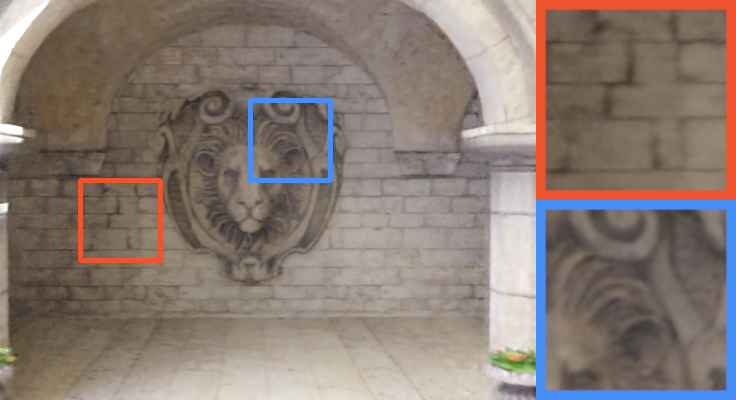}\\
            \rotatebox{90}{\qquad \; \resizebox{0.035\textwidth}{!}{\texttt{LOU}}} & \includegraphics[width=0.32\linewidth]{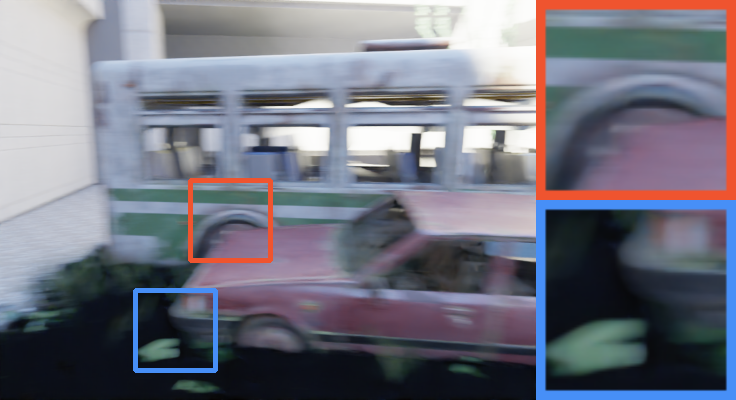}& \includegraphics[width=0.32\linewidth]{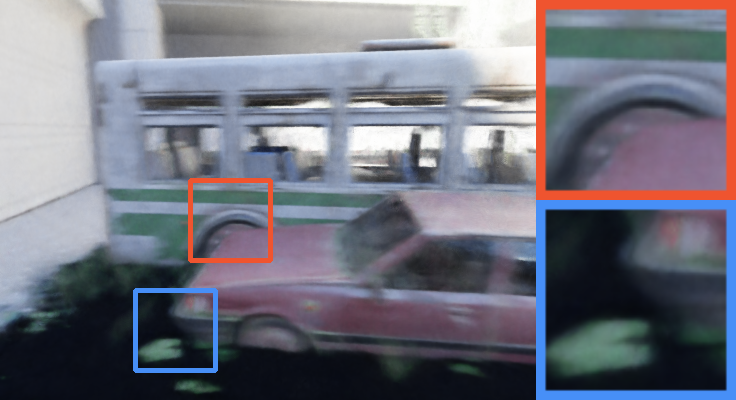} & \includegraphics[width=0.32\linewidth]{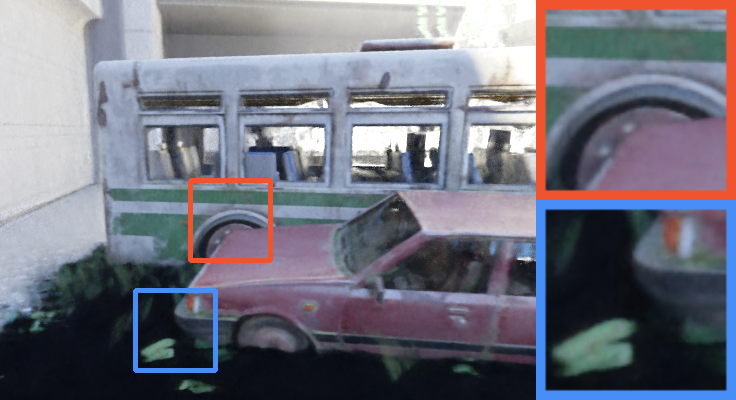}\\
            \rotatebox{90}{\: \resizebox{0.14\textwidth}{!}{\texttt{Italian-flat-0}}} & \includegraphics[width=0.32\linewidth]{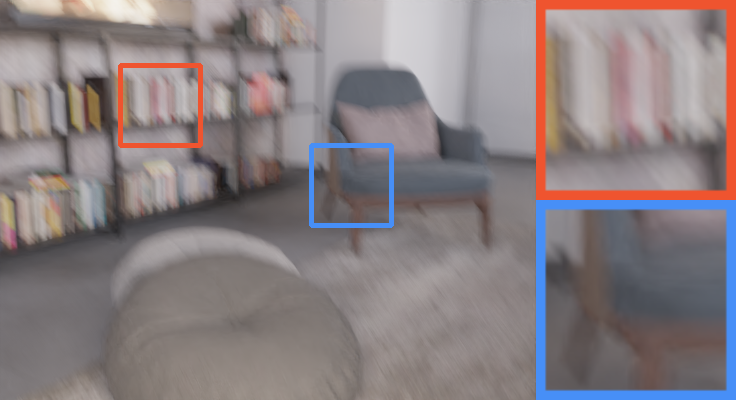}& \includegraphics[width=0.32\linewidth]{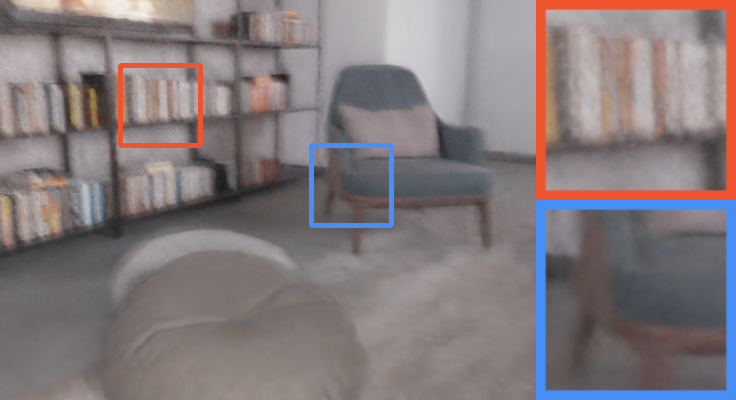} & \includegraphics[width=0.32\linewidth]{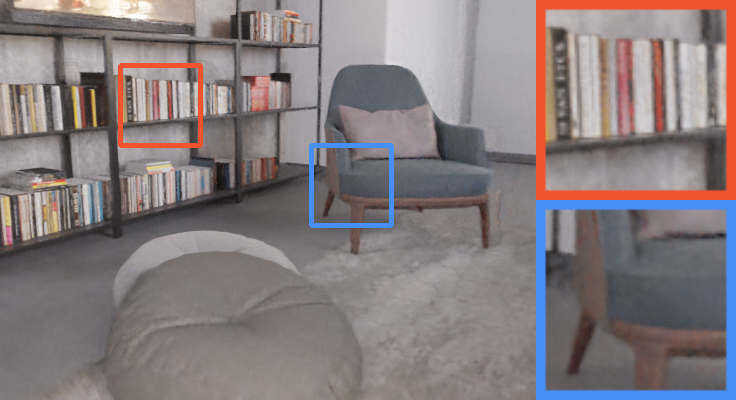}\\
        \end{tabular}
        \caption{Synthetic Dataset}
        \label{fig:qualitative_rgb_synthetic}
    \end{subfigure}\\
    \begin{subfigure}{0.9\linewidth}
        \begin{tabular}{@{}c@{\,}c@{\,}c@{\,}c@{\,}}
            & \resizebox{0.12\textwidth}{!}{Input Frame} & \resizebox{0.18\textwidth}{!}{$\text{NeRF-SLAM}^\dagger$~\cite{rosinol2023nerf}} & \resizebox{0.09\textwidth}{!}{\islam} \\
            \multirow{2}{*}{\rotatebox{90}{\qquad \resizebox{0.1\textwidth}{!}{\texttt{fr1/desk}}}} & \includegraphics[width=0.32\linewidth]{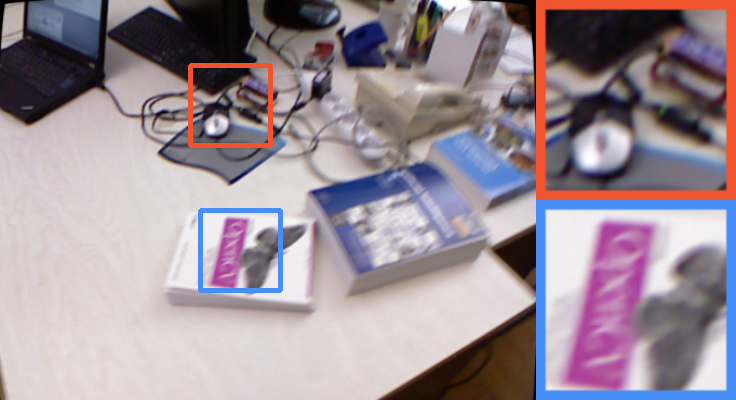}& \includegraphics[width=0.32\linewidth]{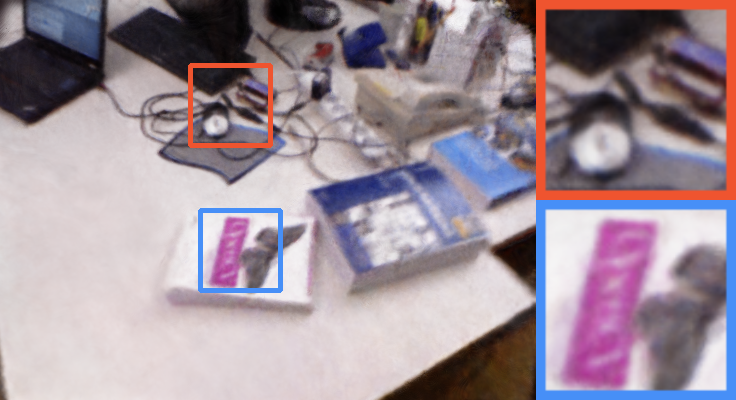} & \includegraphics[width=0.32\linewidth]{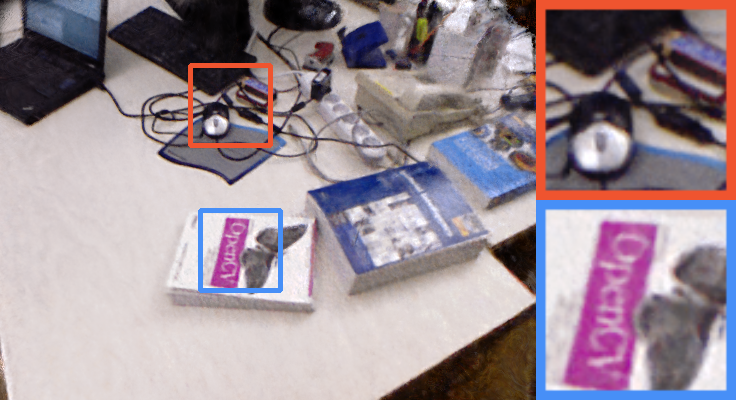}\\
             & \includegraphics[width=0.32\linewidth]{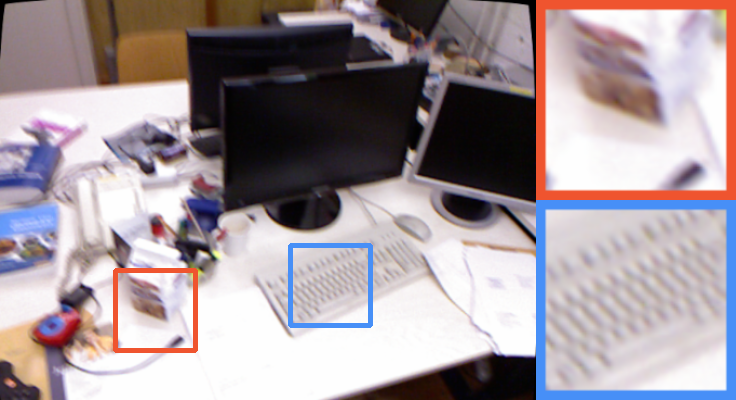}& \includegraphics[width=0.32\linewidth]{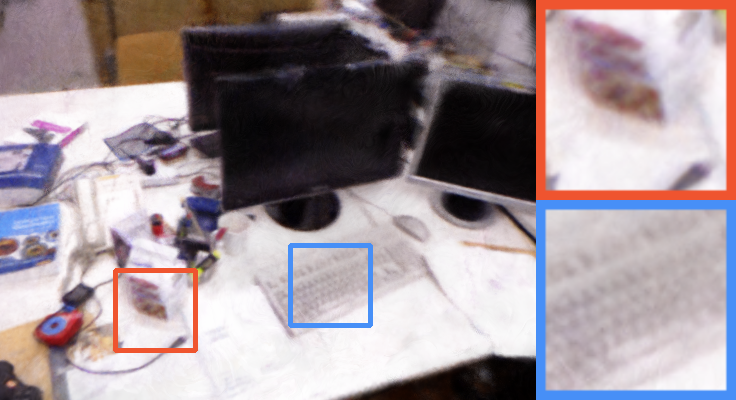} & \includegraphics[width=0.32\linewidth]{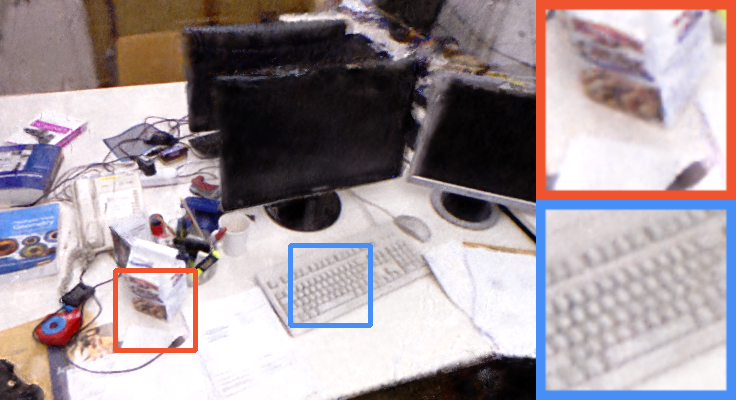}\\
            \multirow{2}{*}{\rotatebox{90}{\quad\; \resizebox{0.12\textwidth}{!}{\texttt{fr3/office}}}} & \includegraphics[width=0.32\linewidth]{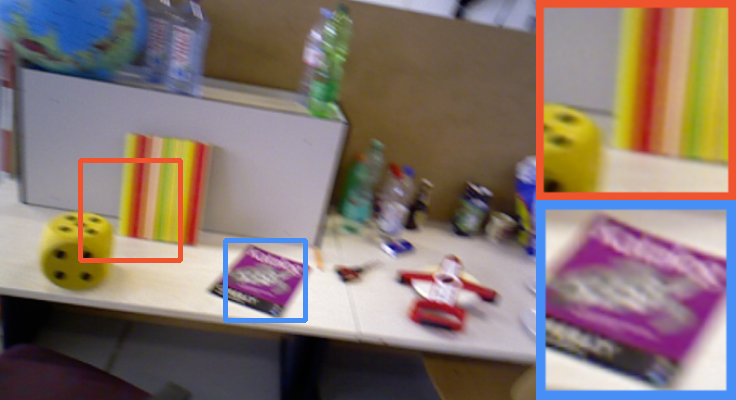}& \includegraphics[width=0.32\linewidth]{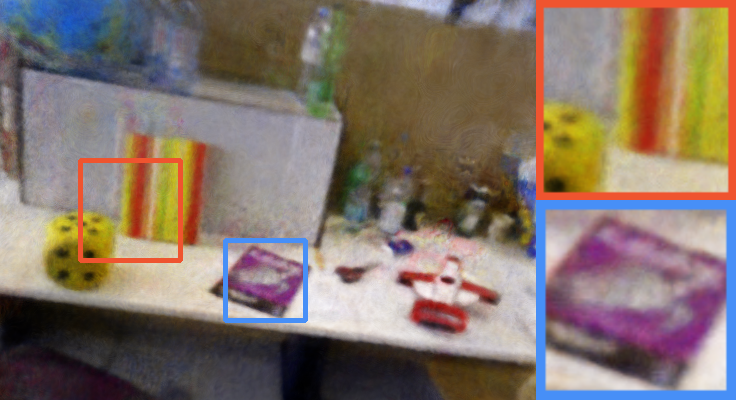} & \includegraphics[width=0.32\linewidth]{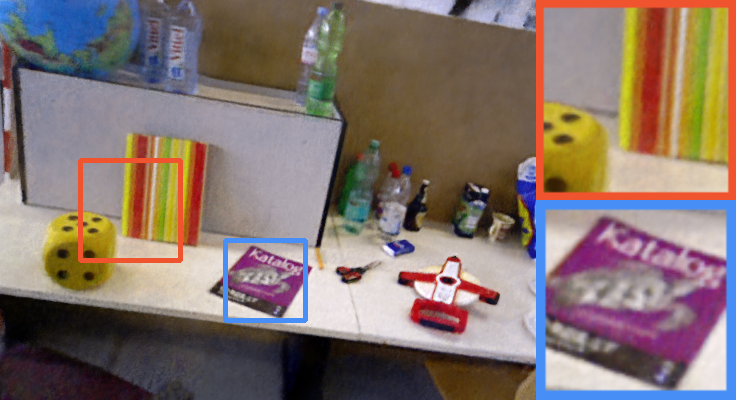}\\
             & \includegraphics[width=0.32\linewidth]{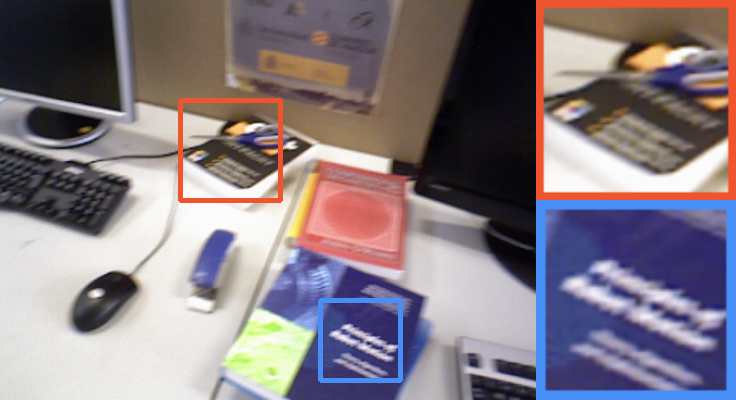}& \includegraphics[width=0.32\linewidth]{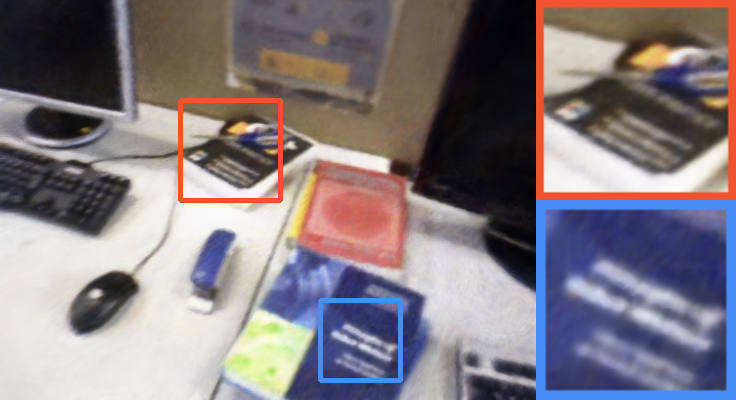} & \includegraphics[width=0.32\linewidth]{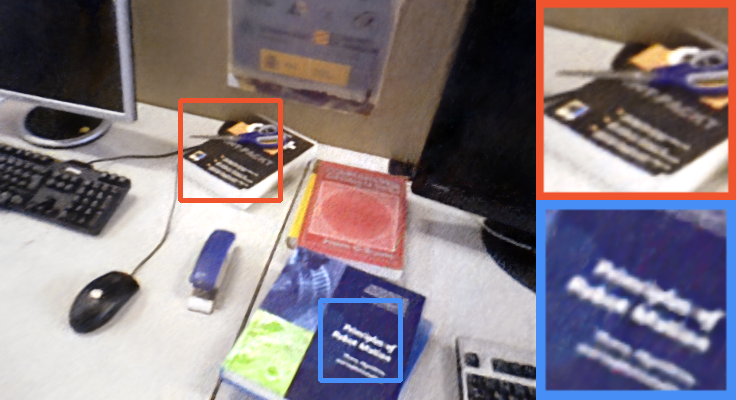}\\
        \end{tabular}
        \caption{TUM-RGBD~\cite{sturm2012benchmark}}
        \label{fig:qualitative_rgb_tum}
    \end{subfigure}
    \caption{Qualitative results on applying \islam~to the RGB-SLAM method.}
    \label{fig:qualitative_rgb}
\end{figure}

\begin{figure}[!ht]
    \centering
    \begin{subfigure}{0.9\linewidth}
        \begin{tabular}{@{}c@{\,}c@{\,}c@{\,}c@{\,}}
            & \resizebox{0.12\textwidth}{!}{Input Frame} & \resizebox{0.13\textwidth}{!}{SplaTAM~\cite{keetha2023splatam}} & \resizebox{0.09\textwidth}{!}{\islam} \\
            \rotatebox{90}{\; \resizebox{0.15\textwidth}{!}{\texttt{Italian-flat-1}}} & \includegraphics[width=0.32\linewidth]{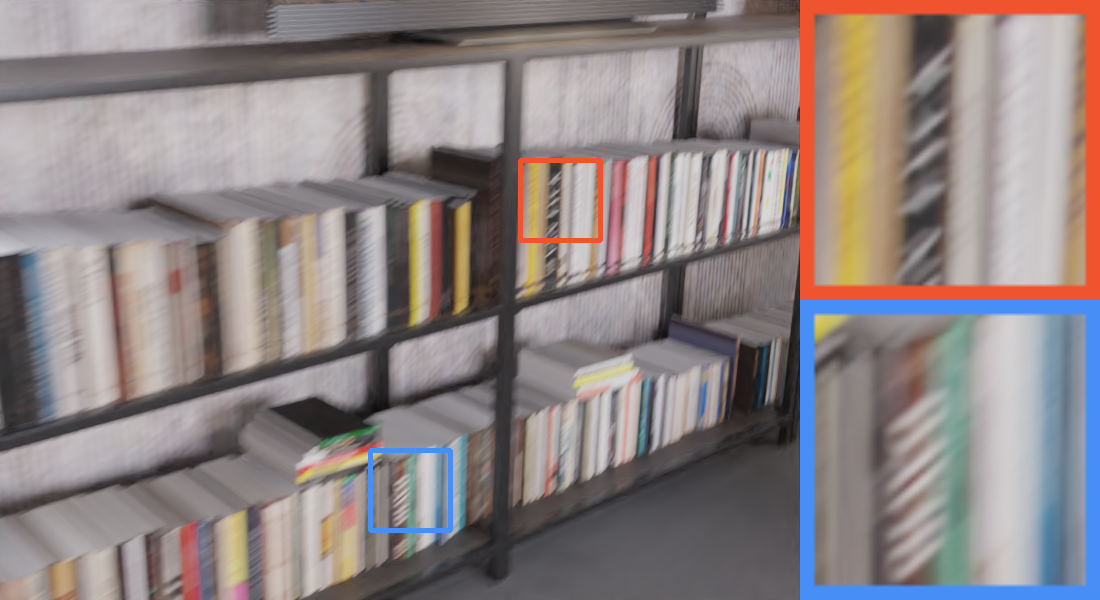}& \includegraphics[width=0.32\linewidth]{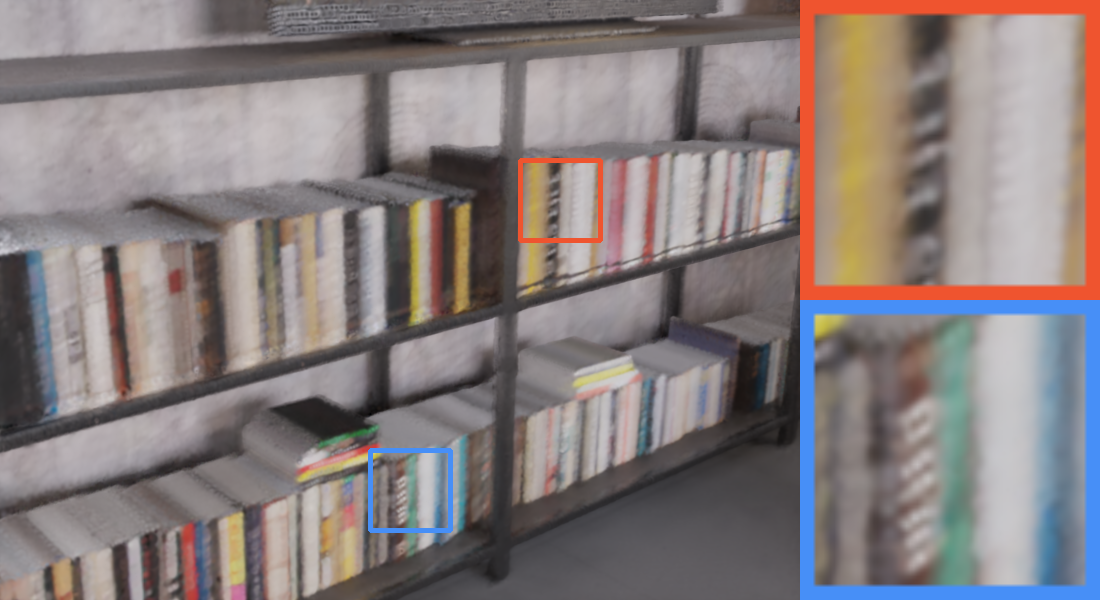} & \includegraphics[width=0.32\linewidth]{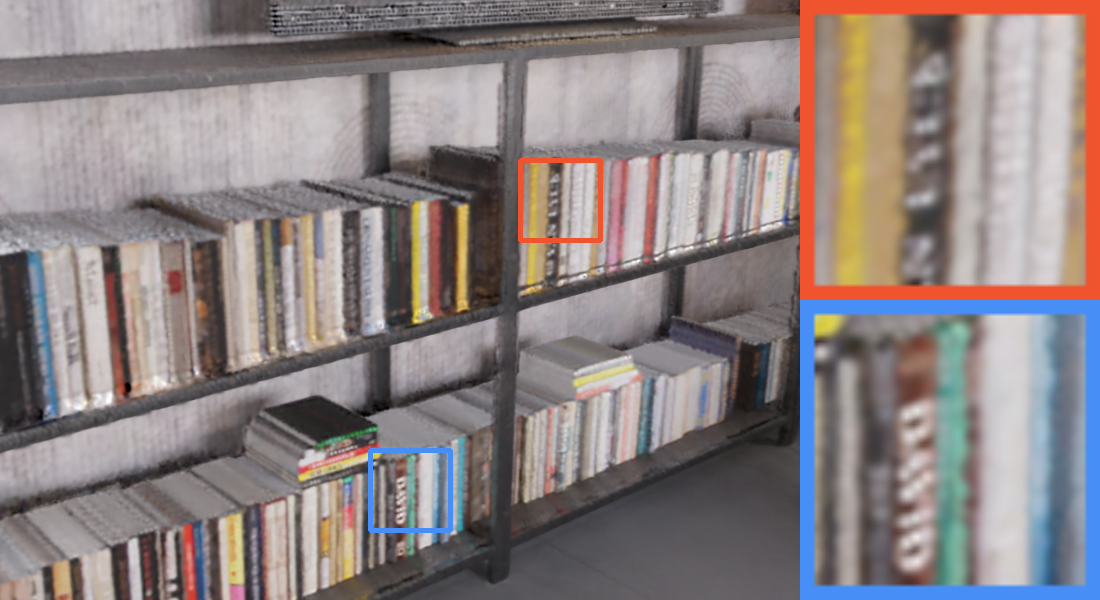}\\
            \rotatebox{90}{\quad\; \resizebox{0.07\textwidth}{!}{\texttt{Sponza}}} & \includegraphics[width=0.32\linewidth]{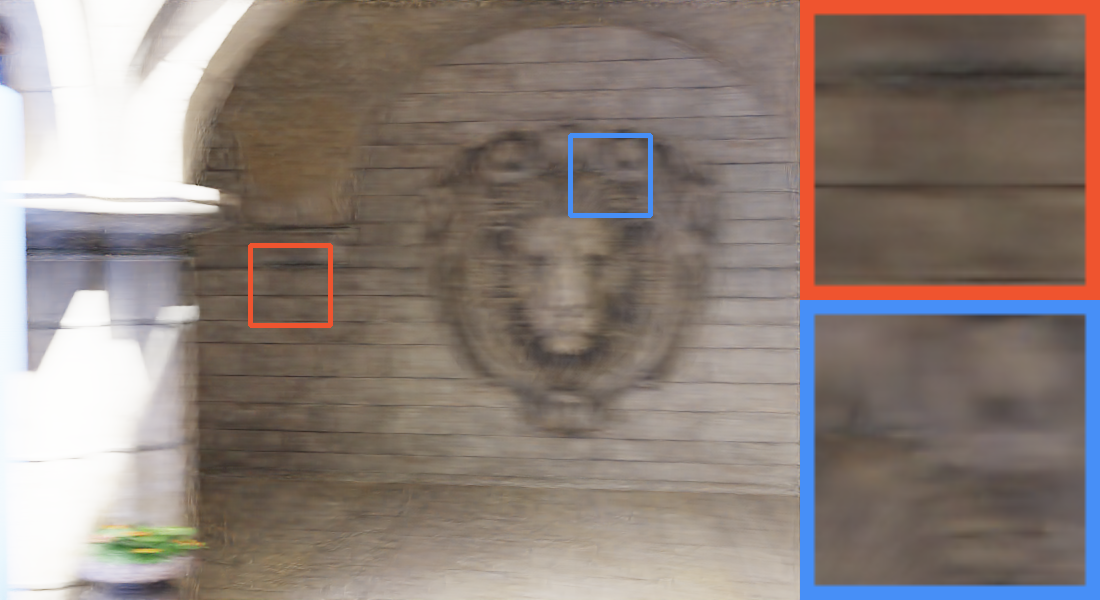}& \includegraphics[width=0.32\linewidth]{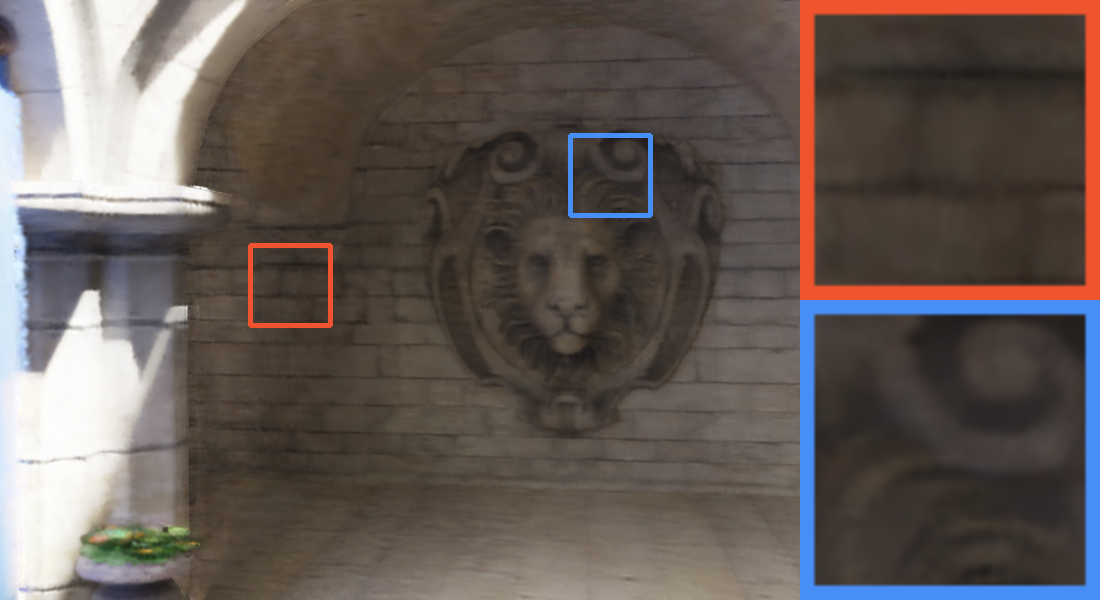} & \includegraphics[width=0.32\linewidth]{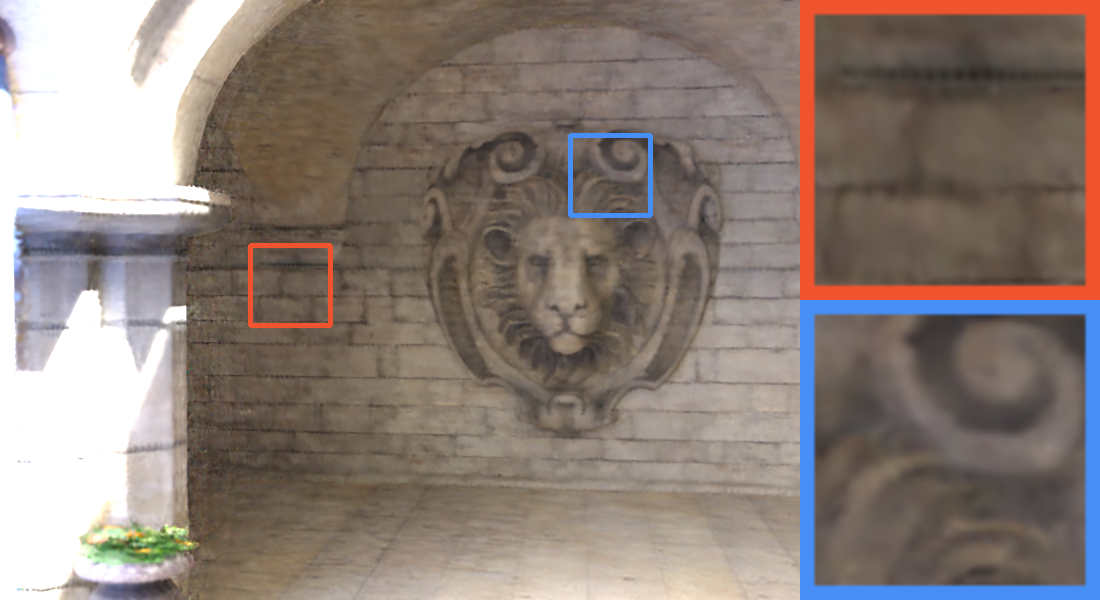}\\
        \end{tabular}
        \caption{Synthetic Dataset}
        \label{fig:qualitative_rgb_synthetic}
    \end{subfigure}\\
    \begin{subfigure}{0.9\linewidth}
        \begin{tabular}{@{}c@{\,}c@{\,}c@{\,}c@{\,}}
            & \resizebox{0.12\textwidth}{!}{Input Frame} & \resizebox{0.13\textwidth}{!}{SplaTAM~\cite{keetha2023splatam}} & \resizebox{0.09\textwidth}{!}{\islam} \\
            \rotatebox{90}{\quad\; \resizebox{0.08\textwidth}{!}{\texttt{0024-01}}} & \includegraphics[width=0.32\linewidth]{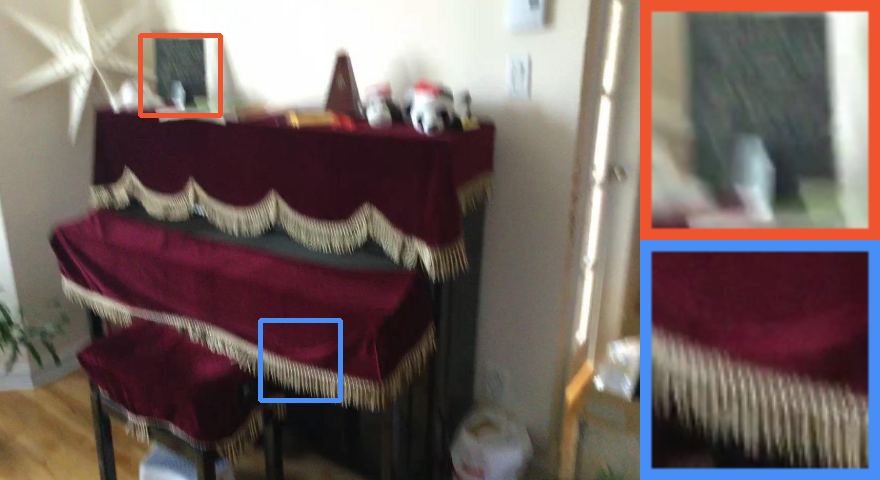}& \includegraphics[width=0.32\linewidth]{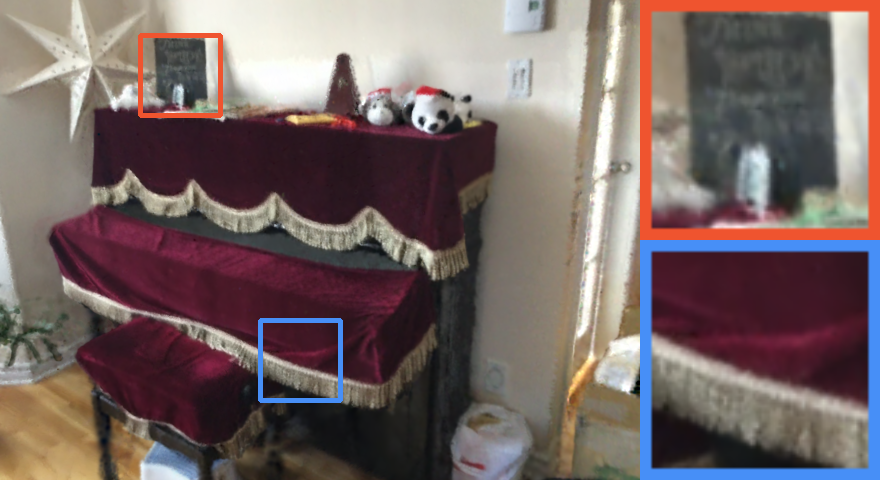} & \includegraphics[width=0.32\linewidth]{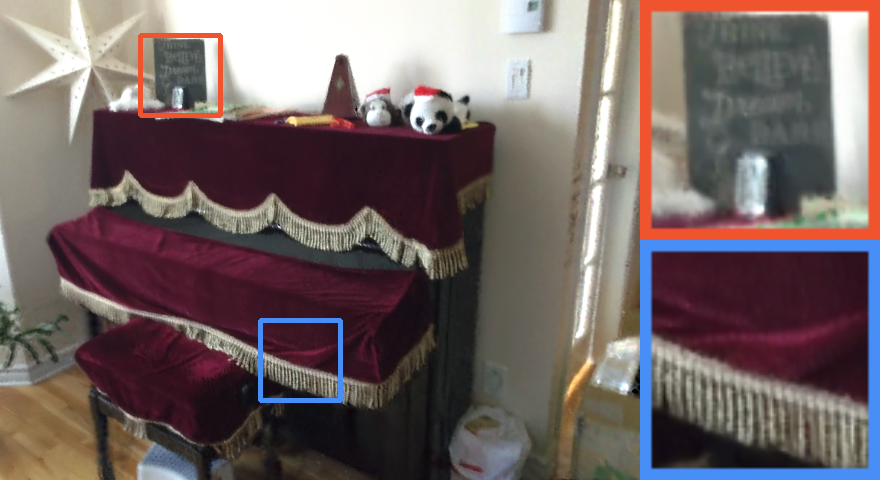}\\
            \rotatebox{90}{\quad\; \resizebox{0.08\textwidth}{!}{\texttt{0736-00}}} & \includegraphics[width=0.32\linewidth]{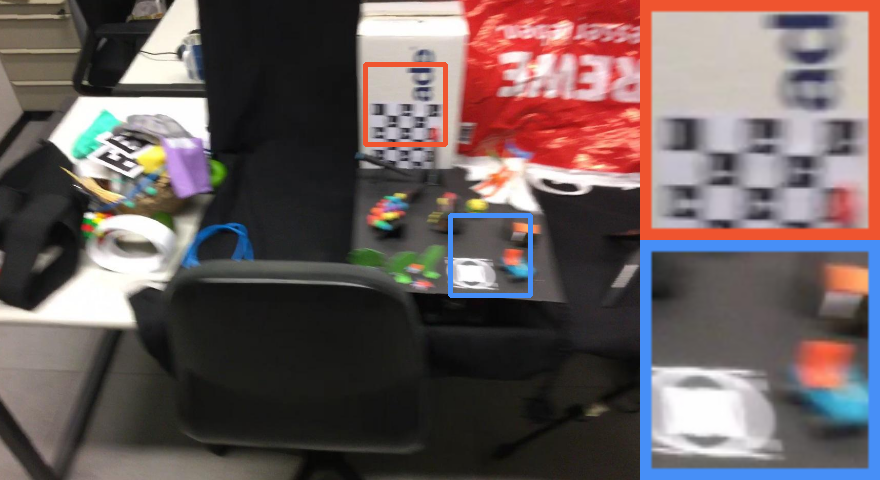}& \includegraphics[width=0.32\linewidth]{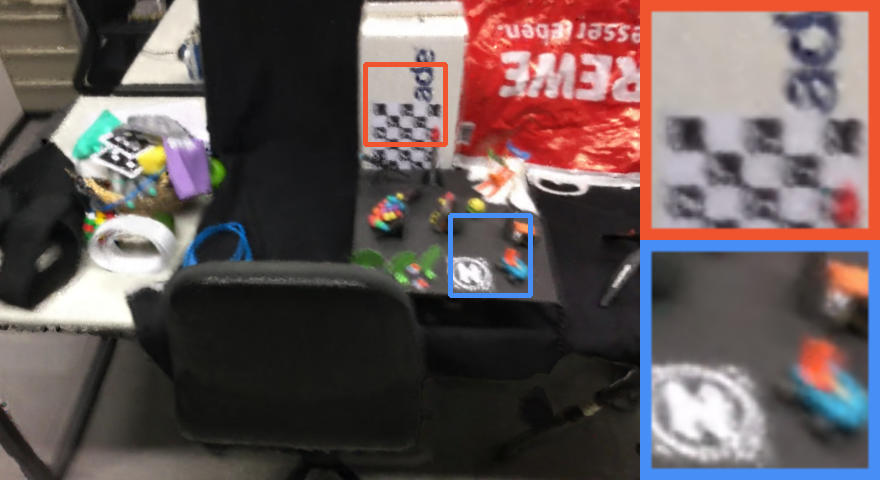} & \includegraphics[width=0.32\linewidth]{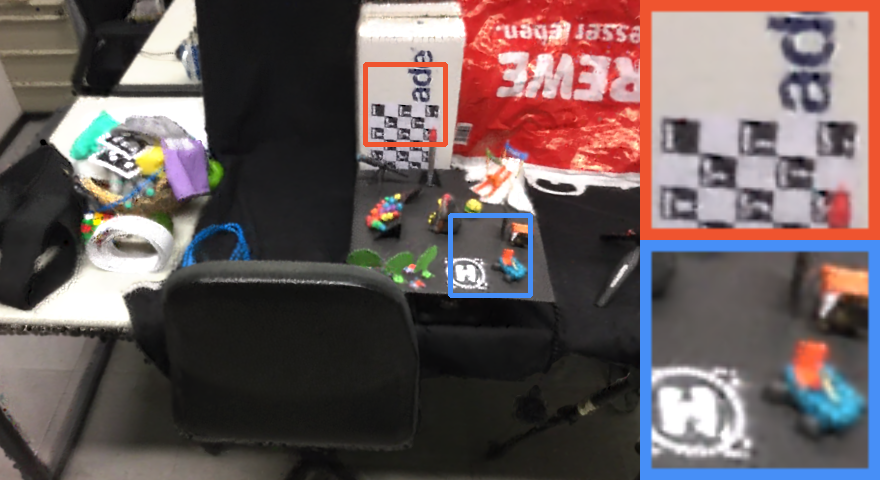}\\
            \multirow{2}{*}{\rotatebox{90}{\quad\; \resizebox{0.08\textwidth}{!}{\texttt{0785-00}}}} & \includegraphics[width=0.32\linewidth]{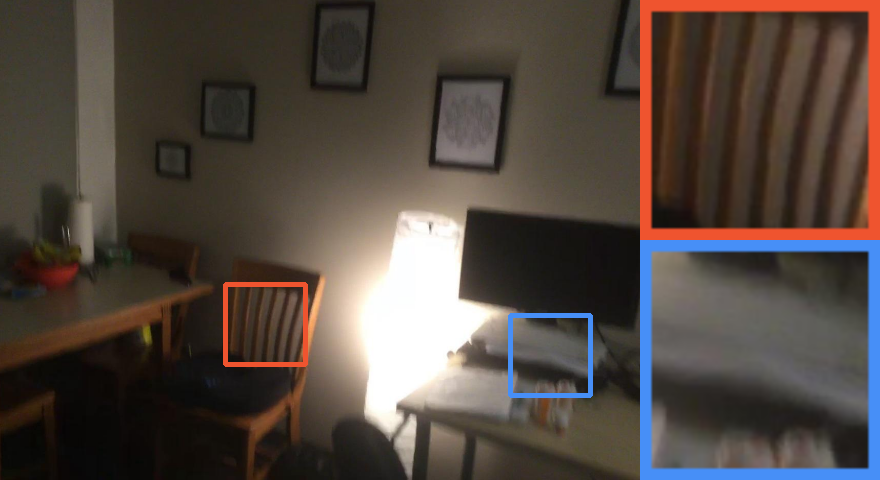}& \includegraphics[width=0.32\linewidth]{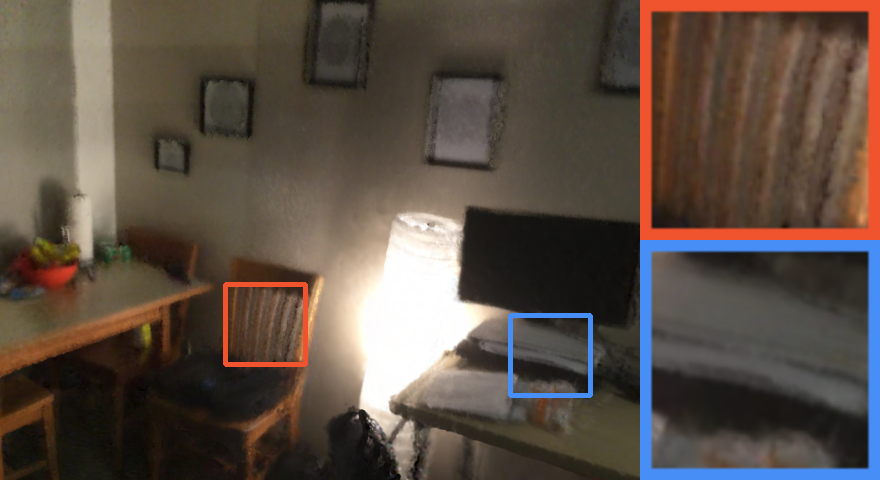} & \includegraphics[width=0.32\linewidth]{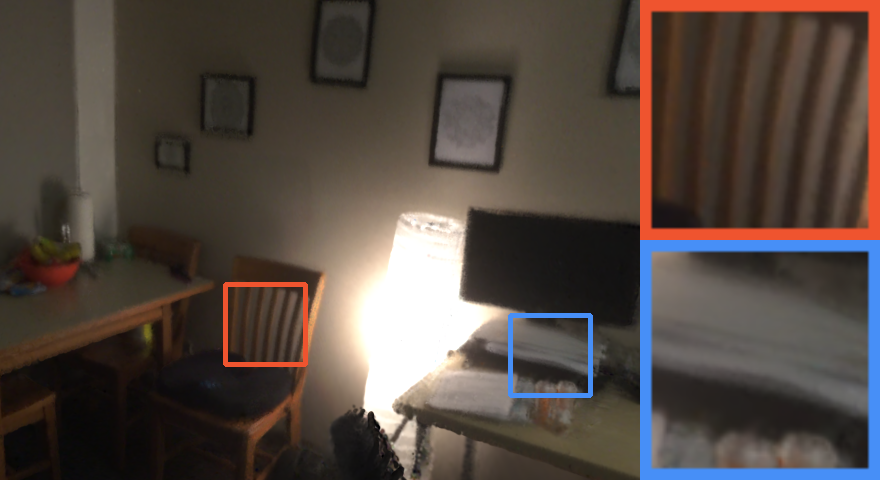}\\
            & \includegraphics[width=0.32\linewidth]{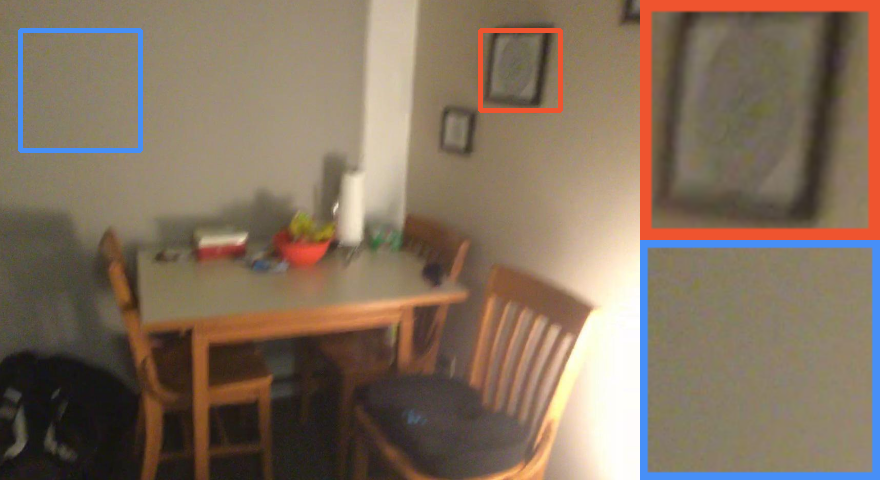}& \includegraphics[width=0.32\linewidth]{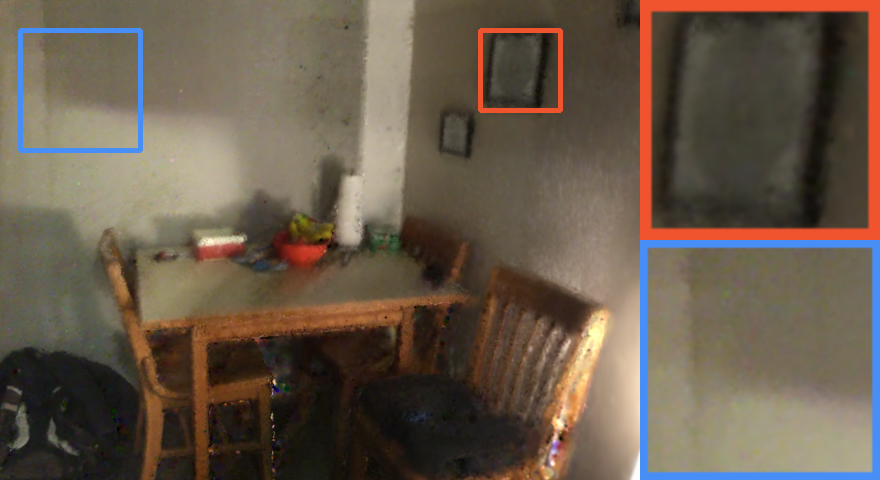} & \includegraphics[width=0.32\linewidth]{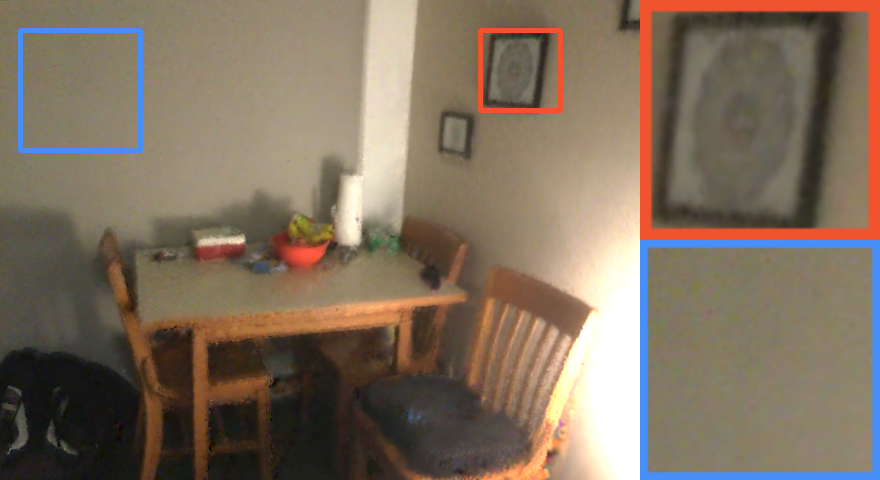}\\
        \end{tabular}
        \caption{ScanNet~\cite{dai2017scannet}}
        \label{fig:qualitative_rgb_tum}
    \end{subfigure}
    \caption{Qualitative results on applying \islam~to the RGBD-SLAM method.}
    \label{fig:qualitative_rgbd}
\end{figure}

\subsubsection{Qualitative Results}
We demonstrate the rendered images of \islam~with $\text{NeRF-SLAM}^\dagger$ in \cref{fig:qualitative_rgb}.
As shown in \cref{fig:qualitative_rgb} (a), \islam~successfully renders sharp images, whereas $\text{NeRF-SLAM}^\dagger$ has blurry artifacts come from the degraded input, for example, bricks in \texttt{Sponza}, boundary of the tire in \texttt{LOU}, complex texture of books and thin legs of the chair in \texttt{Italian-flat-0}.
More importantly, our approach also shows remarkable performance in reconstructing sharp maps from TUM-RGBD data which contains the real-world camera motion.
The texts on the books and objects on the table are clearly rendered with \islam~when we compare to the results of baseline and blurred input frames.
In \cref{fig:qualitative_rgbd}, we show samples of rendered images in RGBD scenario.
We can observe the consistent improvement over SplaTAM in generating a sharp map in our synthetic dataset.
The letters on the book cover and the bricks are clearly rendered with \islam.
Also, we observe significant map quality enhancement in the ScanNet dataset.
The letters on the board and tassels in \texttt{0024-01}, the detection mark and objects on the table in \texttt{0736-00}, and the pictures in \texttt{0785-00} are deblurred with our approach.
Furthermore, since the color of Gaussians in SplaTAM is consistent regardless of the exposure of images, SplaTAM does not appropriately model the brightness changes as we can see in the darker color of bricks in \texttt{Sponza} and tiled artifacts in \texttt{0785-00}.
Such artifacts are removed by modeling the tone mapping process.

\subsection{Ablation Study}
\label{subsec:ablation}

\newcommand{\cmark}{\ding{51}}
\newcommand{\xmark}{\ding{55}}
\begin{table}[t]
\begin{minipage}{.4\linewidth}
\caption{Ablation study.}
\vspace{-1em}
\label{tab:ablation}
\centering
\resizebox{\linewidth}{!}{
\begin{tabular}{ccccccc}
\toprule
\multirow{2}{*}{\begin{tabular}[c]{@{}c@{}}Traj. \\ Reg.\end{tabular}} &
  \multirow{2}{*}{\begin{tabular}[c]{@{}c@{}}Tone \\ Mapping\end{tabular}} &
  \multirow{2}{*}{\begin{tabular}[c]{@{}c@{}}Motion \\ Blur\end{tabular}} &
  \multirow{2}{*}{ATE} &
  \multirow{2}{*}{PSNR} &
  \multirow{2}{*}{SSIM} &
  \multirow{2}{*}{LPIPS} \\
       &        &        &      &       &       &       \\ \hline

\cmark & \cmark & \cmark & \textbf{2.56} & \textbf{25.62} & \textbf{0.801} & \textbf{0.195} \\
\xmark & \cmark & \cmark & 2.66 & 24.80 & 0.769 & 0.203 \\
\cmark & \xmark & \cmark & 2.60 & 22.39 & 0.756 & 0.226 \\
\xmark & \xmark & \cmark & 2.63 & 22.36 & 0.755 & 0.228 \\
\xmark & \xmark & \xmark & 2.71 & 23.05 & 0.793 & 0.235 \\
\bottomrule
\end{tabular}
}
\end{minipage}
\begin{minipage}{.53\linewidth}
\caption{Runtime analysis.}
\vspace{-1em}
\label{tab:runtime}
\centering
\resizebox{\textwidth}{!}{
\begin{tabular}{lcccccc}
\toprule
          & Mapping & Tracking & \multirow{2}{*}{ATE} & \multirow{2}{*}{PSNR} & \multirow{2}{*}{SSIM} & \multirow{2}{*}{LPIPS} \\
          & /Frame  & /Frame   &                           &                       &                       &                        \\ \midrule
\islam   & 2.058s  & 8.494s   & 1.42                      & 24.16                 & 0.786                 & 0.211                  \\
\islam-\textit{S} & 0.436s  & 1.867s   & 1.50                      & 23.92                 & 0.795                 & 0.230                  \\
SplaTAM~\cite{keetha2023splatam}   & 0.424s  & 1.433s   & 1.81                      & 22.26                 & 0.800                 & 0.224                  \\
SplaTAM-\textit{S}~\cite{keetha2023splatam} & 0.086s  & 0.377s   & 2.40                      & 21.43                 & 0.777                 & 0.243                  \\
\bottomrule
\end{tabular}
}
\end{minipage}
\end{table}

We conduct an ablation study in ScanNet~\cite{dai2017scannet} dataset.
\Cref{tab:ablation} shows that each component plays a crucial role in improving tracking or mapping performance.
In~\cref{tab:ablation}, Traj. Reg. refers to the trajectory regularization in~\cref{subsec:mapping_and_tracking}.
Motion blur-aware rendering improves tracking performance by preventing the accumulation of tracking errors.
HDR map shows improved rendering quality over LDR representations.
Trajectory regularization affects multi-camera optimization, resulting in enhancements in both tracking and mapping.

\subsection{Runtime analysis}
\label{subsec:runtime}

\begin{figure}[t]
    \vspace{-1em}
    \centering
    \begin{subfigure}[b]{0.35\linewidth}
    \centering
    \includegraphics[width=\linewidth]{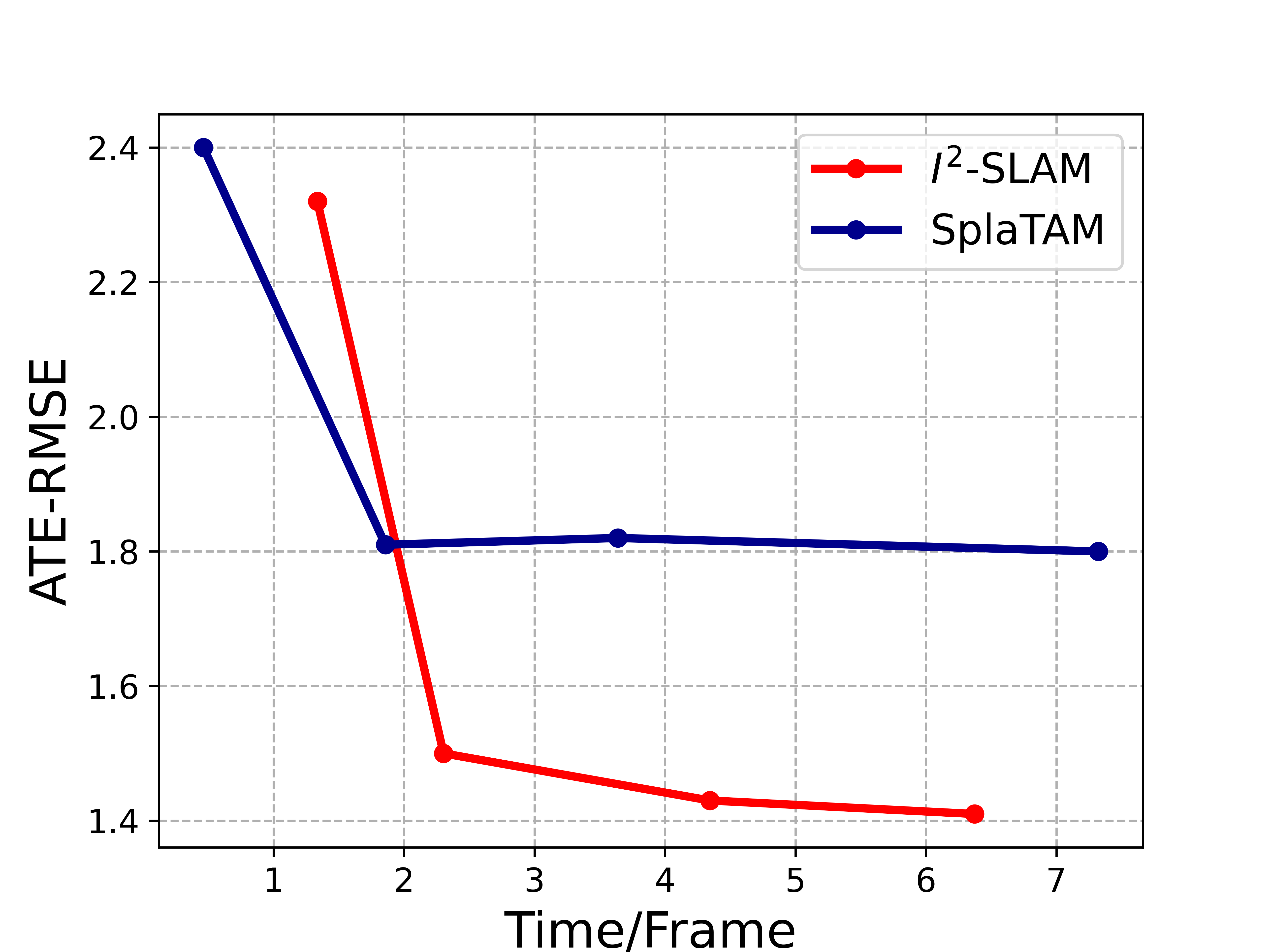}
    \caption{ATE-RMSE over iteration time}
    \label{fig:sub1}
    \end{subfigure} 
    \begin{subfigure}[b]{0.35\linewidth}
    \centering
    \includegraphics[width=\linewidth]{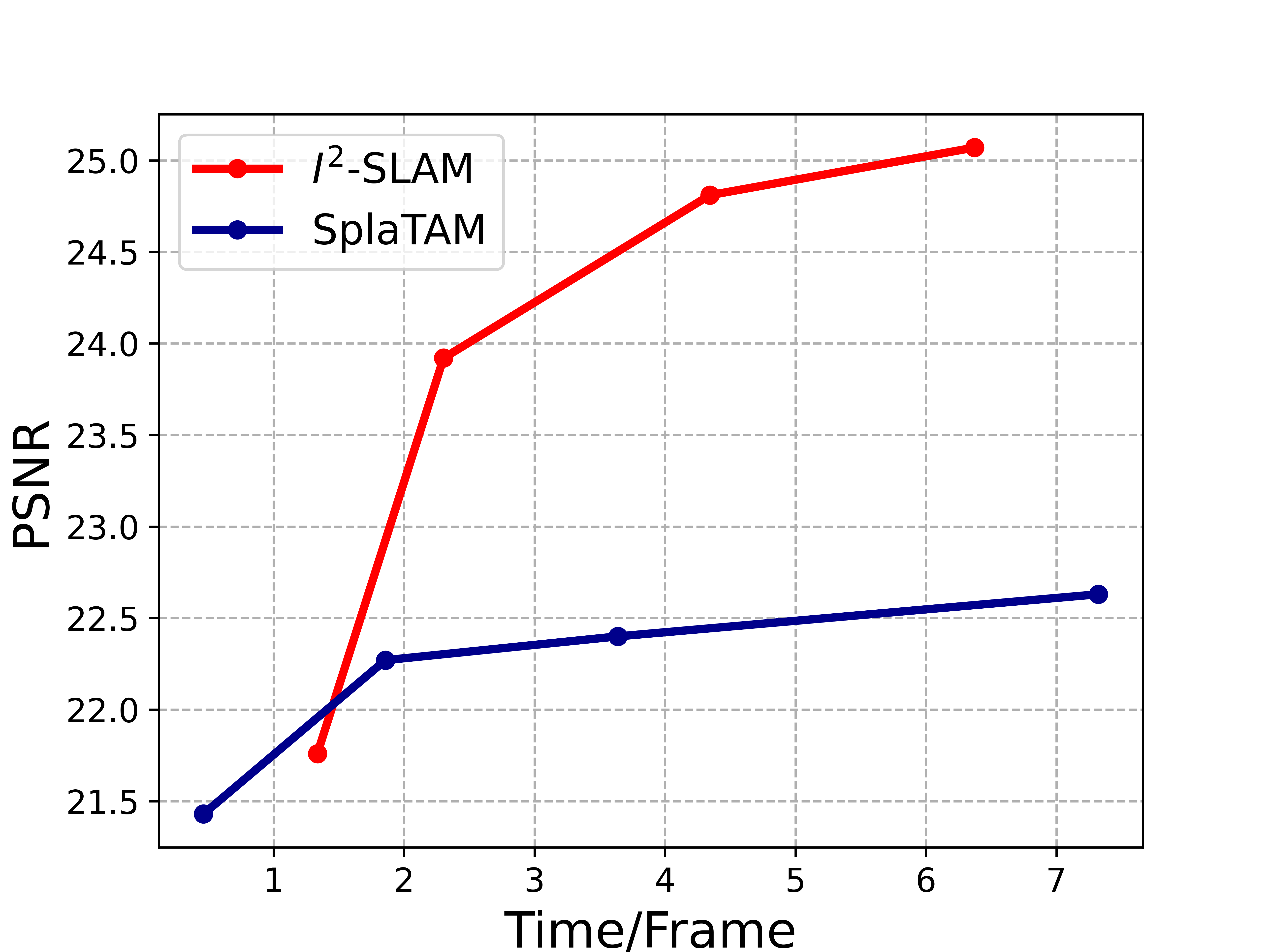}
    \caption{PSNR over iteration time}
    \label{fig:sub1}
    \end{subfigure} \hfill
    \caption{Performance variations over iteration time of \islam~and SplaTAM~\cite{keetha2023splatam}.}
    \label{fig:runtime_graph}
    \vspace{-1em}
\end{figure}

We analyze how much our approach affects the runtime and performance with ScanNet \texttt{0024-01} in~\cref{tab:runtime}.
\islam~exhibits better ATE-RMSE and PSNR metrics compared to SplaTAM~\cite{keetha2023splatam}, but it comes at the cost of slower speed.
However, \islam\textit{-S}, which uses $20\%$ iterations, takes a similar runtime but yields better performance.
In contrast, SplaTAM\textit{-S}, which also uses $20\%$ iterations, shows a significant decrease in performance.

In~\cref{fig:runtime_graph}, we analyze the trend of performance changes concerning the number of iterations. Four models with different numbers of iterations are compared. In most cases, our method demonstrates better tracking and rendering performance when using a similar runtime.

\section{Conclusion}
\label{sec:conclusion}
We present \islam, a generic module that improves the quality of existing visual SLAM approaches by inverting the image formation process for casually captured videos.
We show the effectiveness of our module by incorporating it into the state-of-the-art methods in RGB and RGBD-SLAM approaches.
With \islam, existing visual SLAM pipelines can effectively handle the varying appearance and motion blur that are prevalent in-the-wild capturing scenarios.

Although \islam~is able to reconstruct sharp HDR maps while estimating the camera trajectory from casually captured videos, there are some limitations.
The motion blur simulation with the sum of multiple cameras accompanies the longer rendering and optimization time.
An efficient approximation method to simulate camera motion blur would be an interesting research direction to improve our approach.

\subsubsection{Acknowledgements}
This work is supported by the National Research Foundation of Korea(NRF) grant (No. RS-2023-00218601) and IITP grant [NO.RS-2021-II211343, Artificial Intelligence Graduate School Program (Seoul National University)] funded by the Korea government(MSIT). 

\clearpage

\bibliographystyle{splncs04}
\bibliography{references}

\clearpage
\appendix
\section{Implementation Details}
\paragraph{RGB-SLAM}
We reimplement NeRF-SLAM~\cite{rosinol2023nerf} with torch-ngp~\cite{torch-ngp}. 
Our model takes the same inputs as in NeRF-SLAM~\cite{rosinol2023nerf}.
We utilize camera poses, depth estimates, and uncertainty maps of keyframes from NeRF-SLAM.
We sequentially add keyframes to the training set.
The learning rate for camera tracking is $1e-4$.
The learning rate for the tone-mapping module is $1e-4$.
We utilize camera preconditioning~\cite{park2023camp} for virtual camera pose optimization.
We use photometric errors instead of 2D projections to make camera poses to change in motion-blur direction.

\paragraph{RGBD-SLAM}
For ScanNet~\cite{dai2017scannet}, we follow the experimental setup of SplaTAM~\cite{keetha2023splatam}, except for unpadding edges.
We unpad $24$ pixels in color images and $12$ pixels in depth images to ignore black regions caused by the calibration process.
\islam~parameterizes virtual camera poses with center poses and velocities.
We use the same learning rate for the center pose and translational velocity as SplaTAM, while the learning rate for the rotational velocity is set to one-tenth.
The learning rate for the tone-mapping module is $1e-2$.
The learning rate for the trajectory regularizer is $1e6$.
\islam~optimizes monochromatic WB for ScanNet.
For the synthetic dataset, \islam~optimizes monochromatic WB and monochromatic CRF.
We initialize virtual camera poses by interpolating between the current frame and the previous frame by a factor of $1/10$.
The learning rate for trajectory regularizer is $1e4$.
The learning rate for camera tracking is $5e-5$.
The mapping interval is every five frames.
We use $30$ iterations for mapping and $200$ iterations for tracking.

\section{Dataset}
\paragraph{Synthetic dataset}
We create a new synthetic dataset that simultaneously considers autoexposure and motion blur with Blender's \cite{blender} Cycles engine. 
We use three Blender scenes with trajectories manually set for each scene pretending a human is capturing the scene.
At first, we render HDR images along the trajectory without motion blur for ground truth images. 
Then we render LDR images from the HDR images based on the classical Reinhard tone-mapping  \cite{10.1145/566654.566575}. 
We define our tone-mapping function following HDR-NeRF \cite{huang2022hdr}:
\begin{equation}
    I_{LDR} = 255((\Delta t I_{HDR}/ (\Delta t I_{HDR}+1))^{1/2.2},
\end{equation}
where $I_{LDR}$ and $ I_{HDR}$ are the pixel value of LDR and HDR images, $\Delta t=2^{EV}$ is the exposure time and $EV$ is the exposure value. 
We set $EV$ for each frame by converting the HDR RGB image to gray values and mapping the LDR output of the median gray value to $255/2$. 
Finally, we blur the LDR images using Blender's built-in motion blur function. 
We set the shutter curve as a square function. The shutter value ${SV}_i$ of the $i$th frame is decided by the exposure as follows:
\begin{equation}
    SV_i = a2^{EV_{i}}+b,
\end{equation}
where $a$ and $b$ are the predefined parameters per scene and trajectory. 
Specifically, we set $a$ and $b$ by forcing the minimum and maximum shutter values to be 0.1 and 1 for each trajectory in the scene for our dataset.

\begin{figure}[t]
    \centering
    \begin{subfigure}[b]{0.24\linewidth}
    \includegraphics[width=\linewidth]{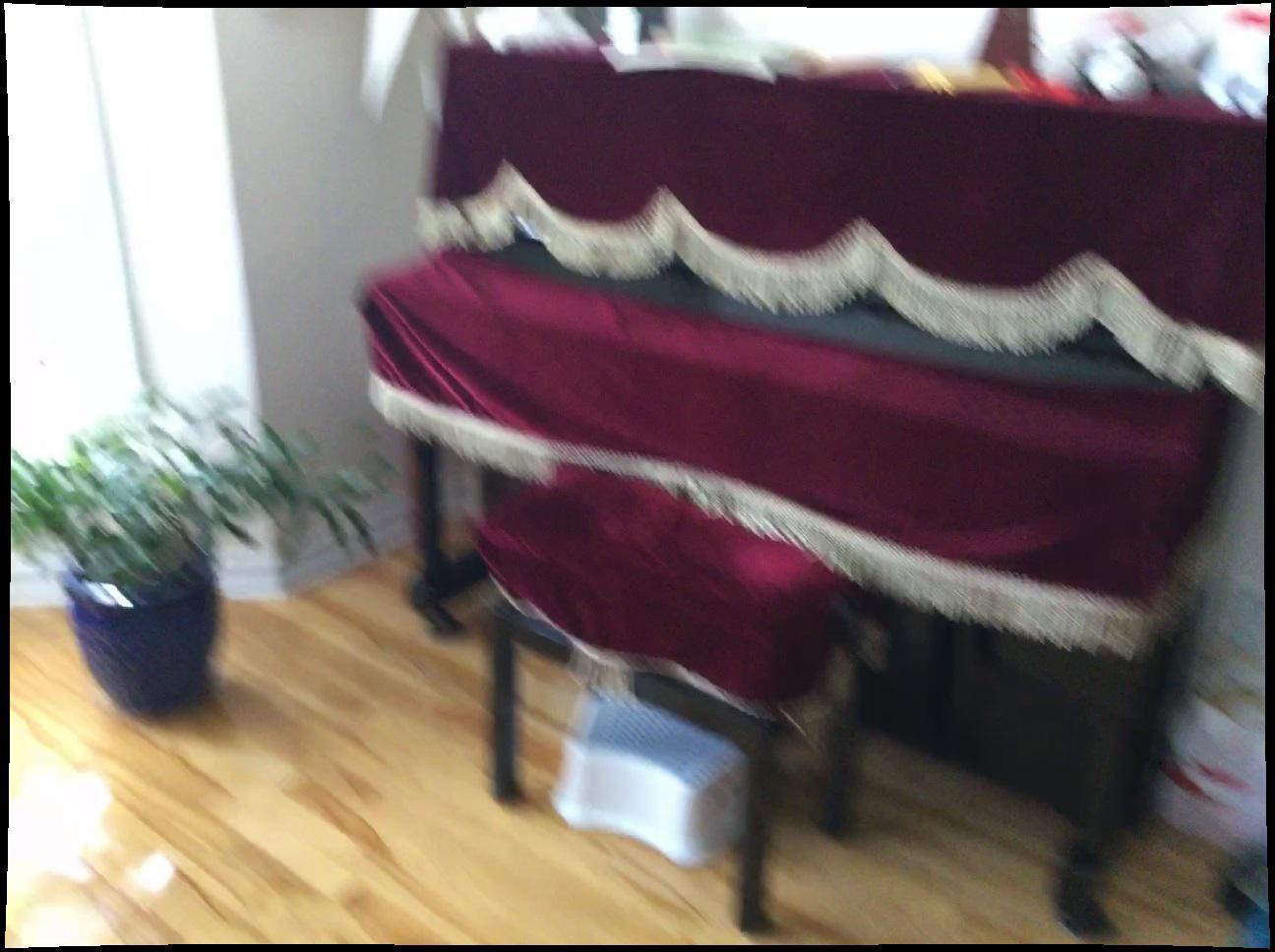}
    \caption{Blurry keyframe\protect\\in ScanNet~\cite{dai2017scannet}}
    \label{fig:sub1}
    \end{subfigure} \hfill
    \begin{subfigure}[b]{0.24\linewidth}
    \includegraphics[width=\linewidth]{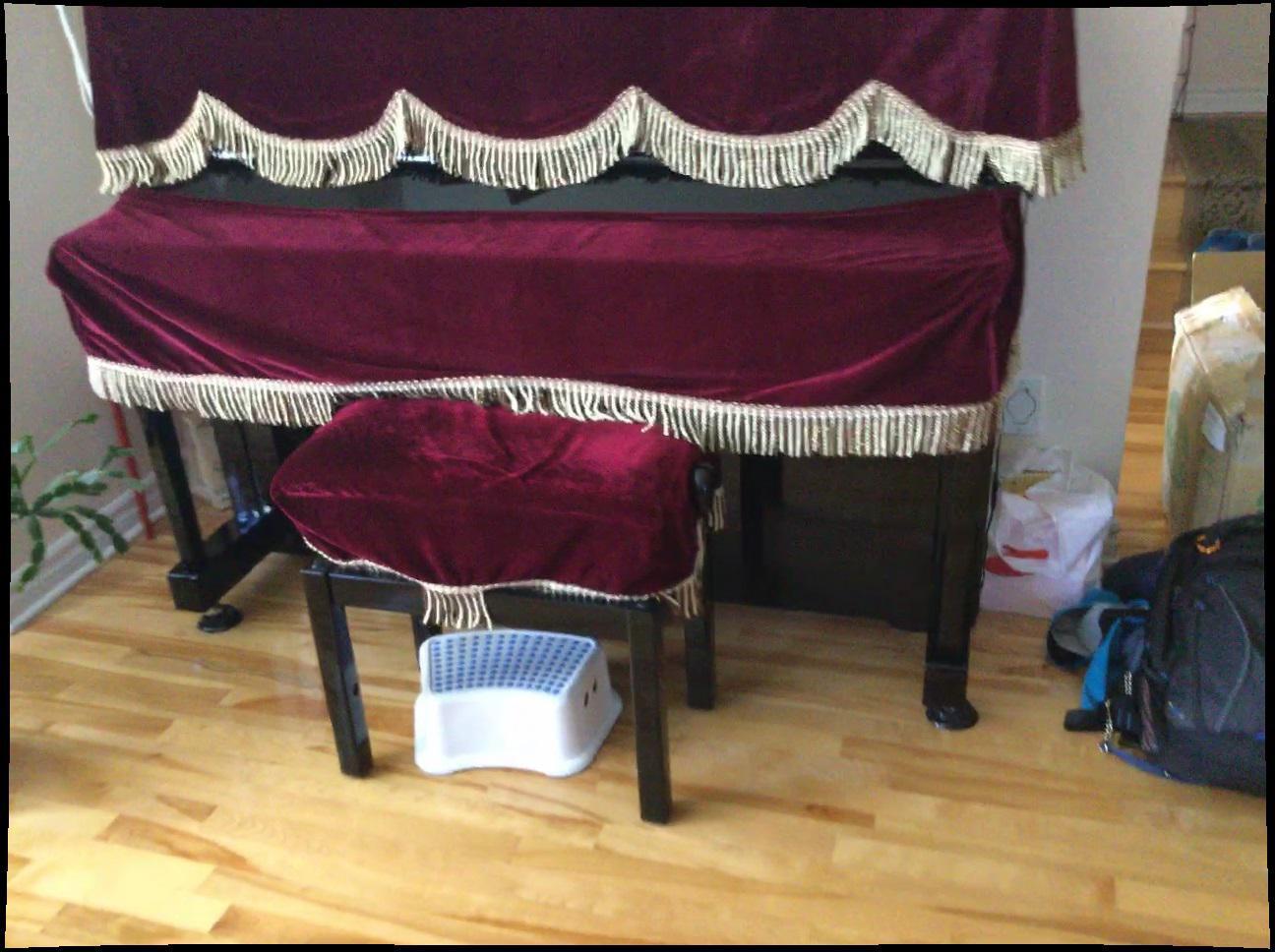}
    \caption{Sharp keyframe\protect\\in ScanNet~\cite{dai2017scannet}}
    \label{fig:sub1}
    \end{subfigure} \hfill
    \begin{subfigure}[b]{0.24\linewidth}
    \includegraphics[width=\linewidth]{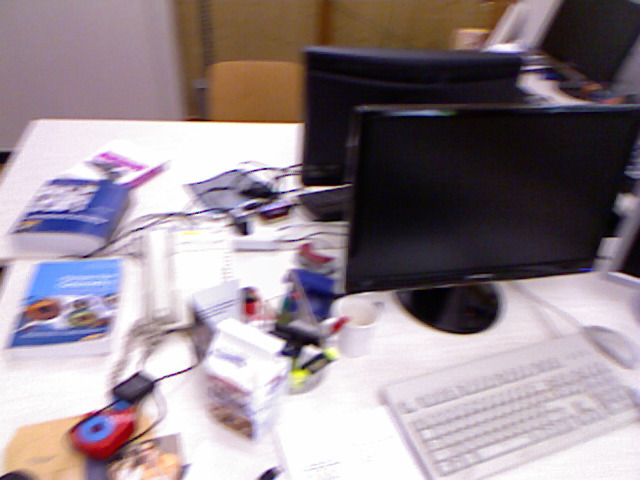}
    \caption{Blurry keyframe\protect\\in TUM-RGBD~\cite{sturm2012benchmark}}
    \label{fig:sub1}
    \end{subfigure} \hfill
    \begin{subfigure}[b]{0.24\linewidth}
    \includegraphics[width=\linewidth]{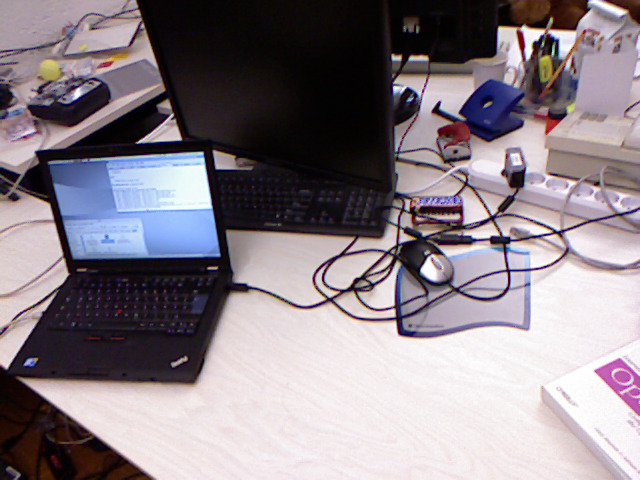}
    \caption{Sharp keyframe\protect\\in TUM-RGBD~\cite{sturm2012benchmark}}
    \label{fig:sub1}
    \end{subfigure} \hfill
    \caption{Examples of sharp keyframes and blurry keyframes in real-world datasets.}
    \label{fig:examples}
\end{figure}

\paragraph{Real-world dataset}
We use part of the sequences as input instead of the entire sequences for experiments in Scannet (0$\sim$300 in \texttt{0024-01}, 0$\sim$300 in~\texttt{0031-00}, 1000$\sim$1400 in~\texttt{0736-00}, and 0$\sim$500 in~\texttt{0785-00}).
As real-world datasets are heavily affected by motion blur, it is unlikely to evaluate the image quality with the entire input as a reference.
Therefore, we manually annotate sharp images from input image sequences to evaluate rendering quality in real-world datasets. 
$9.5\%$ of input frames from ScanNet~\cite{dai2017scannet} and $29.7\%$ from TUM-RGBD~\cite{sturm2012benchmark} are selected for evaluation. 
We select the entire frame of \texttt{fr2/xyz} of TUM-RGBD for evaluation since the corresponding scene was carefully captured to avoid motion blur. 
~\Cref{fig:examples} depicts examples of selected sharp and blurry images.

\section{Trajectory Regularization}
In this section, we explain our trajectory loss terms (Eq. 16 and 17 in the main manuscript) in more detail, especially describing the relationship between the global scale parameter $a$ and the frame rate of input video.

The trajectory loss guides the start and end camera poses to be located on the global trajectory.
Here, global trajectory stands for the trajectory defined by the middle camera position of each frames.
Specifically, we define trajectory loss as a L2-norm between start and end camera poses and the one point on linear interpolant of $\mathbf{t}^i$ and $\mathbf{t}^{i-1}$.
Also, the interpolating parameter should be determined by the timestamp of start and end poses.

Also, when we assume the linear motion during the exposure time, $t^i-t^{i-1}=\dfrac{1}{f}$, where $t^i=\dfrac{t^i_s+t^i_e}{2}$ and $f$ is the frequency of the input video samples.
Also, $t^i - t^i_s = \dfrac{\Delta t^i}{2}$.
Therefore, the translation vector of the start position $\mathbf{t}(t^i_s)$ should be located on LERP$(\mathbf{t}^{i-1}, \mathbf{t}^{i}, 1-\dfrac{f\Delta t^i}{2})$.
Now, we can find the relationship between the global scale parameter $a$ and the video frame rate $f$, $a=\dfrac{f}{2}$.
Same analogy holds for the end camera location and rotation vectors.

\section{Additional Experimental Results}

\begin{table}[t]
    \caption{Rendering quality comparison against RGBD-SLAM baselines on ScanNet~\cite{dai2017scannet}. \islam~in this table represents our RGBD-SLAM model which is incorporated into SplaTAM~\cite{keetha2023splatam}.  
    }
    \label{tab:more_baselines_rendering}
    \centering
    \resizebox{0.7\textwidth}{!}{
    \begin{tabular}{lcccccccc}
        \toprule
        \multirow{2}{*}{Methods} & \multirow{2}{*}{Metrics} & \multicolumn{4}{c}{ScanNet~\cite{dai2017scannet}} \\
        & & \texttt{0024-01} & \texttt{0031-00} & \texttt{0736-00} & \texttt{0785-00} \\
         \midrule
        \multirow{3}{*}{Co-SLAM~\cite{wang2023co}} & PSNR & 17.34 & 21.05 & 16.38 & 19.89 \\
         & SSIM  & 0.570 & 0.645 & 0.525 & 0.647 \\
         & LPIPS  & 0.542 & 0.473 & 0.651 & 0.618 \\
         \midrule
        \multirow{3}{*}{Point-SLAM~\cite{sandstrom2023point}} & PSNR & 17.67 & 21.09 & 16.21 & 20.71 \\
         & SSIM  & 0.533 & 0.605 & 0.508 & 0.677 \\
         & LPIPS  & 0.439 & 0.417 & 0.508 & 0.483 \\
         \midrule
        \multirow{3}{*}{SplaTAM~\cite{keetha2023splatam}} & PSNR & 21.60 & 24.64 & \textbf{24.50} & 19.63 \\
         & SSIM  & \textbf{0.786} & 0.773 & \textbf{0.847} & 0.719 \\
         & LPIPS  & 0.236 & 0.275 & 0.182 & 0.340 \\
         \midrule
        \multirow{3}{*}{\islam} & PSNR  & \textbf{23.39} & \textbf{26.89} & 24.07 & \textbf{26.40} \\
         & SSIM & 0.780 & \textbf{0.796} & 0.828 & \textbf{0.762} \\
         & LPIPS & \textbf{0.180} & \textbf{0.236} & \textbf{0.175} & \textbf{0.238} \\
         \bottomrule
    \end{tabular}
    }
\end{table}

\begin{figure}[t]
    \centering
    \includegraphics[width=0.6\linewidth]{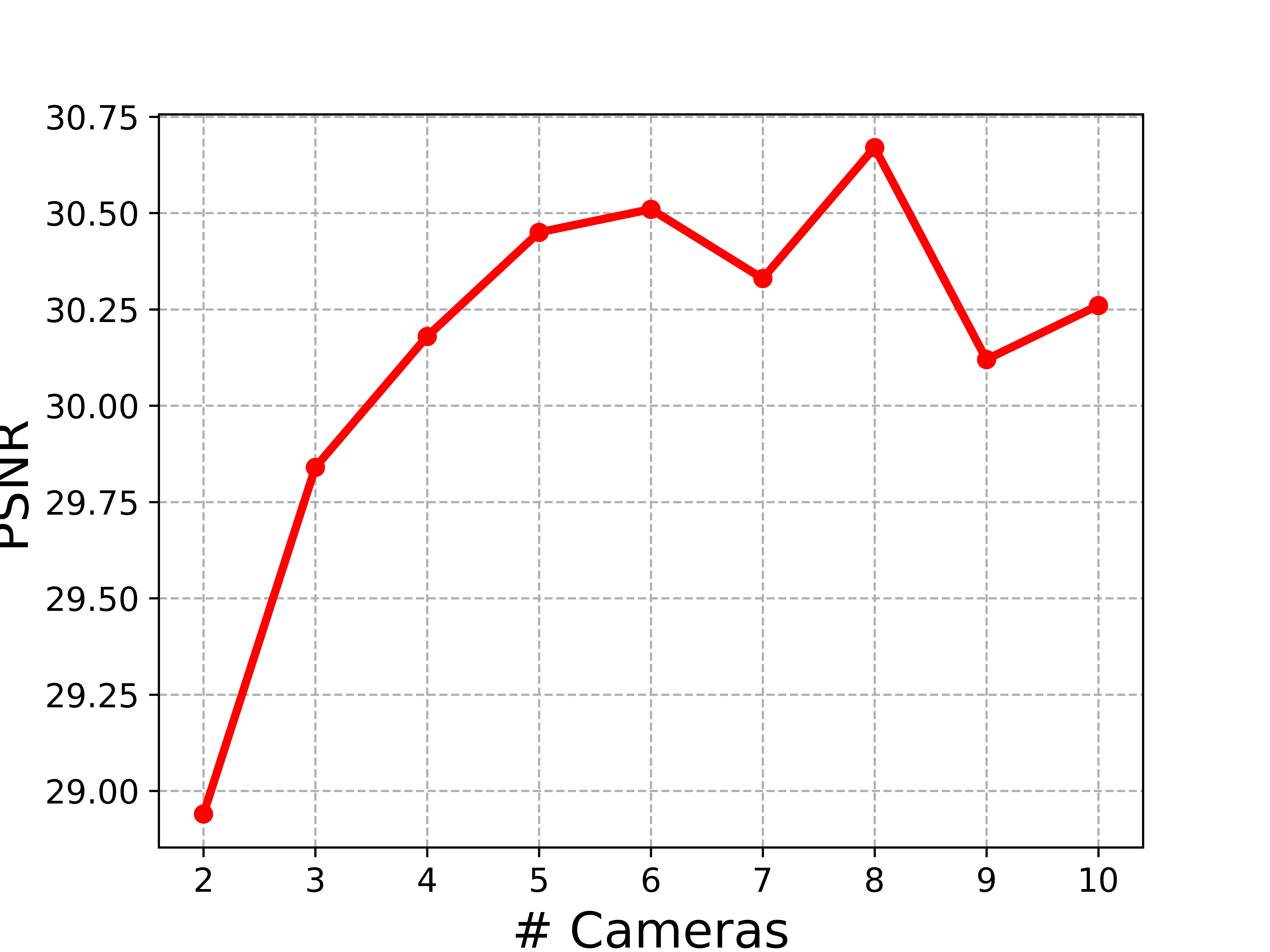}
    \caption{PSNR variation over the number of virtual cameras}
    \label{fig:cam_num_ablation}
\end{figure}

\begin{figure}[t]
    \centering
    \includegraphics[width=\linewidth]{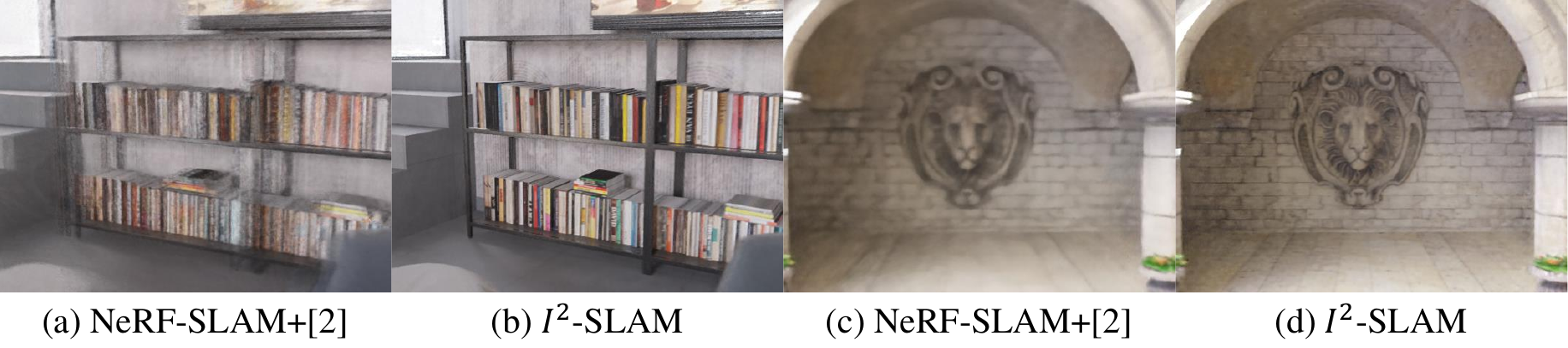}
    \caption{Renderings of RGB-SLAM with image enhancements.}
    \label{fig:image_deblurring}
\end{figure}

\begin{figure}[!ht]
    \centering
        \begin{tabular}{@{}c@{\,}c@{\,}c@{\,}c@{\,}c@{\,}c@{\,}}
            & \resizebox{0.12\textwidth}{!}{Input Frame} & \resizebox{0.12\textwidth}{!}{Co-SLAM~\cite{wang2023co}} &\resizebox{0.15\textwidth}{!}{Point-SLAM~\cite{sandstrom2023point}} &\resizebox{0.12\textwidth}{!}{SplaTAM~\cite{keetha2023splatam}} & \resizebox{0.09\textwidth}{!}{\islam} \\
            
            \rotatebox{90}{\quad\: \resizebox{0.07\textwidth}{!}{\texttt{0024-01}}} & \includegraphics[width=0.19\linewidth]{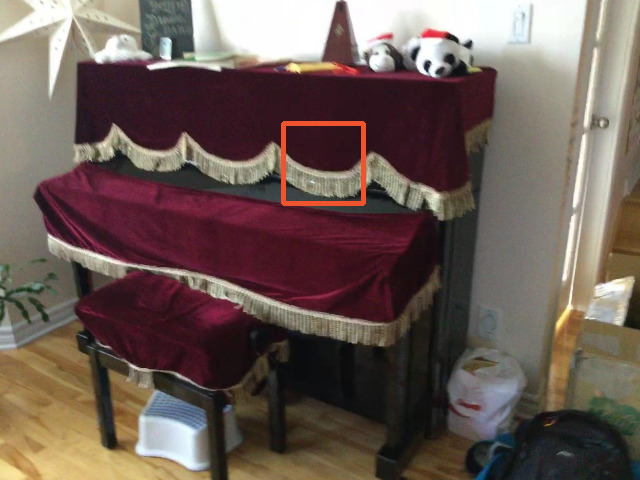}& \includegraphics[width=0.19\linewidth]{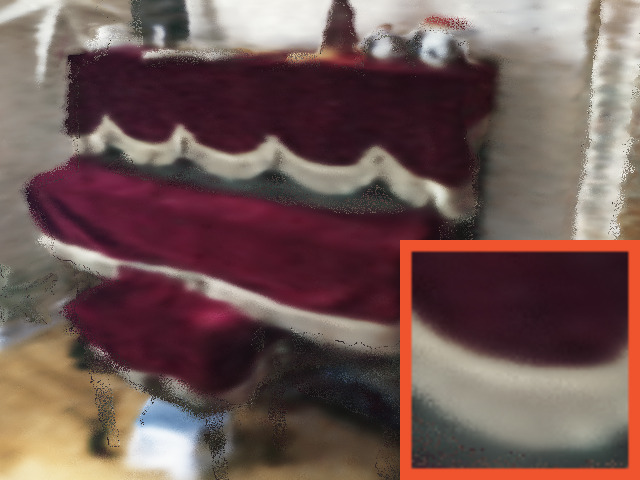} & \includegraphics[width=0.19\linewidth]{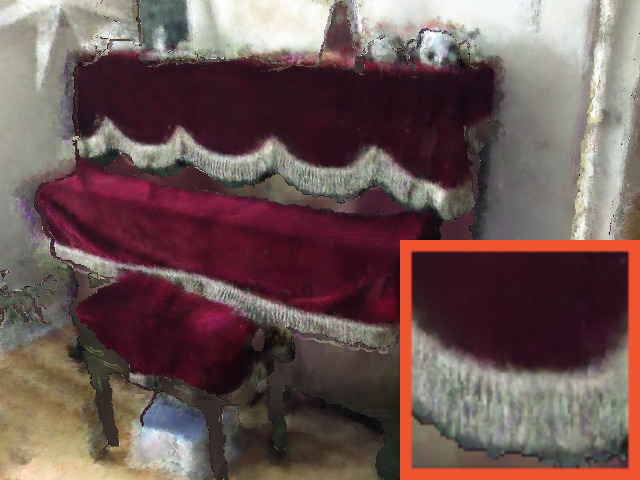} & \includegraphics[width=0.19\linewidth]{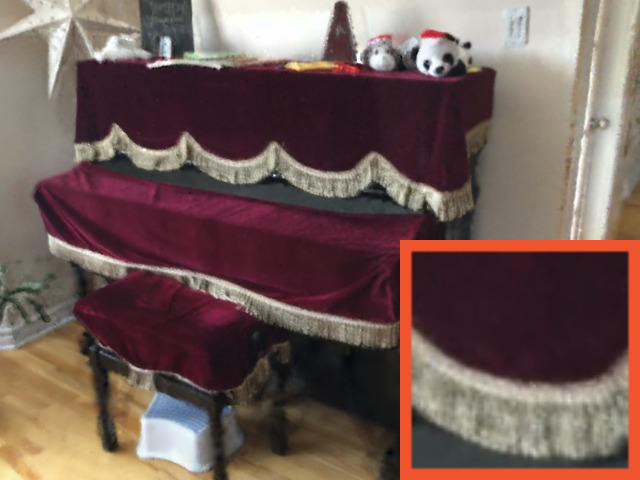} & \includegraphics[width=0.19\linewidth]{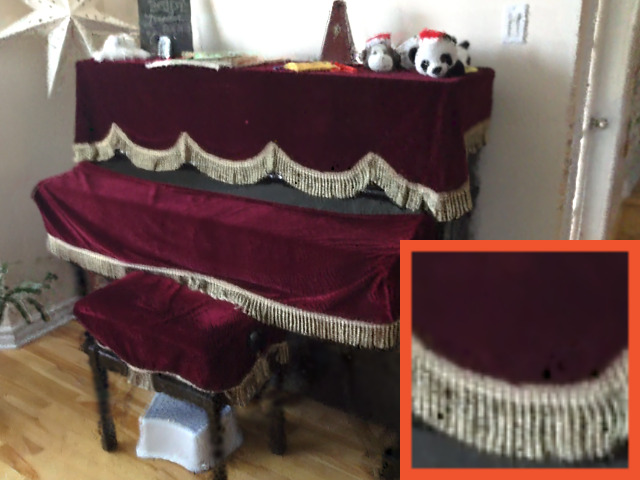} \\
            
            \rotatebox{90}{\quad\: \resizebox{0.07\textwidth}{!}{\texttt{0031-00}}} & \includegraphics[width=0.19\linewidth]{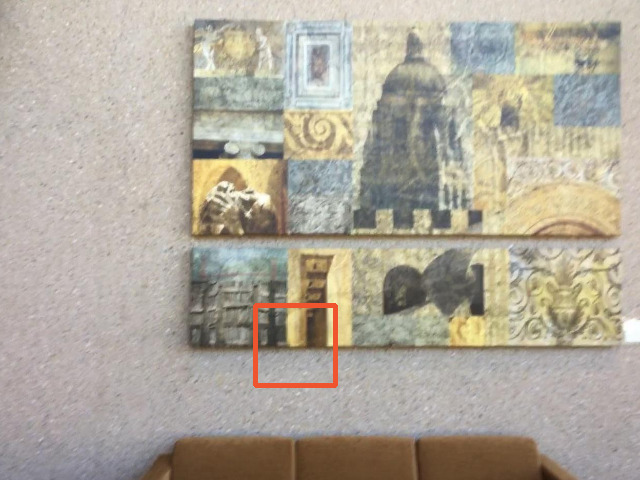}& \includegraphics[width=0.19\linewidth]{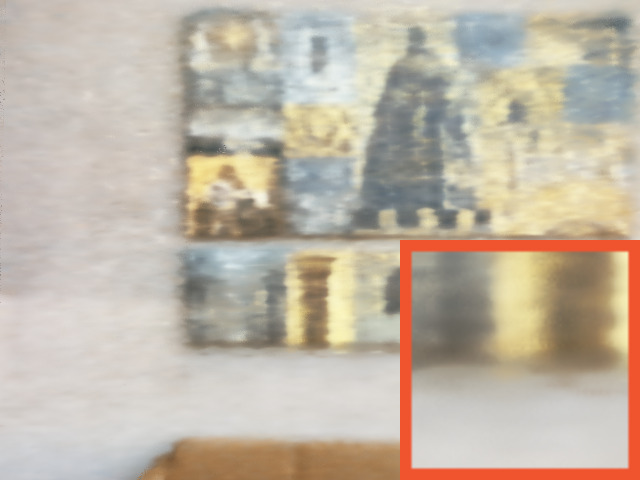} & \includegraphics[width=0.19\linewidth]{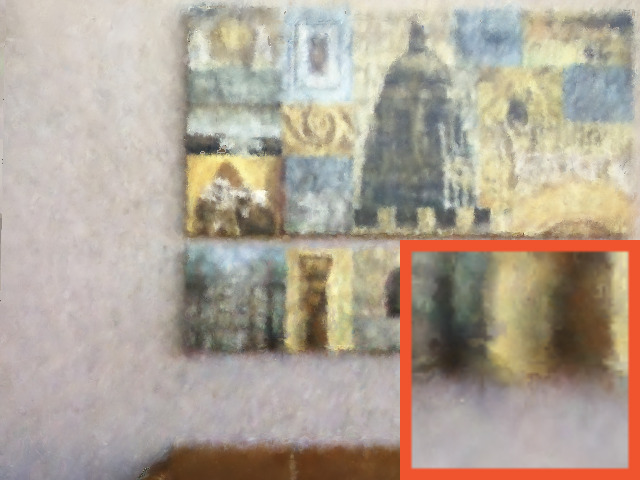} & \includegraphics[width=0.19\linewidth]{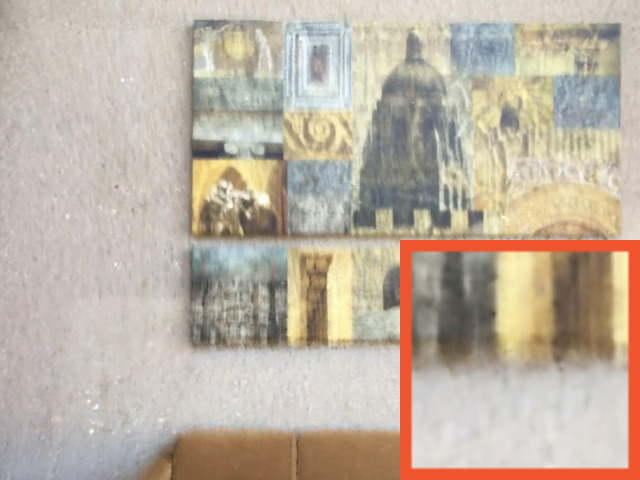} & \includegraphics[width=0.19\linewidth]{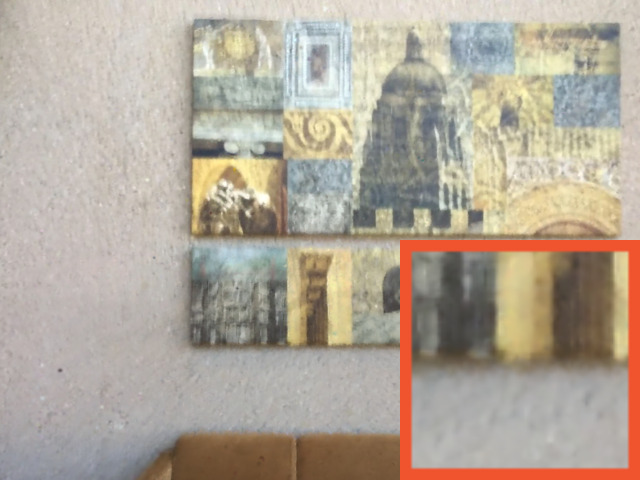} \\
            
            \rotatebox{90}{\quad\: \resizebox{0.07\textwidth}{!}{\texttt{0736-00}}} & \includegraphics[width=0.19\linewidth]{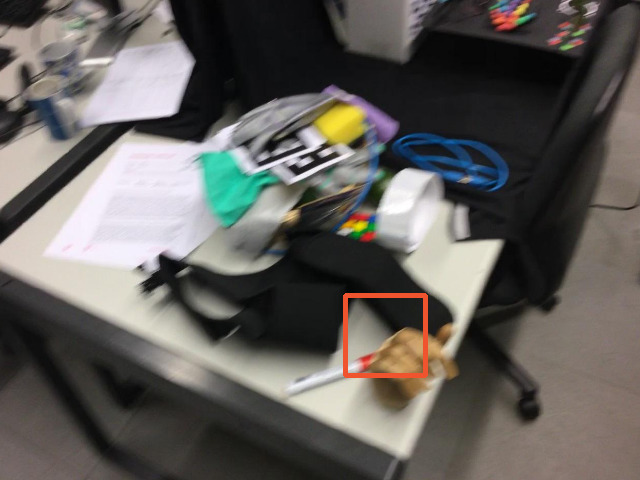}& \includegraphics[width=0.19\linewidth]{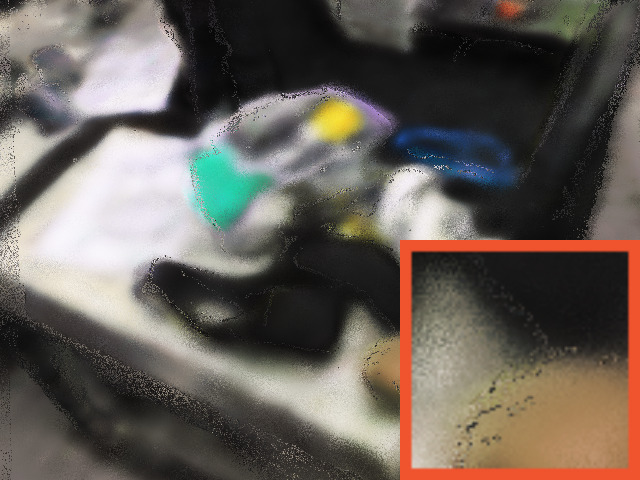} & \includegraphics[width=0.19\linewidth]{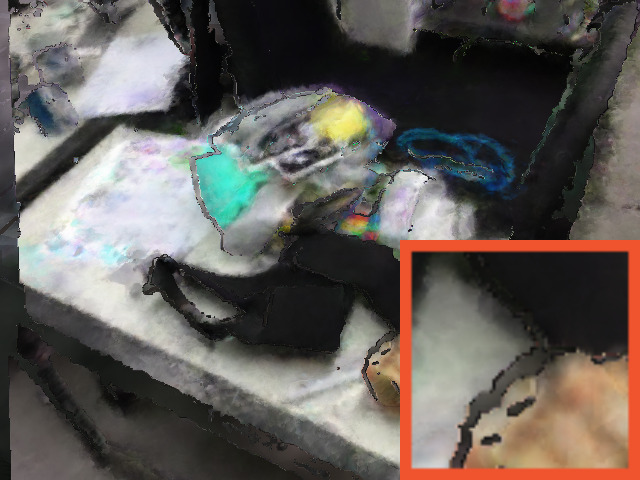} & \includegraphics[width=0.19\linewidth]{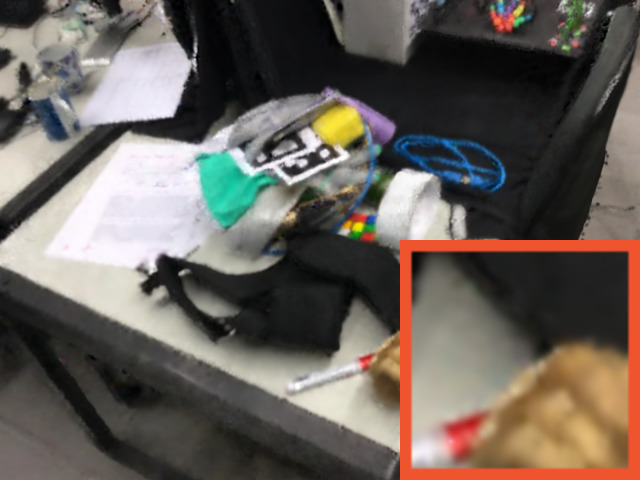} & \includegraphics[width=0.19\linewidth]{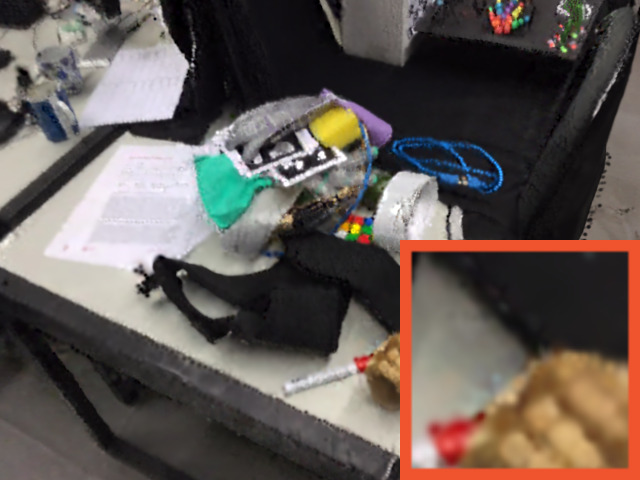} \\
            
            \rotatebox{90}{\quad\: \resizebox{0.07\textwidth}{!}{\texttt{0736-00}}} & \includegraphics[width=0.19\linewidth]{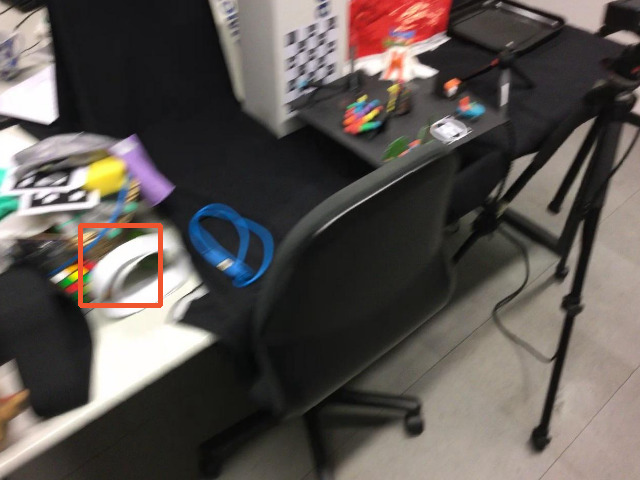}& \includegraphics[width=0.19\linewidth]{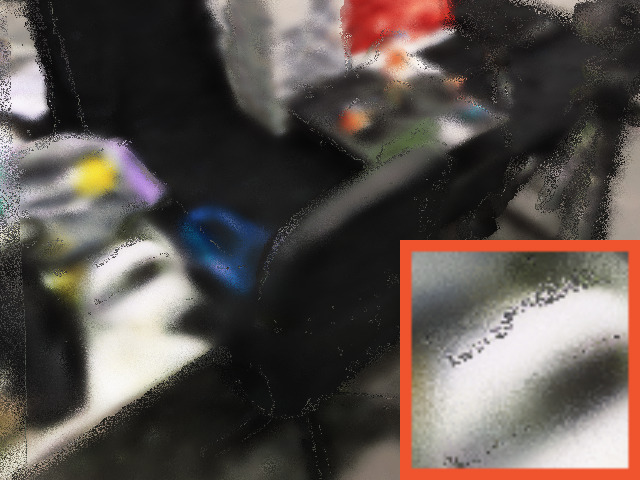} & \includegraphics[width=0.19\linewidth]{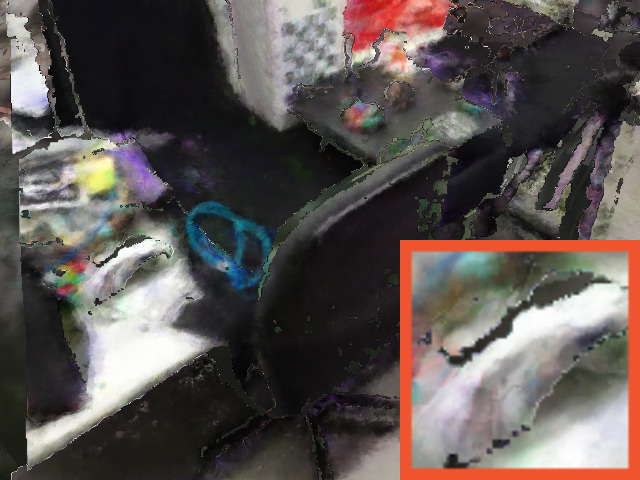} & \includegraphics[width=0.19\linewidth]{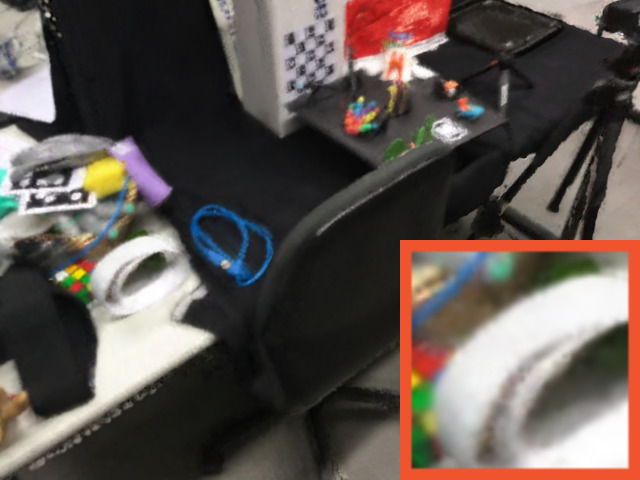} & \includegraphics[width=0.19\linewidth]{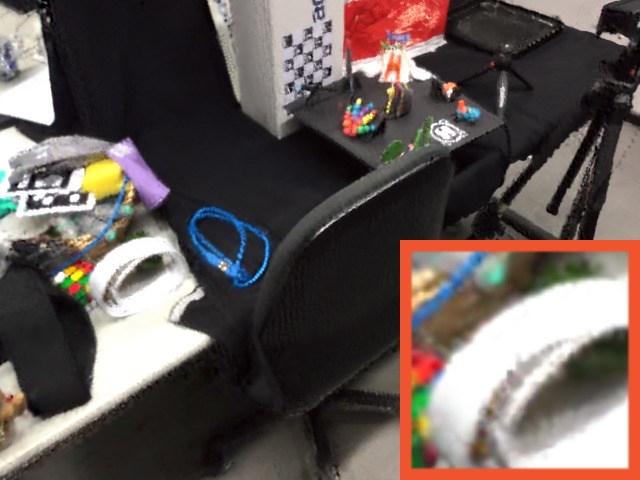} \\
            
            \rotatebox{90}{\quad\: \resizebox{0.07\textwidth}{!}{\texttt{0785-00}}} & \includegraphics[width=0.19\linewidth]{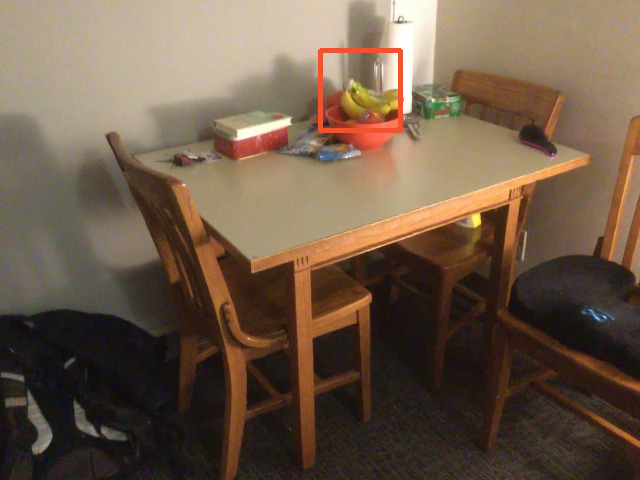}& \includegraphics[width=0.19\linewidth]{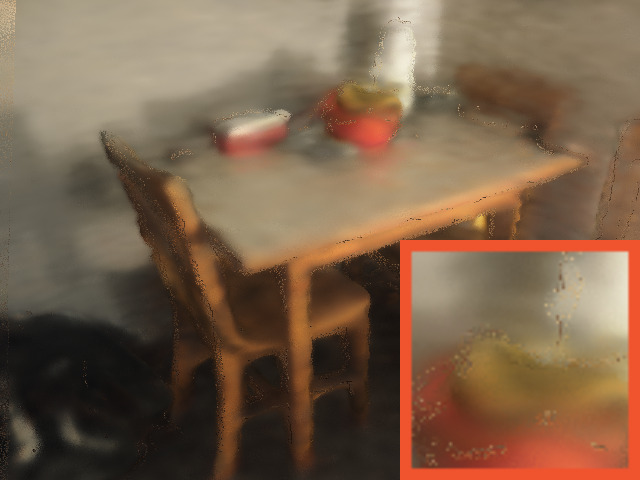} & \includegraphics[width=0.19\linewidth]{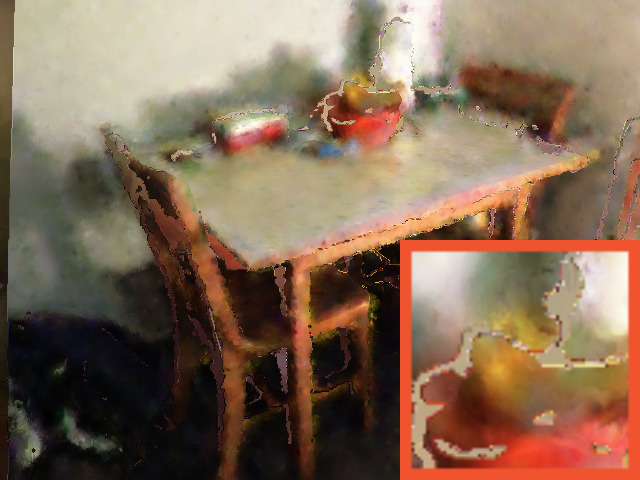} & \includegraphics[width=0.19\linewidth]{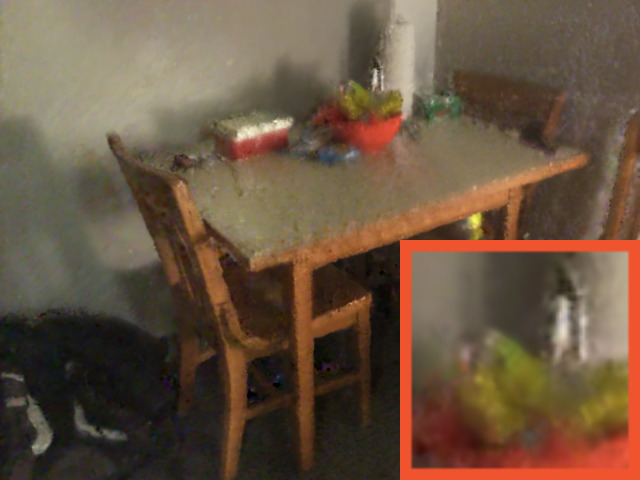} & \includegraphics[width=0.19\linewidth]{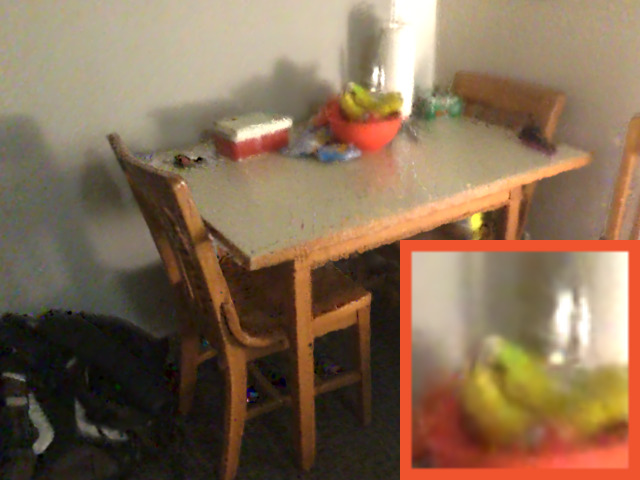} \\
            
            \rotatebox{90}{\quad\: \resizebox{0.07\textwidth}{!}{\texttt{0785-00}}} & \includegraphics[width=0.19\linewidth]{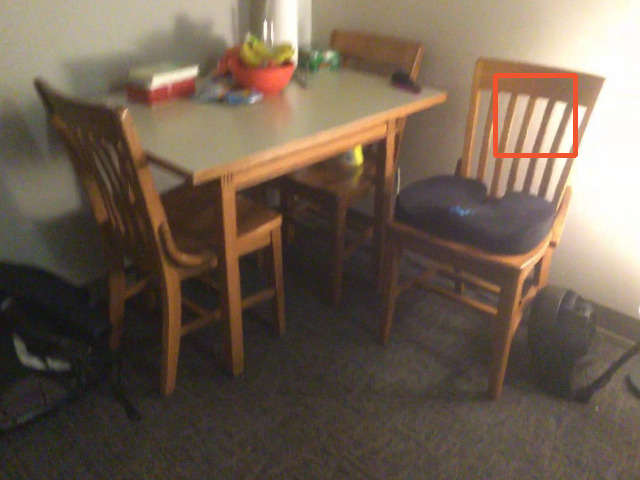}& \includegraphics[width=0.19\linewidth]{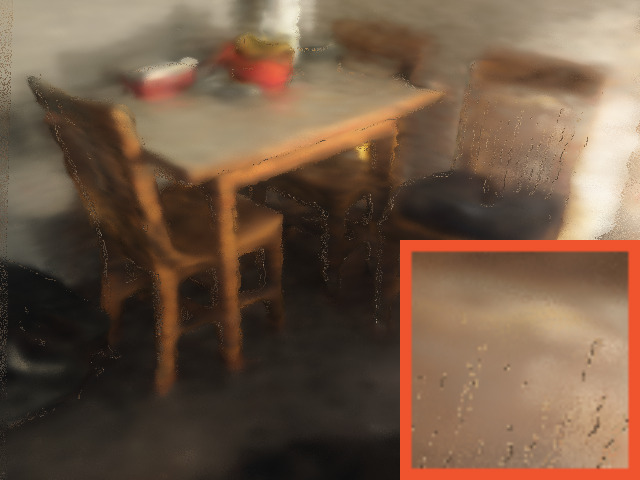} & \includegraphics[width=0.19\linewidth]{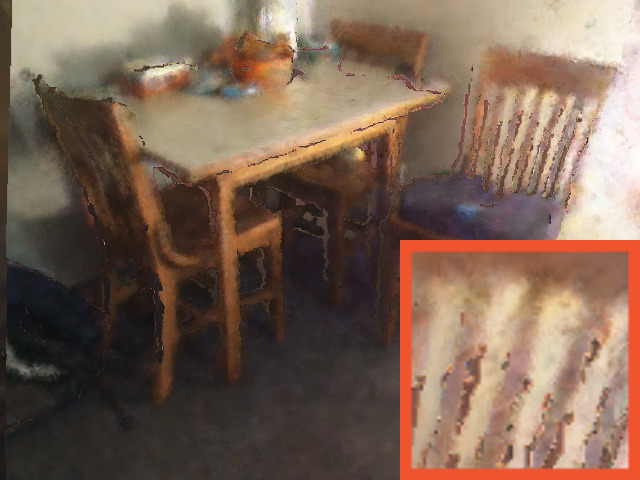} & \includegraphics[width=0.19\linewidth]{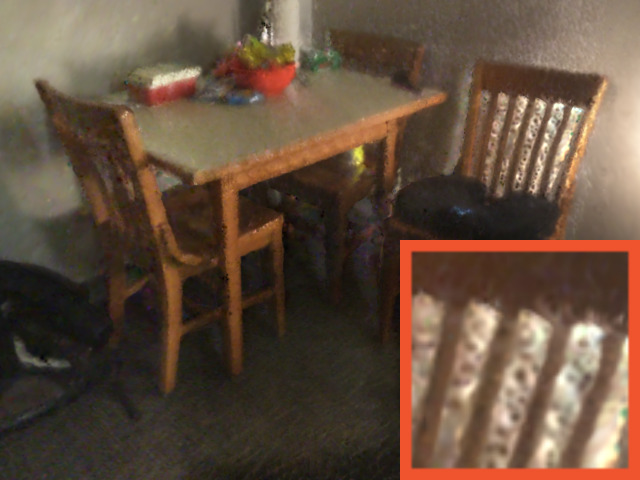} & \includegraphics[width=0.19\linewidth]{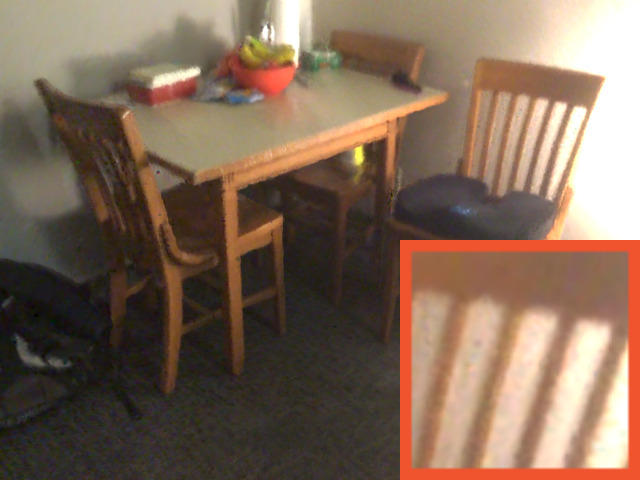} \\
            
            \rotatebox{90}{\quad\: \resizebox{0.07\textwidth}{!}{\texttt{0785-00}}} & \includegraphics[width=0.19\linewidth]{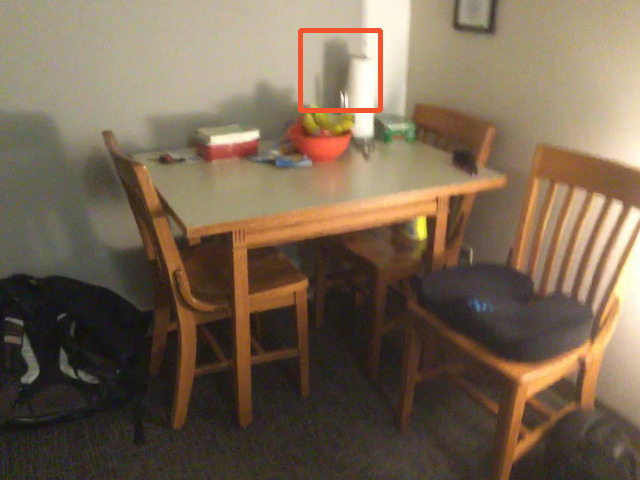}& \includegraphics[width=0.19\linewidth]{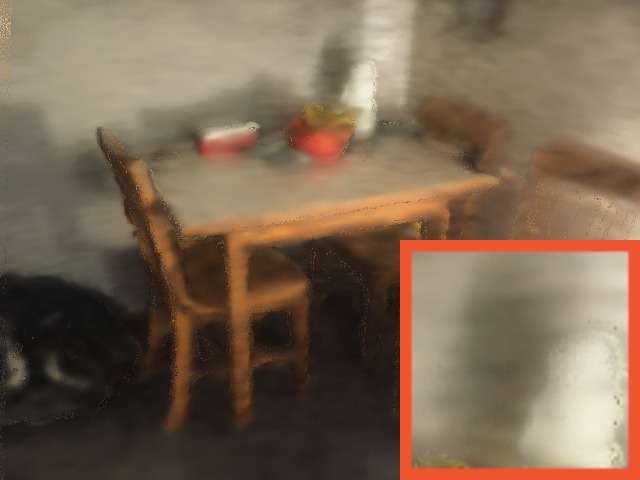} & \includegraphics[width=0.19\linewidth]{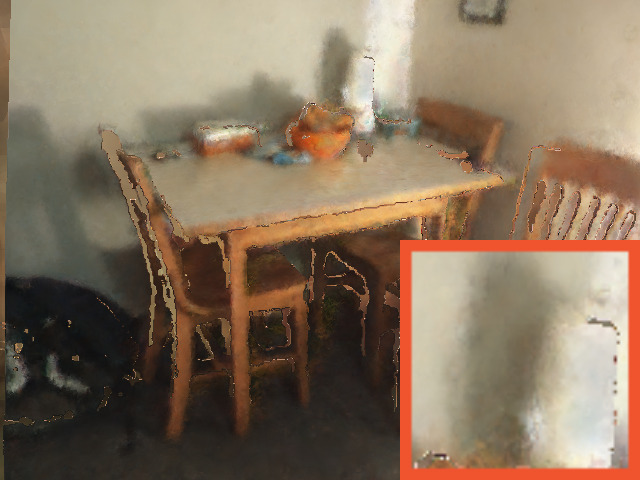} & \includegraphics[width=0.19\linewidth]{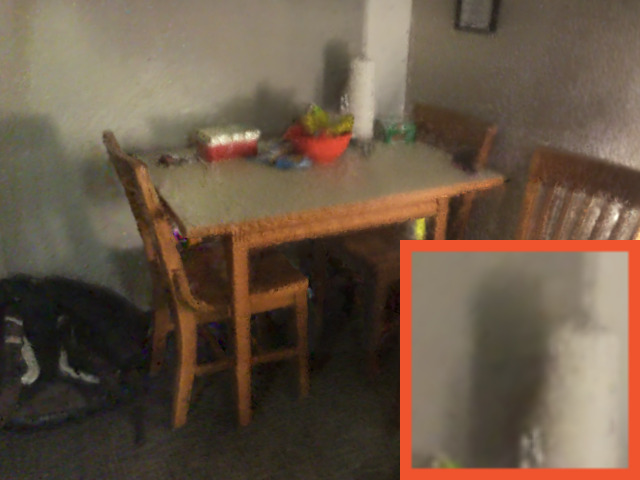} & \includegraphics[width=0.19\linewidth]{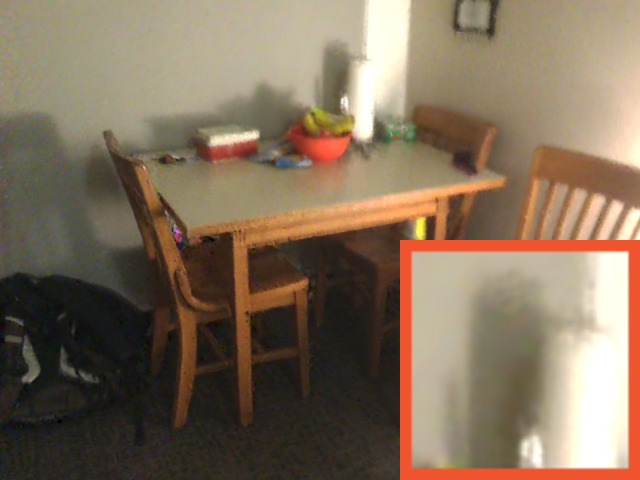} \\
            
            \rotatebox{90}{\quad\: \resizebox{0.07\textwidth}{!}{\texttt{0785-00}}} & \includegraphics[width=0.19\linewidth]{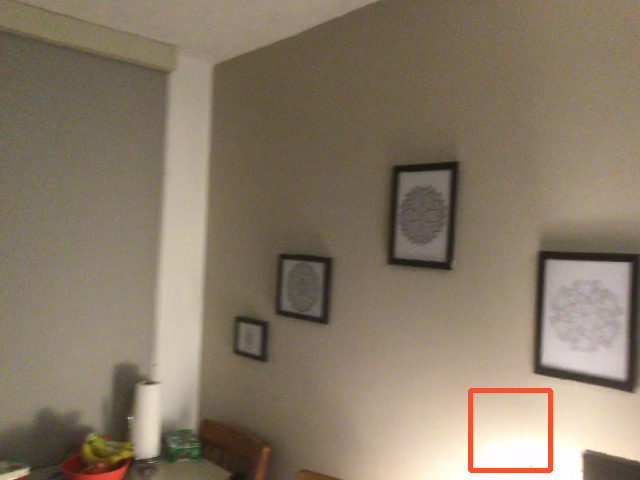}& \includegraphics[width=0.19\linewidth]{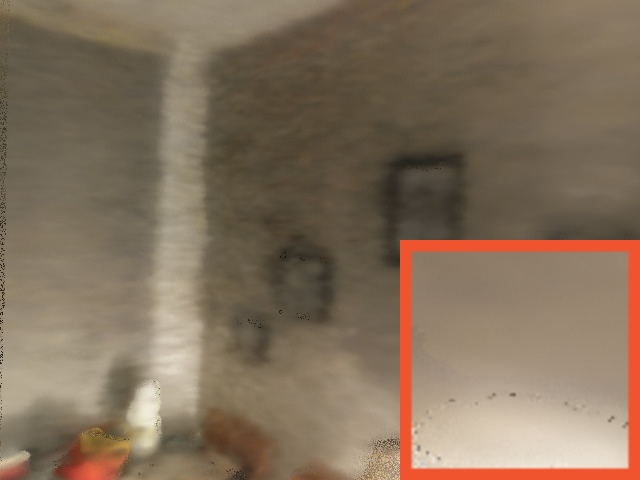} & \includegraphics[width=0.19\linewidth]{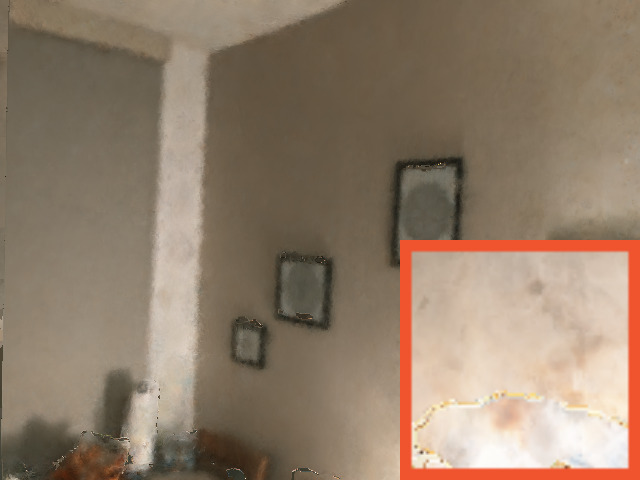} & \includegraphics[width=0.19\linewidth]{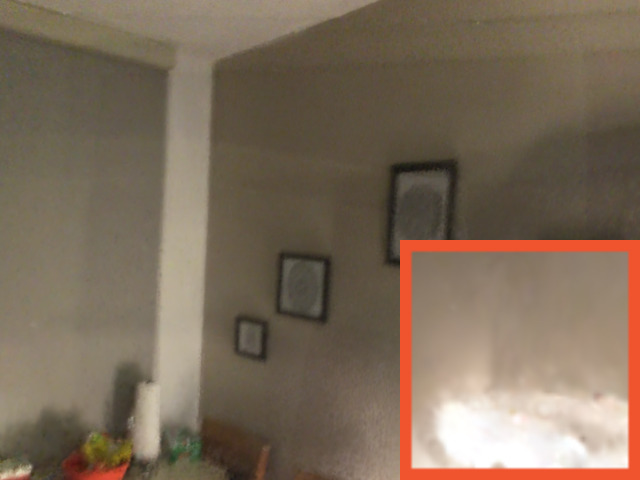} & \includegraphics[width=0.19\linewidth]{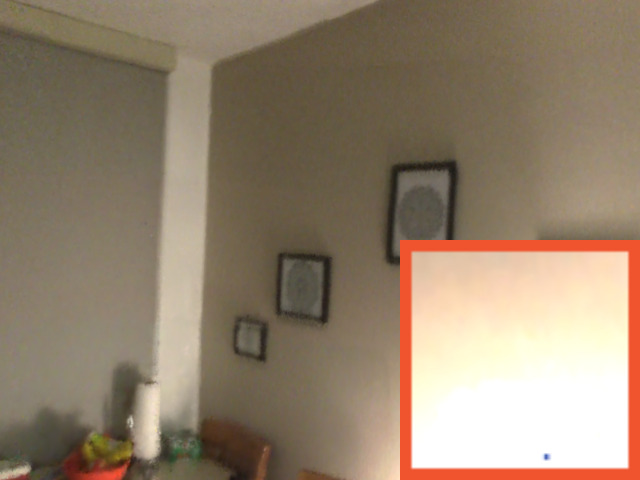} \\
            
            \rotatebox{90}{\quad\: \resizebox{0.07\textwidth}{!}{\texttt{0785-00}}} & \includegraphics[width=0.19\linewidth]{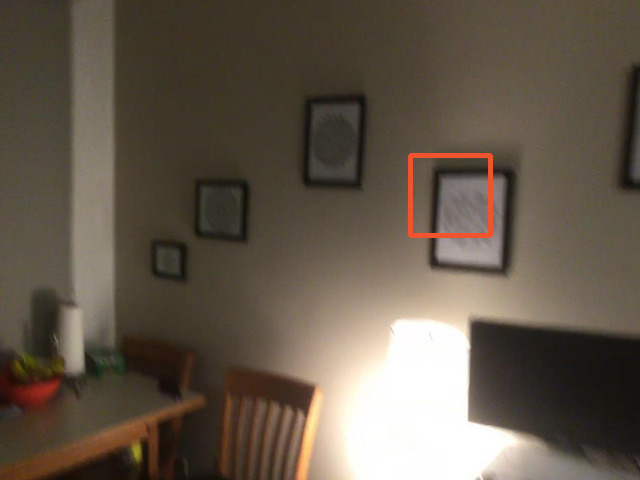}& \includegraphics[width=0.19\linewidth]{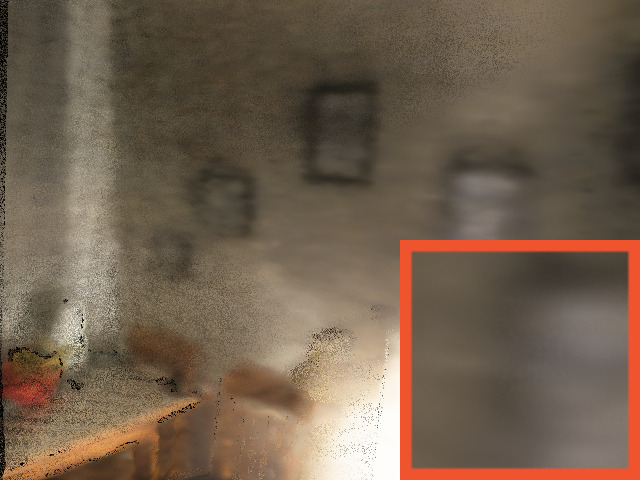} & \includegraphics[width=0.19\linewidth]{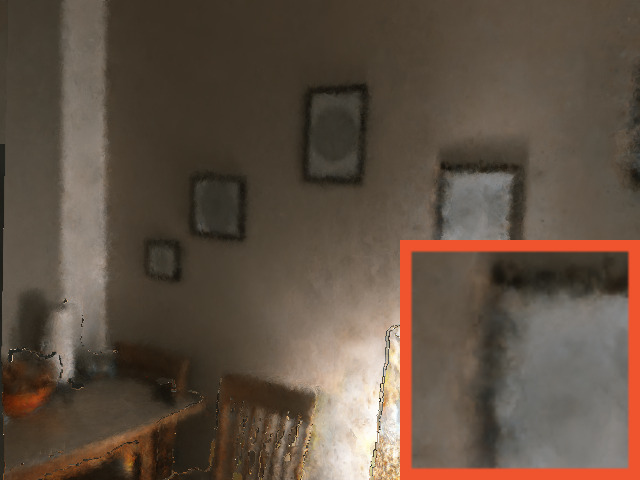} & \includegraphics[width=0.19\linewidth]{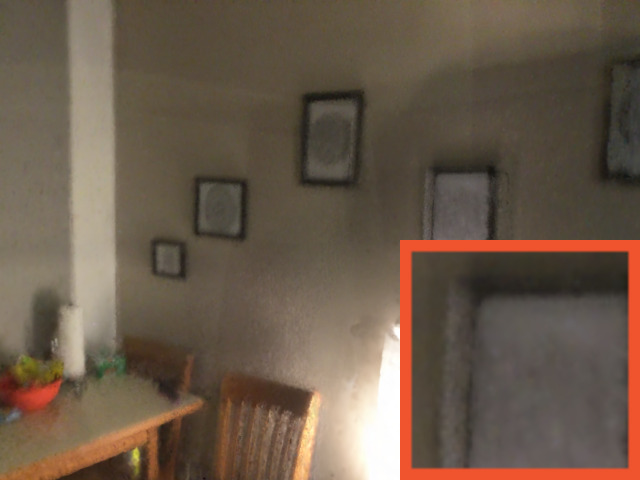} & \includegraphics[width=0.19\linewidth]{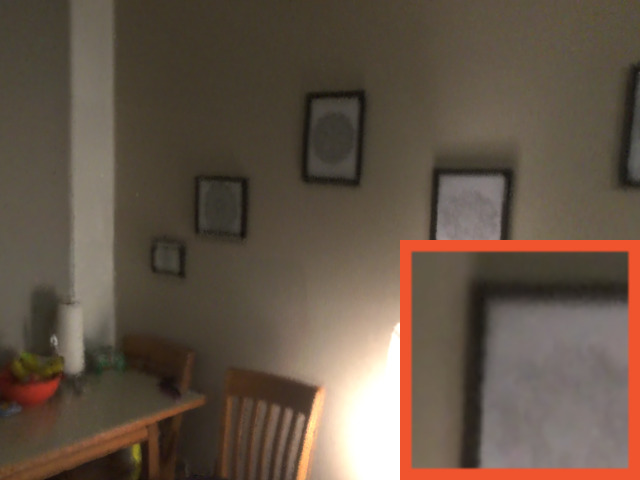} \\
            
            \rotatebox{90}{\quad\: \resizebox{0.07\textwidth}{!}{\texttt{0785-00}}} & \includegraphics[width=0.19\linewidth]{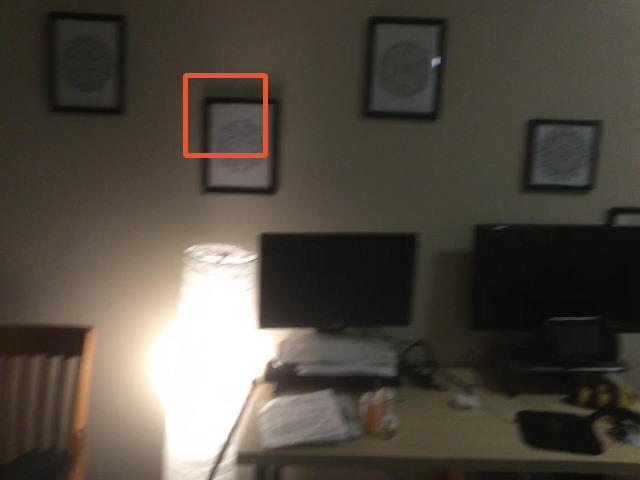}& \includegraphics[width=0.19\linewidth]{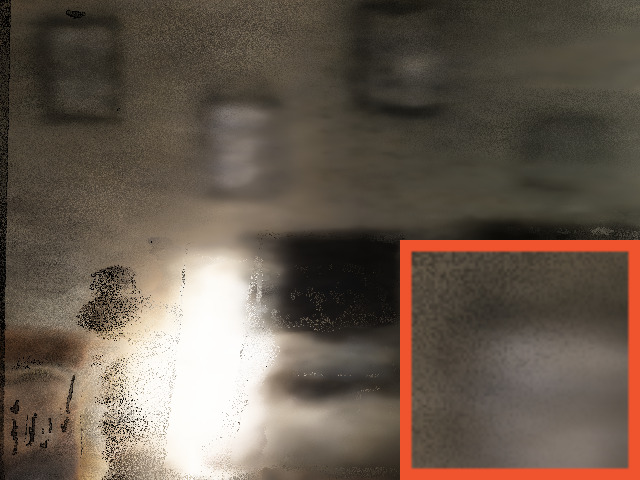} & \includegraphics[width=0.19\linewidth]{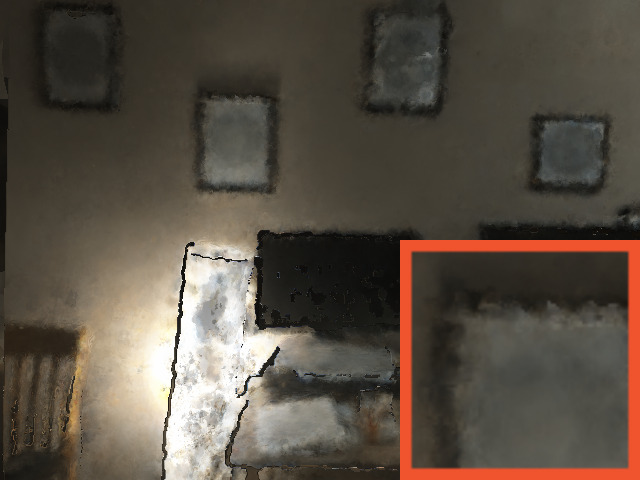} & \includegraphics[width=0.19\linewidth]{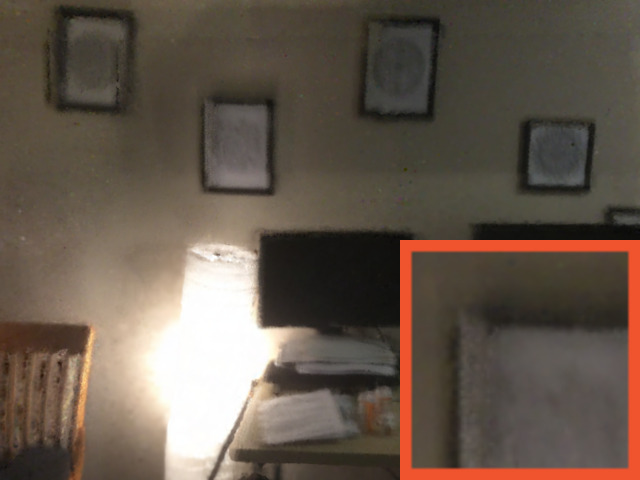} & \includegraphics[width=0.19\linewidth]{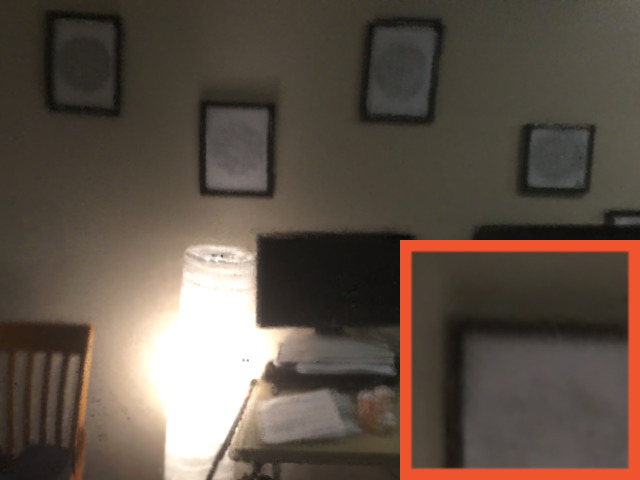} \\
        \end{tabular}
    \caption{Additional qualitative results in ScanNet~\cite{dai2017scannet} dataset.}
    \label{fig:additional_scannet}
\end{figure}

\begin{figure}[!ht]
    \centering
        \begin{tabular}{@{}c@{\,}c@{\,}c@{}}
            \resizebox{0.12\textwidth}{!}{Input Frame} & \resizebox{0.165\textwidth}{!}{$\text{NeRF-SLAM}^\dagger$~\cite{rosinol2023nerf}} & \resizebox{0.09\textwidth}{!}{\islam} \\
            
            \includegraphics[width=0.31\linewidth]{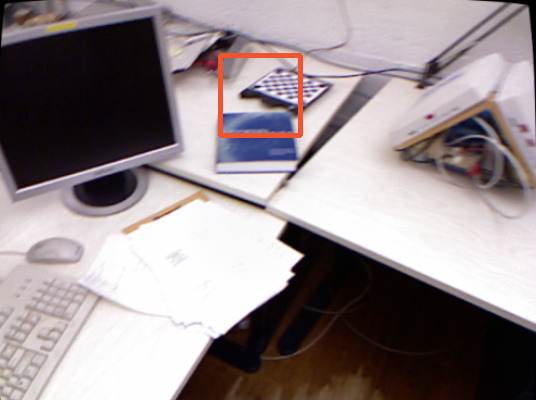}& \includegraphics[width=0.31\linewidth]{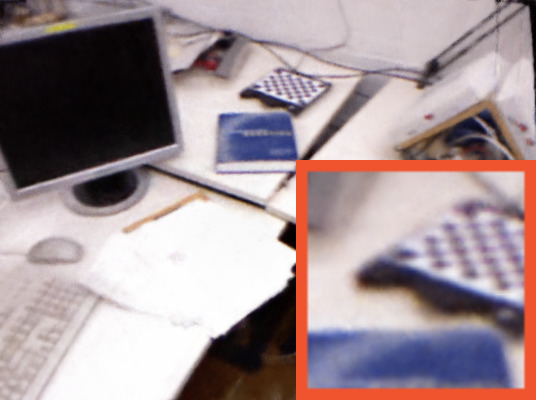} & \includegraphics[width=0.31\linewidth]{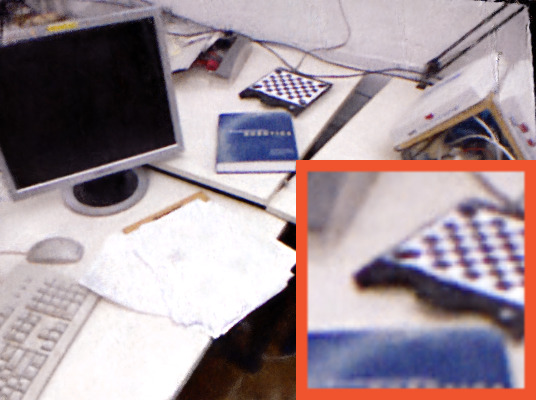} \\
            
            \includegraphics[width=0.31\linewidth]{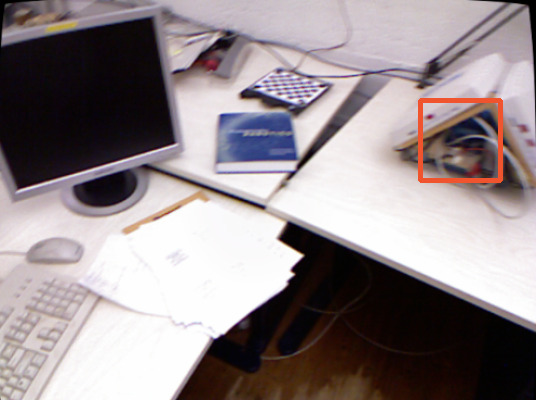}& \includegraphics[width=0.31\linewidth]{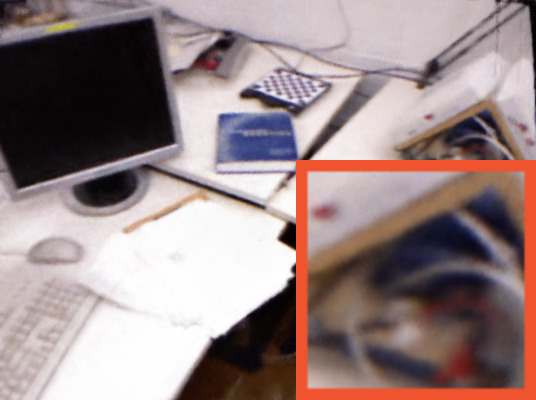} & \includegraphics[width=0.31\linewidth]{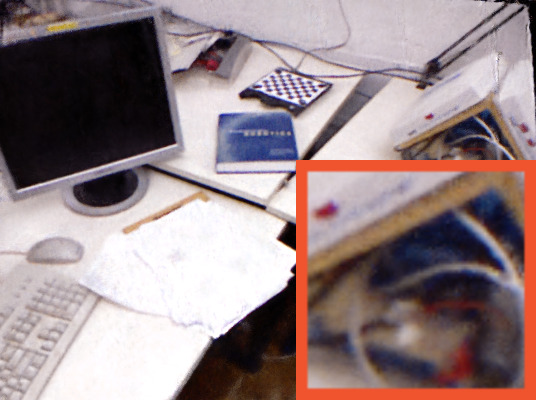} \\
            
            \includegraphics[width=0.31\linewidth]{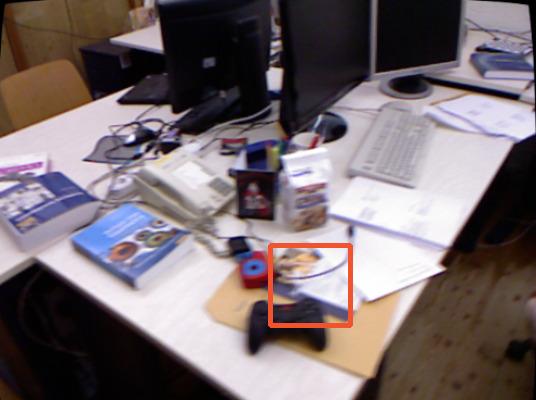}& \includegraphics[width=0.31\linewidth]{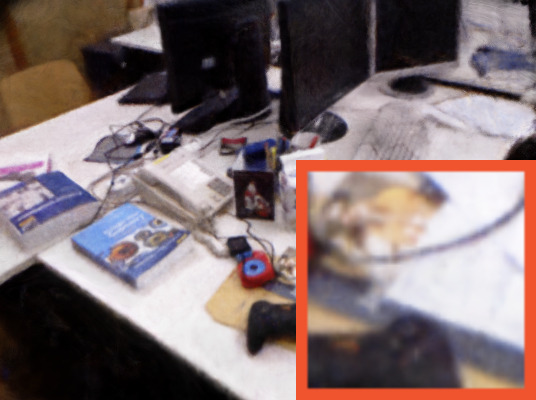} & \includegraphics[width=0.31\linewidth]{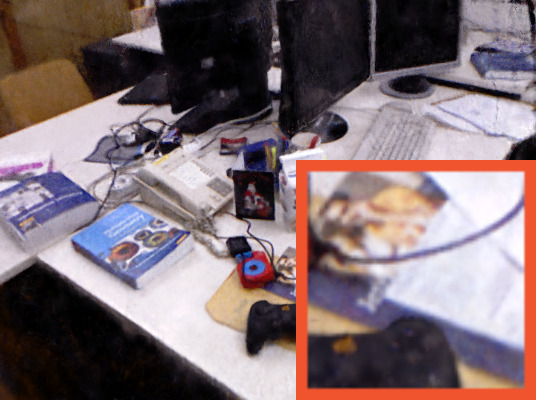} \\
        \end{tabular}
    \caption{Additional qualitative results in~\texttt{fr1/desk} of TUM-RGBD~\cite{sturm2012benchmark}}
    \label{fig:additional_tum_fr1}
\end{figure}

\begin{figure}[!ht]
    \centering
        \begin{tabular}{@{}c@{\,}c@{\,}c@{}}
            \resizebox{0.12\textwidth}{!}{Input Frame} & \resizebox{0.165\textwidth}{!}{$\text{NeRF-SLAM}^\dagger$~\cite{rosinol2023nerf}} & \resizebox{0.09\textwidth}{!}{\islam} \\
            
            \includegraphics[width=0.31\linewidth]{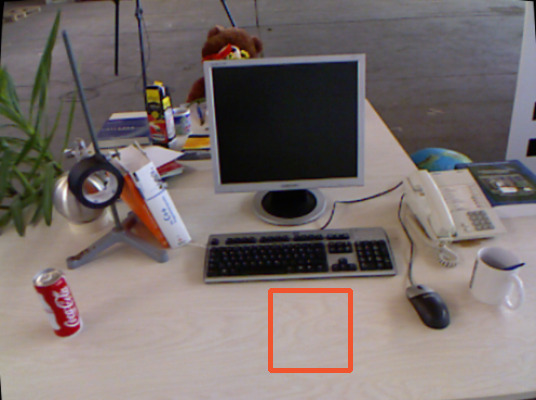}& \includegraphics[width=0.31\linewidth]{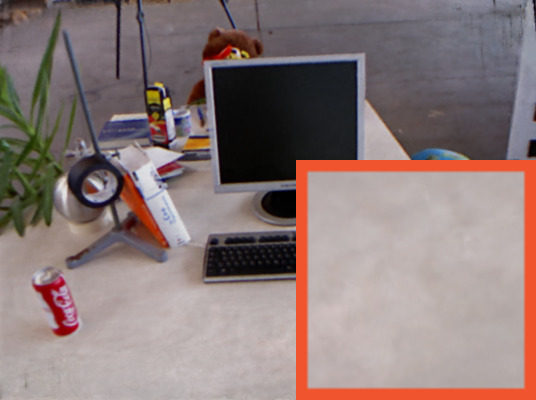} & \includegraphics[width=0.31\linewidth]{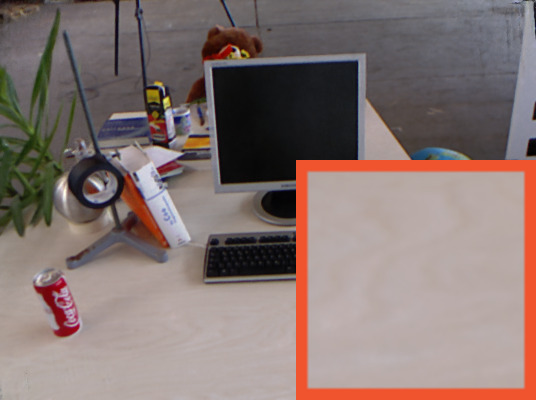} \\
            
            \includegraphics[width=0.31\linewidth]{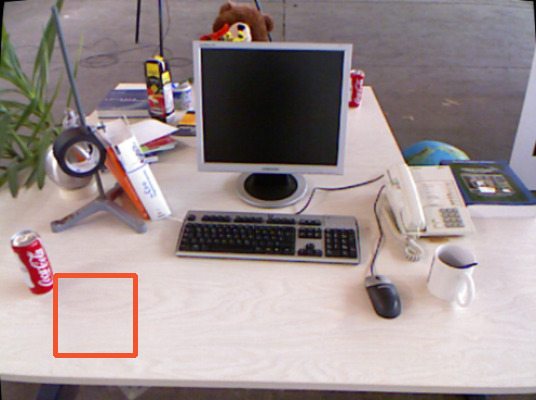}& \includegraphics[width=0.31\linewidth]{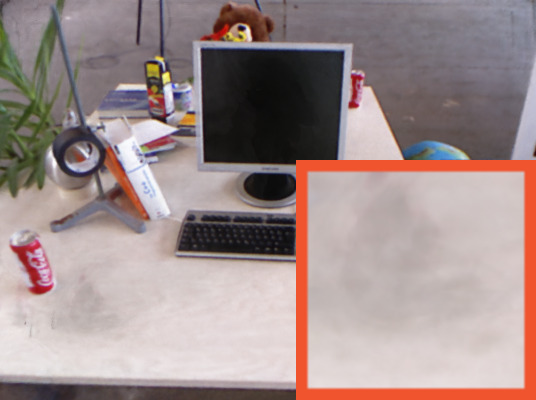} & \includegraphics[width=0.31\linewidth]{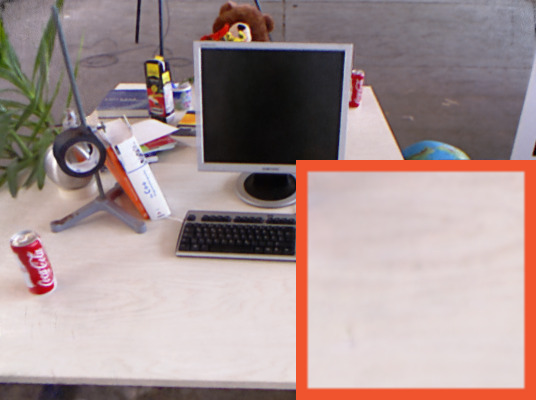} \\
            
        \end{tabular}
    \caption{Additional qualitative results in~\texttt{fr2/xyz} of TUM-RGBD~\cite{sturm2012benchmark}}
    \label{fig:additional_tum_fr2}
\end{figure}

\begin{figure}[!ht]
    \centering
        \begin{tabular}{@{}c@{\,}c@{\,}c@{}}
            \resizebox{0.12\textwidth}{!}{Input Frame} & \resizebox{0.165\textwidth}{!}{$\text{NeRF-SLAM}^\dagger$~\cite{rosinol2023nerf}} & \resizebox{0.09\textwidth}{!}{\islam} \\
            
            \includegraphics[width=0.31\linewidth]{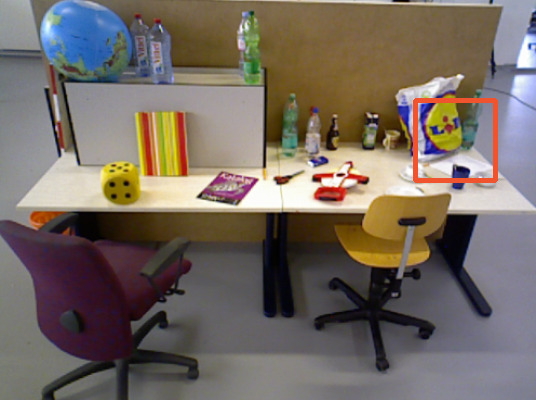}& \includegraphics[width=0.31\linewidth]{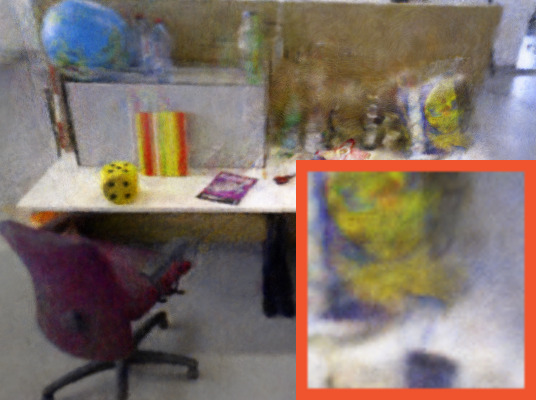} & \includegraphics[width=0.31\linewidth]{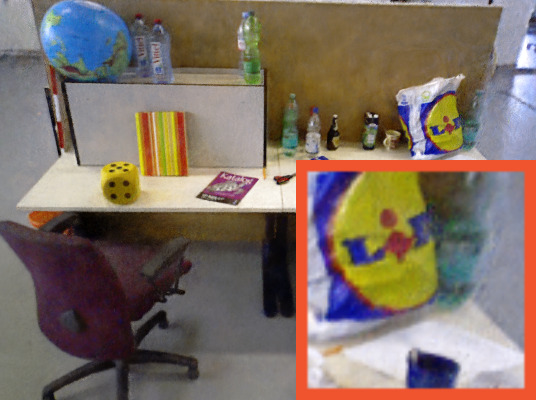} \\
            
            \includegraphics[width=0.31\linewidth]{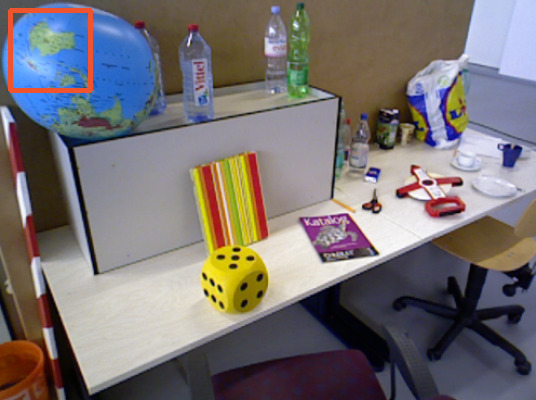}& \includegraphics[width=0.31\linewidth]{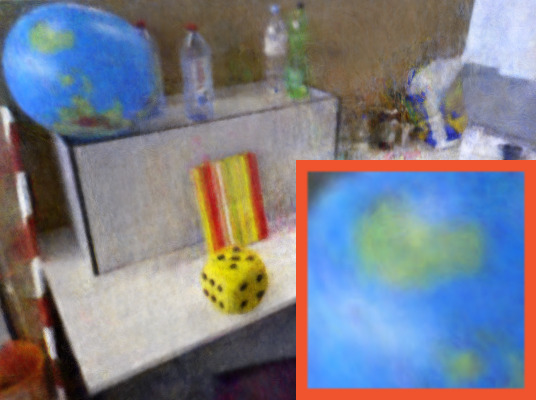} & \includegraphics[width=0.31\linewidth]{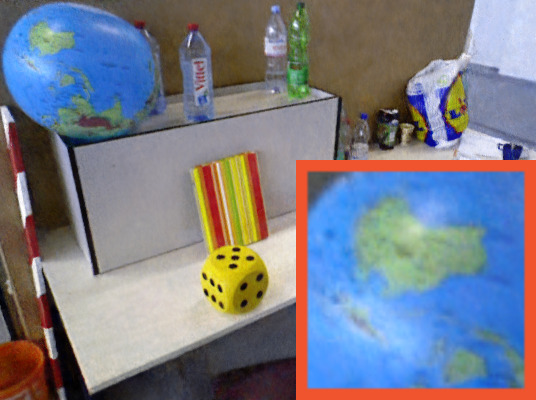} \\
            
            \includegraphics[width=0.31\linewidth]{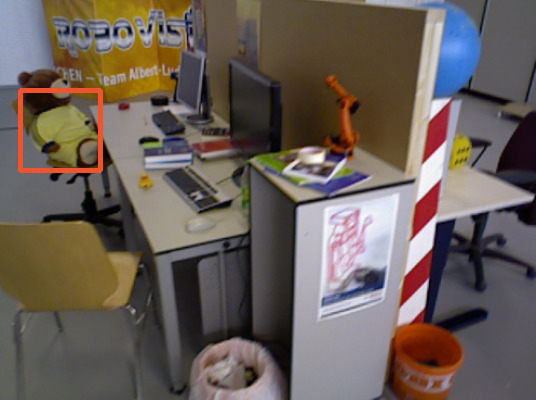}& \includegraphics[width=0.31\linewidth]{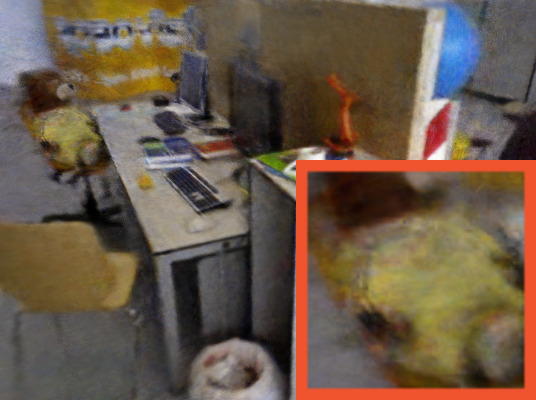} & \includegraphics[width=0.31\linewidth]{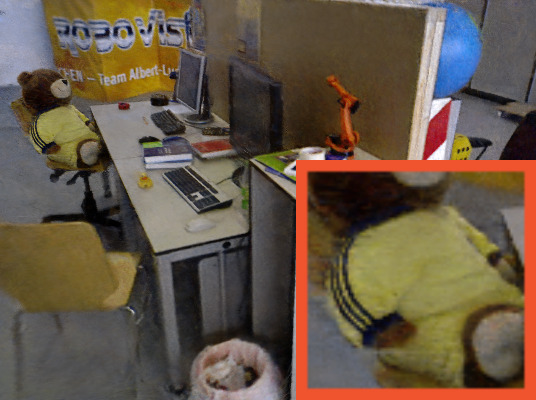} \\
            
            \includegraphics[width=0.31\linewidth]{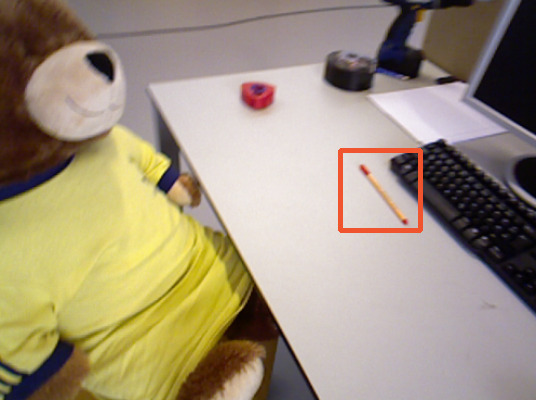}& \includegraphics[width=0.31\linewidth]{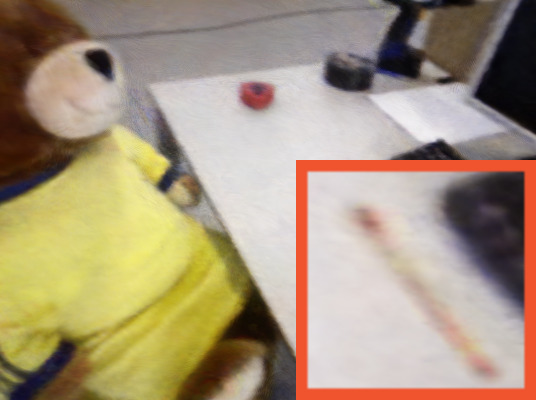} & \includegraphics[width=0.31\linewidth]{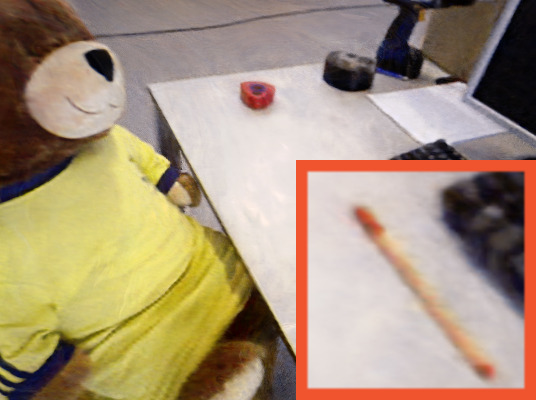} \\
            
            \includegraphics[width=0.31\linewidth]{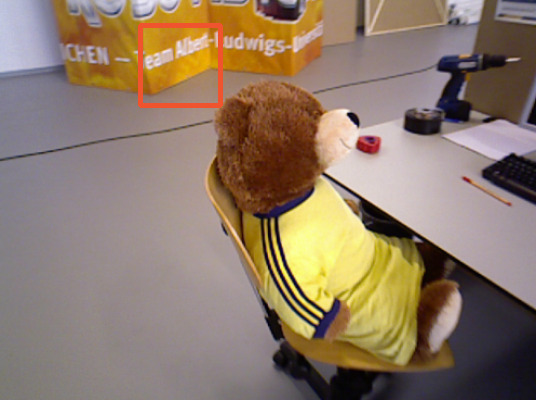}& \includegraphics[width=0.31\linewidth]{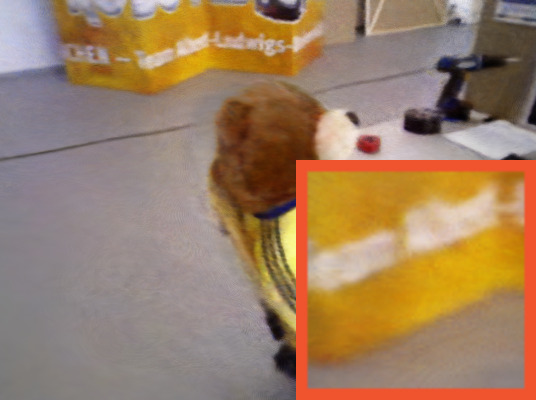} & \includegraphics[width=0.31\linewidth]{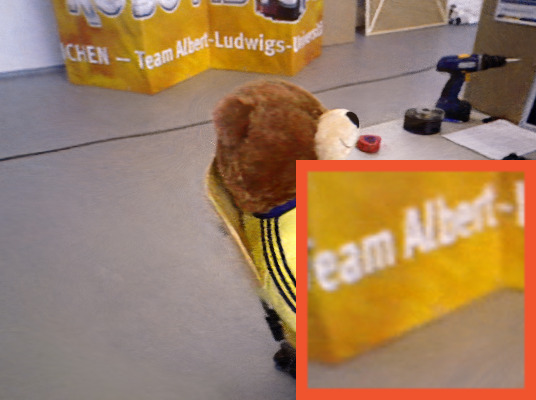} \\
            
            \includegraphics[width=0.31\linewidth]{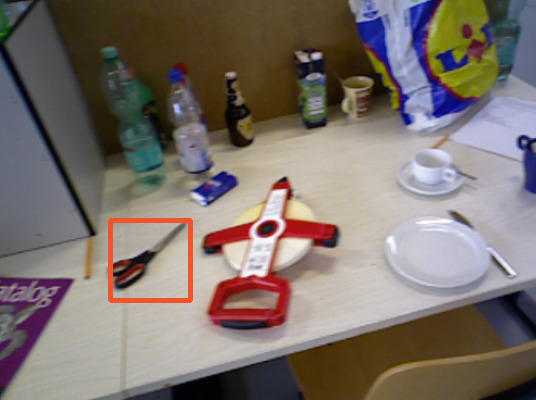}& \includegraphics[width=0.31\linewidth]{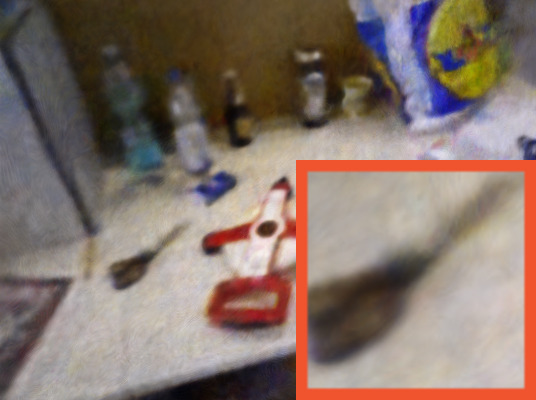} & \includegraphics[width=0.31\linewidth]{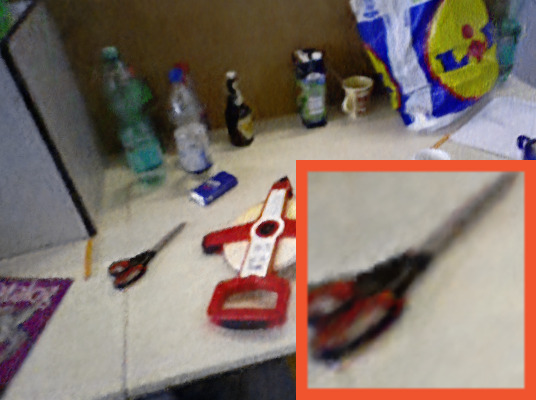} \\
            
        \end{tabular}
    \caption{Additional qualitative results in~\texttt{fr3/office} of TUM-RGBD~\cite{sturm2012benchmark}}
    \label{fig:additional_tum_fr3}
\end{figure}

\paragraph{Comparison with more baselines}
We further compare \islam~with two widely used neural RGBD-SLAM methods, Co-SLAM~\cite{wang2023co} and Point-SLAM~\cite{sandstrom2023point}. 
In~\cref{tab:more_baselines_rendering}, Both two methods show a lack of rendering performance in ScanNet's all scenes except scene~\texttt{0785-00}. 
As depicted in~\cref{fig:additional_scannet}, scene~\texttt{0785-00} contains severe appearance changes which can cause inaccurate mapping on homogeneous surfaces. 
SplaTAM shows inferior performance than other baselines.
However, by integrating our method into SplaTAM, it outperforms other methods with a large margin.

\paragraph{Qualitative results}
We conduct more qualitative comparison in \cref{fig:additional_scannet,fig:additional_tum_fr1,fig:additional_tum_fr2,fig:additional_tum_fr3}.
\Cref{fig:additional_scannet} demonstrates qualitative results of our RGBD-SLAM model in ScanNet.
\islam~renders sharp images even when blurry frames are used for mapping.
In the scene~\texttt{0785-00}, \islam~successfully reflects extreme appearance changes.
\Cref{fig:additional_tum_fr1,fig:additional_tum_fr3} show that our method reconstructs small objects,~\eg, a pen and a scissors, and areas with complex texture.
The effect of our tone-mapper can be seen through the removal of artifacts in~\cref{fig:additional_tum_fr2}.

\paragraph{Effects of the number of virtual cameras}
We use five virtual cameras to approximately simulate motion blur.
We conduct an ablation study on the number of virtual cameras. \Cref{fig:cam_num_ablation} shows results with our RGB-SLAM model in ~\texttt{Italian-flat-1}.
With an increasing number of cameras, the performance sees an upward trend.
Considering the trade-off between performance and runtime, we use five virtual cameras.

\section{Comparison with SLAM using 2D Image Deblurring}

\begin{table}[t]
\vspace{-0.5em}
\caption{Results of RGB-SLAM with enhanced image inputs.}
\label{tab:image_deblurring}
\centering
\resizebox{\linewidth}{!}{
\begin{tabular}{lcccccccc}
\toprule
\multicolumn{1}{c}{} & \multicolumn{4}{c}{TUM-RGBD}                                     & \multicolumn{4}{c}{Synthetic}                                    \\
Methods              & ATE ↓     & PSNR ↑          & SSIM ↑          & LPIPS ↓         & ATE ↓     & PSNR ↑          & SSIM ↑          & LPIPS ↓         \\
\midrule
NeRF-SLAM            & 3.21          & 26.89          & 0.817          & 0.227          & 2.47          & 27.41          & 0.828          & 0.310          \\
NeRF-SLAM+\cite{chen2022simple}    & 3.27          & 26.61          & 0.825          & 0.202          & 3.77          & 27.65          & 0.820          & \textbf{0.237} \\
NeRF-SLAM+\cite{chen2022simple}+TM & 2.80          & 26.22          & 0.814          & 0.215          & 3.48          & 27.13          & 0.809          & 0.254          \\
$I^2$-SLAM              & \textbf{1.28} & \textbf{29.40} & \textbf{0.861} & \textbf{0.151} & \textbf{1.48} & \textbf{29.59} & \textbf{0.878} & 0.256          \\
\bottomrule
\end{tabular}
}
\end{table}

We conduct comparisons with the RGB-SLAM with enhanced image inputs.
We first deblur input frames with NAFNet~\cite{chen2022simple} and run NeRF-SLAM with deblurred images.
In~\cref{tab:image_deblurring} and~\cref{fig:image_deblurring}, even with the deblurred inputs, there is no significant improvement in tracking and rendering quality.
While integrating our tone mapper~(TM) enhances trajectory accuracy, it still shows lower performance compared to $I^2$-SLAM.
This is attributed to the multi-view inconsistency in image deblurring process.

\end{document}